\documentclass[letterpaper, 10 pt, conference]{IEEEtran}  % Comment this line out if you need a4paper

\IEEEoverridecommandlockouts 
\usepackage{cite}

\usepackage{amsmath,amssymb,amsfonts}
\usepackage{graphicx}
\usepackage{textcomp}

\DeclareMathOperator*{\argmin}{arg\,min}
\usepackage[table,xcdraw]{xcolor}
\usepackage{authblk}

\usepackage{enumitem}
\usepackage{hyperref}
\usepackage{breqn}
\usepackage{algpseudocode}
\usepackage[linesnumbered,ruled,lined]{algorithm2e}

\usepackage{array}
\usepackage{makecell}
\usepackage{tabularray}

\usepackage{svg}
\usepackage{graphicx}
\usepackage[caption=false,font=footnotesize]{subfig}
\usepackage{multirow}
\usepackage{balance}

\renewcommand\IEEEkeywordsname{Keywords}

\allowdisplaybreaks

\definecolor{revision_color}{rgb}{0.0, 0.0, 0.0}

\begin{document} 

\title{\LARGE \bf Flow-Opt: Scalable Centralized Multi-Robot Trajectory Optimization with Flow Matching and Differentiable Optimization}
% \author{Anonymous Authors}
\author{Simon Idoko, Prajyot Jadhav, Arun Kumar Singh\thanks{\small{All authors are with the University of Tartu, Estonia. This research was in part supported by grant PSG753 from Estonian Research Council, collaboration project LLTAT21278 with Bolt Technologies and co-funded by European Union and Estonian Research Council through the  TEM-TA101 project.
Emails: cisimon7@gmail.com, arun.singh@ut.ee. Our code will be available at \url{https://github.com/cisimon7/FlowOpt}}} 
}
\maketitle

% \definecolor{reed}{rgb}{0.0, 0.0, 0.0}

%%%%%%%%%%%%%%%%%%%%%%%%%%%%%%%%%%%%%%%%%%%%%%%%%%%%%%%%%%%%%%%%%%%%%%%%%%%%%%%%%%%%%%%%%%%%%%%%%%%%%%%%%%%%%%%%%%%%%%%%%%%%%%%%%%%%%%%%%%%%%%%%%%%%%%%%%%%%%%%
\begin{abstract}

Centralized trajectory optimization in the joint space of multiple robots allows access to a larger feasible space that can result in smoother trajectories, especially while planning in tight spaces. Unfortunately, it is often computationally intractable beyond a very small swarm size. In this paper, we propose Flow-Opt, a fast \textcolor{black}{learning-based approach for providing high-quality approximations of centralized multi-robot trajectory optimization}. Specifically, we reduce the problem to first learning a generative model to sample different candidate trajectories and then using a learned Safety-Filter(SF) to ensure fast inference-time constraint satisfaction. We propose a flow-matching model based on a diffusion transformer (DiT) augmented with state and map encoders, as the generative model. We develop a custom solver for our SF and equip it with a neural network that predicts context-specific initialization. The initialization network is trained in a self-supervised manner, taking advantage of the differentiability of the SF solver. We advance the state-of-the-art in the following respects. First,  we show that we can generate trajectories for tens of robots in cluttered environments in a few tens of milliseconds. This is several times faster than existing centralized optimization approaches. Moreover, our approach generates smoother trajectories orders of magnitude faster than competing baselines based on diffusion models. Second, each component of our approach can be batched, allowing us to solve a few tens of problem instances in a fraction of a second. We believe this is the first such result; no existing approach provides such capabilities. Finally, our approach can generate a diverse set of trajectories between a given set of start and goal locations, which can capture different collision-avoidance behaviors.

\end{abstract}

\begin{IEEEkeywords}
Optimization and control, Data-based approaches, multi-robot systems.
\end{IEEEkeywords}

\def\abstractname{Note to Practitioners}

\begin{abstract}
In applications like warehouse automation, the quality of multi-robot trajectories is critical for maximizing throughput and efficiency. While centralized optimizers can produce high-quality, globally coordinated plans, they are often considered too slow for practical use, forcing a trade-off for faster, sub-optimal distributed methods. Our work eliminates this compromise. By leveraging a combination of supervised and self-supervised learning, our approach achieves the best of both worlds: it generates high quality (in terms of smoothness, arc length), fixed-final-time trajectories for tens of robots at near-real-time speeds. This makes it possible to coordinate entire teams of robots in milliseconds, ensuring that they meet precise timing constraints. The speed of our method also unlocks new operational paradigms, such as centralizing and coordinating multiple geographically separate robot teams in parallel within the same warehouse.
\end{abstract}

\section{Introduction}
Deploying multi-robot systems, including quadrotors and autonomous vehicles, plays a crucial role in applications ranging from search and rescue operations to warehouse automation and large-scale environmental mapping. Generating feasible trajectories that coordinate the robots behaviors is a fundamental requirement in multi-robot deployment. Furthermore, generating diverse and feasible multi-robot motions can also be used to construct data-driven simulations to train navigation policies, as in \cite{8865441, Mavrogiannis2020BGAPBS, prorok2022holy}. 

Current approaches to multi-robot motion planning are primarily divided into two paradigms. On the one hand, we have distributed approaches that allow each robot to plan its motion independently \cite{adajania2023amswarm, Luis2019OnlineTG, Grfe2022EventtriggeredAD, chen2023toward} using communication with other robots or predictions of their future trajectories to maintain coordination. These methods are computationally fast but only consider inter-robot interactions implicitly, which limits the feasible solution space. In contrast, centralized methods plan in the joint space of all robots \cite{6385823, Rastgar2020GPUAC}. This can dramatically improve the trajectory quality in terms of smoothness, arc-length, and other metrics. Moreover, the enhanced feasible space can also prove crucial when planning in constrained or cluttered environment. \textcolor{black}{However, the primary challenge of centralized trajectory optimization is that the number of inter-robot collision constraints grows steeply with the number of robots \cite{6385823}, leading to poor scalability as the number of robots increases.}

\subsubsection*{Generative Models Hold The Key} In this paper, we aim to improve the scalability of centralized multi-robot trajectory optimizers, as the high-quality trajectories they produce are critical for applications such as warehouse automation. Our approach leverages the recent success of applying generative models for motion planning. Specifically, models such as diffusion policies have been extensively used to plan optimal trajectories for both single \cite{edmp}, \cite{mpd} and multiple robots \cite{mmd}, \cite{smd}, \cite{disco}. The underlying theme in these works is that they first train a generative model on a dataset of expert trajectories. Then at inference time, they sample novel trajectories and further refine them to improve satisfaction of constraints such as collision avoidance. 

\subsubsection*{Core Challenges }While diffusion policies have demonstrated remarkable performance, some key challenges still remain. In particular, the denoising process during inference can be painfully slow. This computational cost further increases when the inference-time correction strategies are embedded in the denoising process. For example, \cite{smd} solves a complex non-convex optimization within each denoising step to guide the diffusion policies towards constraint satisfaction. In the case of multi-robot planning, such optimization can scale poorly with the number of robots.

\subsubsection*{Our Approach and its Novelty} At a higher level, our approach follows the same paradigm of combining generative modeling with rule-based inference-time refinement. However, we differ from prior works such as \cite{edmp}, \cite{mpd}, \cite{mmd}, \cite{smd}  in two key ways. First, instead of diffusion policies, our pipeline is built around Flow Matching. Conceptually, flow policy operates similarly to diffusion models by iteratively transforming random noise into smooth trajectories. However, unlike diffusion, which relies on stochastic differential equations, flow policies are based on ordinary differential equations (ODEs), resulting in a simpler training process and faster inference. We still leverage the Diffusion Transformer (DiT) backbone \cite{peebles2023scalable}, augmented with permutation invariant start-goal and map encoders to build our flow policy. To the best of our knowledge, this is the first application of Flow Matching to the multi-robot trajectory planning problem.

Our second novelty lies in how we perform inference-time refinement. Rather than embedding the refinement strategy within the denoising or unrolling process of the flow policy, we apply it only to the trajectory obtained at the final step. We hypothesize that the intermediate trajectories during early denoising steps are too noisy for meaningful modification. We formulate inference-time refinement as an optimization problem, which we henceforth call the Safety Filter (SF). We develop a custom solver for SF that can be accelerated over GPUs. This allows for hundreds of different instances of our SF to be run in parallel. This proves crucial for simultaneously refining a batch of trajectories. To further improve the computational performance of SF, we equip it with an initialization network that provides context-specific initialization to accelerate the convergence of the SF solver. The initialization network is trained in a self-supervised manner by leveraging the end-to-end differentiability of our custom SF solver.

\subsubsection*{Benefits of Our Approach} Our approach improves upon both model-based optimization and data-driven approaches for centralized multi-robot trajectory optimization. First, it achieves up to an order of magnitude speedup over \cite{Rastgar2020GPUAC} while generating trajectories of comparable quality. It is also both faster and more reliable than the batch-sequential approach of \cite{park2020efficient}. Compared to the recent diffusion-based multi-robot planning approach in \cite{mmd}, our method produces substantially smoother trajectories while reducing computation time by a factor of 160. Moreover, our framework enables tens of independent problem instances to be solved in parallel within a fraction of a second. We believe this capability opens new opportunities for large-scale deployment in warehouse automation and data-driven simulation environments.

% The rest of the paper is organized as follows. Section \ref{sec:problem} presents the problem formulation. Our solution process is described in Section \ref{sec:method}. Section \ref{rel_work} contrasts our work with the existing literature and how our approach fills the current gaps. We kept this related works comparison after description of our methods to better pin-point how we improve upon different existing works. Section \ref{val} presents the validation and benchmarking results. For the ease of exposition, most of the derivations are presented in Appendix \ref{Appendix}

\textcolor{black}{The remainder of the paper is structured as follows. Section \ref{rel_work} reviews related works and highlights our improvement over the state-of-the-art. Section \ref{sec:problem} formulates the problem, and Section \ref{sec:method} details our proposed solution. Finally, Section \ref{val} presents the experimental validation and results, with detailed derivations deferred to Appendix \ref{Appendix} for readability.}

\begin{figure*}
    \centering
    \includegraphics[scale=0.55]{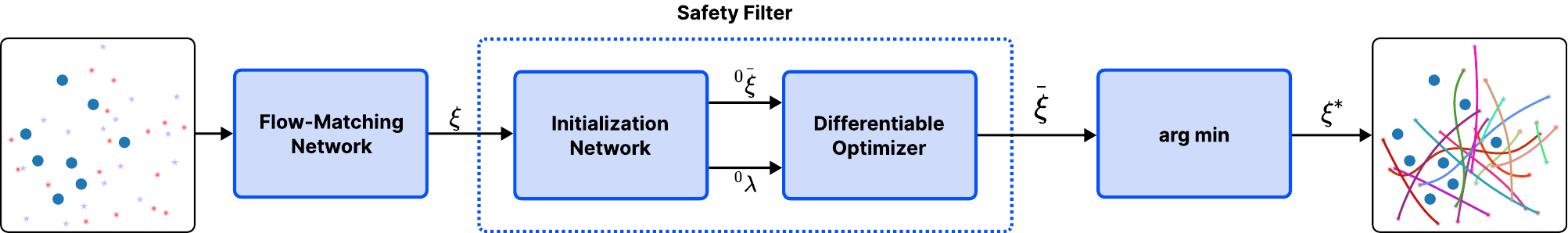}
    % \includesvg[scale=0.55]{figs/pipeline.svg}
    \caption{\footnotesize{The overview of our multi-robot trajectory pipeline. It has two core components: a trained flow policy and a safety-filter.  The trained flow policy takes in start and goal positions of the robots and static obstacle placements (additionally, velocity for dynamic obstacles) and outputs a distribution of trajectories, $\boldsymbol{\xi}$. The multiple flow sampled trajectories are refined in parallel through a safety-filter, and the trajectory with the lowest constraint residual and smoothness is output as the optimal solution. Our SF is accelerated by an initialization network that is conditioned on the samples drawn from the flow policy }}
    \label{fig:pipeline}
\end{figure*}

\section{Related Works}\label{rel_work}
\textcolor{black}{In this section, we review closely connected literature and how our proposed approach fills the key gaps in these works.}

\subsection{Centralized Trajectory Optimizers} 
\noindent The centralized approach can provide many advantages, such as one-shot feasible trajectory generation between start and goal with a fixed-final time and better trajectory quality due to access to a larger feasible space \cite{dabestani2024joint}. This can be particularly important in applications such as warehouse automation, where robots need to reach a pickup location in a specified time. Centralized methods also find application in drone cinematography \cite{nageli2017real}, drone racing \cite{di2023cooperative}, interaction-aware planning \cite{chen2023interactive}, coordination of connected vehicles \cite{dabestani2024joint}, and target tracking \cite{ramachandran2023resilient}. Furthermore, centralized optimizers can provide a framework for generating diverse swarm behaviors, which can be used to train navigation policies through imitation or reinforcement learning \cite{prorok2022holy}.

Given these advantages, significant research effort has focused on improving the computational scalability of centralized trajectory optimization. The underlying idea primarily revolves around decomposing the overall optimization into smaller parallelizable subproblems. To this end, the Alternating Direction Method of Multipliers (ADMM) has emerged as a particularly effective tool \cite{ni2022robust}, \cite{huang2023decentralized}, \cite{salvado2022dimopt}, \cite{bento2013message}, \cite{halsted2021survey}, \cite{10288223}. ADMM exploits the fact that the only coupling between different robots in optimization \eqref{cost}-\eqref{boundary_con} stems from the inter-robot collision avoidance constraints \eqref{inter_robot_con}. Thus, it breaks down the whole centralized problem into decoupled optimization blocks. Other works such as \cite{hamer2018fast}, \cite{Rastgar2020GPUAC} explore a different direction for improving scalability by reformulating the computation in a form that can be easily accelerated over GPU cores.  

\subsubsection*{Our Improvement} Our approach delivers a substantially faster approach for centralized multi-robot trajectory optimization than the above cited works. One of the main reasons is the computational efficiency of the trajectory optimizer underlying our approach. As detailed in the Appendix, our optimizer builds upon \cite{Rastgar2020GPUAC} by incorporating additional constraints for workspace and environment collision. Moreover, we introduce a batchable implementation of \cite{Rastgar2020GPUAC} that can solve several hundred problem instances in parallel in a fraction of a second (see Fig.\ref{fig_scalability}).

% \noindent \emph{Our Improvement over the SOTA:} Our proposed work improves \cite{rastgar2021gpu} by using generative models and self-supervised learning to guide the solution process. Moreover, we develop a batched version of \cite{rastgar2021gpu} to ensure that the SF training and inferencing discussed in the previous section can be performed in a scalable manner.

% \subsection{Combining Learning With Trajectory Optimization} 

% \noindent The use of trajectory optimization to perform inference-time corrections on predictions from learned models has been extensively explored in existing literature. There are two broad ways to do this. We can directly use the predictions from the learned model to initialize the trajectory optimizer \cite{celestini2024transformer}, \cite{pulver2021pilot}. In our context, this would entail directly using the VQ-VAE/CVAE predictions to initialize the solution process of \eqref{eq::form}. Alternatively, one can adopt the SF route and only improve the constraint satisfaction of the model predictions \cite{wabersich2021predictive}. Our work adopts the SF approach and, we show in Section \ref{val} (see Fig.\ref{model_collapse}) that it proves crucial for generating multi-modal feasible solutions.

\subsection{Generative Models for Motion Planning}
\noindent Generative models offer a robust, data-driven approach to motion planning by learning a distribution over expert trajectories, obtained either from human demonstration or synthetic solvers. Conditional Variational Autoencoders (CVAEs) \cite{ichter2018learning} are computationally efficient but often struggle to capture the diverse, multi-modal nature of many planning problems. Vector Quantized Variational Autoencoders (VQ-VAEs) address this by using a discrete latent space \cite{idoko2024learning} capable of representing distinct solution modes, though they can be more challenging to train. More recently, diffusion \cite{edmp}, \cite{mpd}, \cite{mmd} \cite{luo2024pot}, \cite{smd}, \cite{jiang2023motiondiffuser} and flow matching policies \cite{nguyen2025flowmp} have become prominent approaches for motion planning. Both frameworks can effectively learn complex, multi-modal trajectory distributions. A unique strength of diffusion and flow policies is that their learned distribution can be adapted at inference times through cost or constraint functions \cite{edmp}, \cite{mpd}. In this sense, both models can be interpreted as learning structured priors over offline trajectory datasets.

The use of diffusion policies for centralized multi-robot motion planning has only recently been explored, with \cite{mmd}, \cite{smd} being the only works, to the best of our knowledge.  In \cite{mmd}, diffusion priors are learned over the dataset of just single robot motions. At inference time, these distributions are steered through the use of conflict-based search algorithms \cite{sharon2015conflict}. In contrast, \cite{smd} uses the output of \cite{mmd} to learn the prior over multi-robot trajectories followed by the use of trajectory optimization to satisfy safety constraints. The work presented in \cite{jiang2023motiondiffuser} is somewhat related as it learns diffusion policies over multiple vehicles. However, the focus is on trajectory prediction for autonomous driving, hence the constraint residuals reported in \cite{jiang2023motiondiffuser} are too high for the learned policy to be used for navigation.

\subsubsection*{Our Improvement} We present, to the best of our knowledge, the first application of flow matching for multi-robot motion planning. Compared to diffusion-based approaches such as \cite{mmd}, \cite{jiang2023motiondiffuser}, \cite{smd}, inference with our flow model is substantially faster. Furthermore, prior works \cite{mmd}, \cite{smd} build on the unconditional diffusion model of \cite{mpd} whereas our flow policy is explicitly conditioned on start-goal positions and environmental context (Fig.\ref{fig:flow_model}, Section \ref{flow_network}), obtained through a permutation invariant encoder. This conditioning enables more structured and task-aware trajectory generation.

\subsection{Use of Safety Filter (SF)}
\noindent Predictions from learned neural network models typically struggle to satisfy constraints \cite{dontidc3}, even when trained extensively on datasets of feasible trajectories. As a result, it is common to employ a safety filter to perform a correction at the inference time by projecting the predictions from the learned models onto the feasible set \cite{wabersich2021predictive}. Recent works such as \cite{romer2025diffusion}, \cite{xiao2023safediffuser} integrate SF with diffusion models to enforce constraints like collision avoidance.

\subsubsection*{Our Improvement} In many existing data-driven pipelines, the safety filter performs the majority of the computational workload, often becoming the primary bottleneck. Our approach fully acknowledges this and as a result introduces a self-supervised learning pipeline (Fig.\ref{unroll_learning}) to improve the computational efficiency of SF. We are not aware of any existing learning accelerated SF for multi-robot planning.

% Our use of SF differ from the existing works in three aspects. First, unlike works like \cite{sf_qp} that performs only one-step corrections, we project the entire predicted multi-step trajectory onto the feasible set. Second, we learn context-specific initialization policy for the SF in order to improve the computational efficiency of the inference time correction. Finally, our SF can be efficiently operated in batch fashion which comes in handy to simultaneously correct multiple predictions from a generative models. 

% \noindent \textbf{Works on SF:} Predictions from learned neural network models typically struggle to satisfy constraints \cite{dc3}, even though they might have been trained extensively on datasets of feasible trajectories. Thus, it is common to employ a safety filter to perform a correction at the inference time by projecting the predictions from the learned models onto the feasible set \cite{sf_1}, \cite{sf_2}, \cite{sf_qp}. Our use of SF differ from the existing works in three aspects. First, unlike works like \cite{sf_qp} that performs only one-step corrections, we project the entire predicted multi-step trajectory onto the feasible set. Second, we learn context-specific initialization policy for the SF in order to improve the computational efficiency of the inference time correction. Finally, our SF can be efficiently operated in batch fashion which comes in handy to simultaneously correct multiple predictions from a generative models. 

\subsection{Learning to Warm-start Optimization} 
\noindent The performance of non-convex trajectory optimizers depends heavily on the quality of initial guess. Consequently, there is a strong motivation for adopting data-driven approaches to come up with problem specific initialization (warm-start) for the optimizers. A conventional approach involves training a neural network directly over the dataset of optimal solutions. The predictions from the learned model can then be used to initialize the optimizer \cite{celestini2024transformer}, \cite{pulver2021pilot}. However, in our experience such an approach often does not perform satisfactorily on complicated problems such as multi-robot planning (e.g see Fig.\ref{res_16}-\ref{res_32}). This performance gap arises because the initialization model is agnostic to how its predictions are utilized by the downstream solver. To address this, previous works \cite{sambharya2024learning} have proposed hybrid architectures like that shown in Fig.\ref{unroll_learning} in which the optimizer is embedded as a differentiable layer within the training pipeline.

\subsubsection*{Our Improvement} The end-to-end warm-start learning of \cite{sambharya2024learning} has only been applied to convex problems which has two important consequences. First, for this class of problems, the solution process can be easily cast as a chain of differentiable computations. In contrast, off-the-shelf non-convex optimizers often rely on non-differentiable steps such as line-search. Thus, extending \cite{sambharya2024learning} to non-convex problems requires developing custom end-to-end differentiable solvers. The derivations presented in the Appendix \ref{Appendix} precisely fulfill this objective. 

\textcolor{black}{Second, the learning pipeline of \cite{sambharya2024learning} only considers a fixed-point residual as the loss for the self-supervised learning of warm-start  because convex problems have only one solution. In contrast, we augment the loss with additional terms to guide the learning process for SF (see \eqref{NN_loss}).}

\subsection{Reinforcement Learning Approaches}
\noindent Besides generative model driven imitation learning, reinforcement learning has also been used for multi-robot trajectory planning \cite{fan2020distributed}, \cite{tan2020deepmnavigate}, \cite{qin2023srl}, \cite{han2020cooperative}. However, most works including these cited ones consider reactive navigation. In contrast, the focus in our work is on multi-step trajectory generation. 

% Subsequently, the predictions from the learned model can be used to initialize the same optimizer/FP solver from where the dataset was generated \cite{celestini2024transformer}, \cite{pulver2021pilot}. However, such approaches often fall short in practice.  On difficult problems, such as our multi-robot problem, this approach can perform even worse than naive zero initialization of the optimization solver. This is because the notion of "good initialization" is very solver-specific \cite{sambharya2024learning}. In other words, initializations that work for an interior-point solver may not provide any computational gains if we replace them with a gradient descent routine. Thus, authors in \cite{sambharya2024learning} recommend hybrid architectures like that shown in Fig.\ref{unroll_learning} wherein, during the training process, the neural network layers are aware of how the downstream solver uses its predictions. Our work extends \cite{sambharya2024learning} to the non-convex multi-robot trajectory optimization setting. 

\section{Problem Formulation}\label{sec:problem}
\subsubsection*{Notations} We will use normal-font letters to represent scalars. The vectors and matrices will be represented by bold-faced lower and uppercase, respectively. 

\begin{table}[!t]
    \centering
    \caption{Summary of Notation}
    \label{tab:notation}
    \scriptsize
    \begin{tabular}{|l|l|}
        \hline
        \textbf{Symbol} & \textbf{Meaning} \\ 
        \hline
        $t$ & Time-scale of flow ODEs \\ \cline{1-2}
        $k$ & Time-step of robot trajectory \\ \cline{1-2}
        $n$ & Number of robots \\ \cline{1-2}
        $n_{\text{obs}}$ & Number of obstacles in the environment \\ \cline{1-2}
        $n_d$ & Workspace dimension ($n_d \in \{2,3\}$ for 2D/3D) \\ \cline{1-2}
        $(\frac{a}{2}, \frac{a}{2}, \frac{b}{2})$ & Robot Dimension modeled as spheroid   \\ \cline{1-2}
        $(a_o, a_o, b_o)$ & Obstacle dimension inflated by the robot size.   \\ \cline{1-2}
        $n_{\xi}$ & Order of polynomial basis for trajectory representation \\ \cline{1-2}
        $S$ & Flow model hyperparameter: Trajectory sequence length \\ \cline{1-2} 
        $\left[\boldsymbol{\xi}_{i,x}, \boldsymbol{\xi}_{i,y}, \boldsymbol{\xi}_{i,z}\right]$  & Polynomial trajectory basis for $i^{th}$ robot \\  \cline{1-2}
        $\omega$ & Flow model hyperparameter: Obstacle sequence length \\  \cline{1-2} 
        $D$ & Flow model hyperparameter: DiT embedding size \\  \cline{1-2} 
    \end{tabular}
    \normalsize
\end{table}

% \subsection{Robot Dynamics}
% We assume that the robot motion is described by series of integrators.

\subsection{Trajectory Optimization}
\noindent We assume that the robots are modeled as double integrator systems, which are expressive enough to model quadrotors \cite{singletary2021comparative} and holonomic mobile robots. It is a good approximation for on-road vehicles under a broad-set of conditions \cite{dabestani2024joint}. However, our formulation trivially extends to higher-order integrator systems as well. We can leverage the differential flatness property to directly plan in the space of trajectories and control inputs can be extracted post-hoc from the trajectory derivative.

Given $n$ number of robots and a planning period $K+1$, let $\mathbf{p}_{{i|k}}$ represent the x-y-z coordinate of the $i^{th}$ robot at time step $k$. The joint trajectory optimization problem can therefore be written as the following quadratically constrained quadratic program:
\begin{subequations}
\begin{align}
    \min_{\mathbf{p}_{i|0:K}} \frac{1}{2}\sum_{i=1}^{n}\sum_{k=0}^{K} {\lVert \ddot{\mathbf{p}}_{i|k}\rVert ^{2}_{2}} \label{cost} \\ 
    \lVert \mathbf{M}_{r}^{-1} \left(\mathbf{p}_{i|k} - \mathbf{p}_{j|k}\right)\rVert ^{2}_{2} - \mathbf{1} \geq \mathbf{0},\hspace{5mm}\forall k \hspace{3mm}  \forall i, j   \label{inter_robot_con}\\
    \lVert \mathbf{M}_{o}^{-1} \left(\mathbf{p}_{i|k} - \mathbf{p}_{o, m|k}\right)\rVert ^{2}_{2} - \mathbf{1} \geq \mathbf{0},\hspace{5mm} \forall k \hspace{3mm}  \forall i, \forall m \label{collision_con} \\
    \mathbf{p}_{min} \leq \mathbf{p}_{i|k}\leq  \mathbf{p}_{max}, \forall i, k \label{workspace_con}\\
    \mathbf{p}_{i|0} = \mathbf{b}_{i, 0}, \mathbf{p}_{i|K} = \mathbf{b}_{i, K}, \forall i \label{boundary_con}
\end{align}
\end{subequations}

The cost function \eqref{cost} minimizes the sum of squared acceleration magnitudes at each time step. However, we can minimize higher-order derivatives like jerk, snap, etc., without affecting the problem structure. The quadratic inequality \eqref{inter_robot_con} represents the inter-robot collision avoidance constraints. We assume that each robot is modeled as an axis-aligned spheroid with radius $(\frac{a}{2}, \frac{a}{2}, \frac{b}{2})$. Hence, $\mathbf{M}_{r}$ is a $3\times 3$ diagonal matrix formed with $(a, a, b)$. The quadratic inequalities \eqref{collision_con} are the collision avoidance constraints between the $i^{th}$ robot and the $m^{th}$ spheroid obstacle, whose position at time step $k$ is given by $\mathbf{p}_{o, m|k}$. For static obstacles, the position is invariant with respect to time. $\mathbf{M}_o$ is a $3\times 3$ diagonal matrix formed with $(a_o, a_o, b_o)$ which captures the obstacle dimension inflated with the robot's dimension.  The affine constraints \eqref{workspace_con} present the workspace constraints on the robots' positions. Finally, the equality constraints \eqref{boundary_con} enforce the boundary conditions of the robots' trajectories.

\newtheorem{remark}{Remark}
\begin{remark}\label{acc_vel}
We intentionally don't include velocity and acceleration bounds in the optimization formulation to reduce the number of constraints. Instead, we follow \cite{park2020efficient} and use post-hoc scaling of the time-axis to keep the velocity and acceleration within limits. However, velocity and acceleration bounds can be incorporated without affecting the problem structure.
\end{remark}

% \textcolor{black}{Put a remark about velocity and acceleration bounds}

% workspace constraints that ensure that robots' trajectories lie within an ellipsoid centered at $\mathbf{p}_w$. The diagonal matrix $\mathbf{M}_{w}$ is formed with the radius of the workspace ellipsoids $(a_w, a_w, b_w)$. 

% where $\mathbf{M}_{a}$ and $\mathbf{M}_{w}$ are both diagonal matrices with the diagonals representing the radius of the ellipsoids that cover a robot and the entire workspace, respectively. $\mathbf{P}_{o}$ represents the center of the ellipsoidal workspace. 

\subsubsection*{Polynomial Parametrization}
\noindent We parametrize the positional trajectory of the $i^{th}$ robot $\mathbf{p}_{i|0:K}$ in the following manner:

\begin{equation}
\begin{aligned}
    \mathbf{p}_{i| 0: K} = \left[\begin{array}{ccc}
        \mathbf{W} & \mathbf{0} & \mathbf{0} \\
        \mathbf{0} & \mathbf{W} & \mathbf{0} \\
        \mathbf{0} & \mathbf{0} & \mathbf{W}
    \end{array}\right] \left[\begin{array}{c}
        \boldsymbol{\xi}_{i, x} \\ \boldsymbol{\xi}_{i, y} \\ \boldsymbol{\xi}_{i, z} 
    \end{array}\right] = \overline{\mathbf{W}} \boldsymbol{\xi}_{i},
\end{aligned}
\label{eq::poly}
\end{equation}

\noindent where $\mathbf{W}$ is a matrix formed with the time-dependent polynomial basis functions and the $\left[\boldsymbol{\xi}_{i,x}, \boldsymbol{\xi}_{i,y}, \boldsymbol{\xi}_{i,z}\right]$ is a vector of coefficients that define the trajectory. The velocities and accelerations can also be expressed in terms of the coefficients in a similar manner.

\textcolor{black}{
\begin{equation}
\begin{aligned}
    \dot{\mathbf{p}}_{i| 0: K} = \dot{\overline{\mathbf{W}}} \boldsymbol{\xi}_{i},\qquad \ddot{\mathbf{p}}_{i| 0: K} = \ddot{\overline{\mathbf{W}}} \boldsymbol{\xi}_{i}
\end{aligned}
\label{eq::poly}
\end{equation}}

\noindent \textcolor{black}{through matrices $\dot{\overline{\mathbf{W}}}$, $\ddot{\overline{\mathbf{W}}}$ which contain the derivative of the polynomial basis. Since, we assume a double integrator system, the states of the robots are $(\mathbf{p}_k, \dot{\mathbf{p}}_k)$ and the control is $\ddot{\mathbf{p}}_k$, all of which can be expressed in terms of the polynomial coefficients $\boldsymbol{\xi}_{i}$.}

% \begin{equation}
% \begin{aligned}
%     \dot{\mathbf{P}}_{i| 0: K} = \dot{\overline{\mathbf{W}}} \mathbf{c}_{i} \hspace{10mm} \ddot{\mathbf{P}}_{i| 0: K} = \ddot{\overline{\mathbf{W}}} \mathbf{c}_{i}
% \end{aligned}
% \label{eq::poly2}
% \end{equation}

We roll the coefficients of all the robots together into a single vector $\boldsymbol{\xi} = (\boldsymbol{\xi}_{1, x}, \dots, \boldsymbol{\xi}_{n, x}, \boldsymbol{\xi}_{1, y}, \dots, \boldsymbol{\xi}_{n, y}, \boldsymbol{\xi}_{1, z}, \dots, \boldsymbol{\xi}_{n, z})$. Consequently, using \eqref{eq::poly}, we can rewrite the optimization problem into the following more compact form:
\begin{equation}
\begin{aligned}
    \underset{\boldsymbol{\xi}}{\text{minimize}} &\hspace{5mm} \frac{1}{2}\boldsymbol{\xi}^{\top} \mathbf{Q}\, \boldsymbol{\xi} + \mathbf{q}^{\top} \boldsymbol{\xi} \\ 
    \text{subject to} &\hspace{5mm} \mathbf{A}\, \boldsymbol{\xi} = \mathbf{b}   \\
                        &\hspace{5mm} \mathbf{G}\, \boldsymbol{\xi} \leq  \mathbf{h}\\
                      &\hspace{5mm} \mathbf{g}\left(\boldsymbol{\xi}\right) \leq \mathbf{0}  
\end{aligned}
\label{eq::form}
\end{equation}

\noindent where the matrices $\mathbf{Q}, \mathbf{A}, \mathbf{G}$ and vectors $\mathbf{q}, \mathbf{b}, \mathbf{h}$ are constants. The affine equality constraints stem from the boundary constraints \eqref{boundary_con}. The affine inequality models the workspace constraints \eqref{workspace_con}. The function $\mathbf{g}$ contains the inequalities  \eqref{inter_robot_con}-\eqref{collision_con}, expressed in terms of polynomial coefficients.

\begin{figure*}
    \centering
    \includegraphics[scale=0.42]{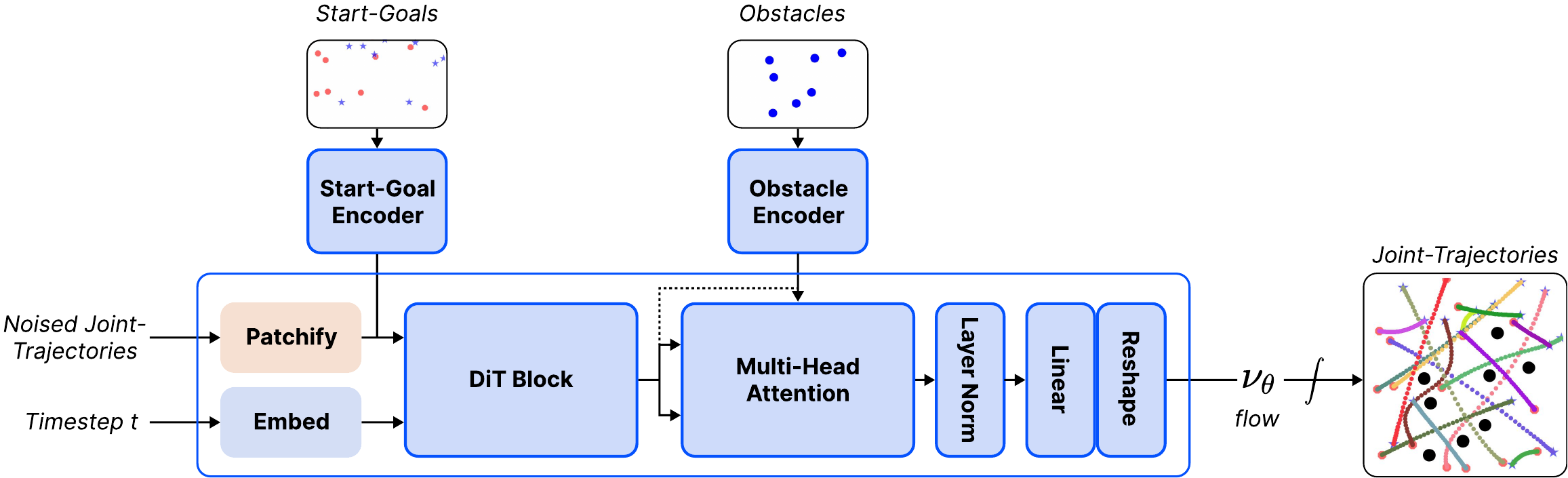}
    % \includesvg[scale=0.51]{figs/flow_model.svg}
    \caption{\footnotesize{Architecture of our Flow Matching Network. Start-goal and obstacle encoders are based on PointNet++ to ensure invariance to the shuffling of obstacle coordinates and start and goal pairs.  }}
    \label{fig:flow_model}
\end{figure*}

%%%%%%%%%%%%%%%%%%%%%%%%%%%%%%%%%%%%%%%%%%%%%%%%%%%%%%%%%%%%%%%%%%%%%%%%%%%%%%%%%%%%%%%%%%%%%%%%%%%%%
\section{Method}\label{sec:method}
We present Flow-Opt, a framework that combines the generative capabilities of a flow-matching policy with a learned differentiable trajectory optimizer to enable scalable centralized multi-robot trajectory optimization. An overview of our pipeline is shown in Fig.\ref{fig:pipeline}. The flow policy takes as input pairs of start and goal positions and learns a distribution of potential trajectory candidates. Trajectories sampled from the learned flow policy may not strictly satisfy collision-avoidance constraints arising from obstacles in the environment or the inter-robot interactions. To enforce feasibility, they are refined through a trajectory optimizer, referred to in this work as the SF. A key novelty of our SF is that it incorporates a neural initialization network conditioned on the flow-generated trajectories to produce good initializations for the underlying optimization solver. This initialization network is trained in a purely self-supervised manner. 

In the following subsections, we describe each component of the pipeline illustrated in Fig.\ref{fig:pipeline} in detail.

\begin{figure*}
    \centering
    \includegraphics[scale=0.2]{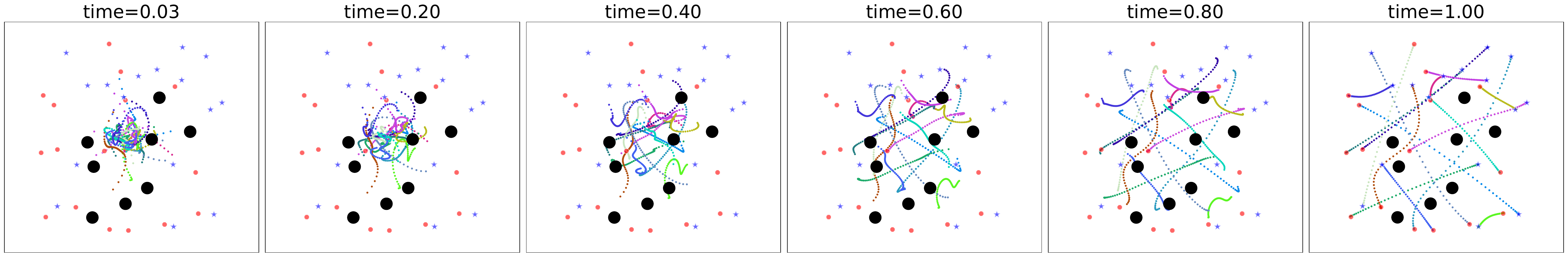}
    \caption{\footnotesize{Process of denoising random noise to feasible multi-robot trajectories through a flow policy.}}
    \label{flow_denoising}
\end{figure*}

% \subsection{Learning Flow Based Generative Prior}
\subsection{Context-Aware Conditional Flow Matching for Trajectories}

\noindent Our objective is to learn a generative model capable of sampling trajectory coefficients $\boldsymbol{\xi}$ from the conditional distribution $q(\boldsymbol{\xi} | \mathbf{c})$, where $\mathbf{c}$ denotes the problem context (e.g., start-goal pairs and environment information). Each joint-trajectory coefficient $\boldsymbol{\xi}$ is treated as a single point in a high-dimensional space. We adopt a conditional flow matching framework that constructs a continuous-time flow transforming samples from a simple, tractable base distribution $q_0(\boldsymbol{\xi})$ into samples from the target conditional distribution $q(\boldsymbol{\xi} | \mathbf{c})$.

\subsubsection{The Conditional Probability Path}

We define a time-dependent conditional probability path $q_t(\boldsymbol{\xi}| \mathbf{c})$ for $t \in [0, 1]$ which smoothly connects the base distribution $q_0(\boldsymbol{\xi})$ (which is independent of $\mathbf{c}$) to the target distribution $q_1(\boldsymbol{\xi} | \mathbf{c}) = q(\boldsymbol{\xi} | \mathbf{c})$. This path is generated by a conditional vector field $u_t(\boldsymbol{\xi} | \mathbf{c})$. The evolution of samples along this path is described by the following ordinary differential equation (ODE):
\begin{equation}
    \frac{d\boldsymbol{\xi}_t}{dt} = u_t(\boldsymbol{\xi}_t | \mathbf{c}) \quad \text{where} \quad \boldsymbol{\xi}_0 \sim q_0(\boldsymbol{\xi}).
    \label{eq:ode_def}
\end{equation}
By integrating this ODE from $t=0$ to $t=1$, the resulting point $\boldsymbol{\xi}_1$ will be a sample from the target distribution $q(\boldsymbol{\xi} | \mathbf{c})$. Note that the time variable of ODE \eqref{eq:ode_def} is different from the planning horizon of the robot trajectories. The goal of flow matching is to train a neural network $v_\theta(\boldsymbol{\xi}_t, t, \mathbf{c})$, parameterized by weights $\boldsymbol{\theta}$, to approximate the unknown ground-truth vector field $u_t(\boldsymbol{\xi} | \mathbf{c})$.

\subsubsection{The Conditional Flow Matching Objective}

Directly simulating the ODE in \ref{eq:ode_def} is intractable since the true vector field $u_t$ is unknown. Flow Matching instead provides a direct regression objective. Given a sample trajectory $\boldsymbol{\xi}_1$ from our dataset $q(\boldsymbol{\xi} | \mathbf{c})$ and a sample $\boldsymbol{\xi}_0$ from the prior $q_0(\boldsymbol{\xi})$, we can construct a path between them. A common and effective choice is a simple linear interpolation:
\begin{equation}
    \boldsymbol{\xi}_t = (1 - t)\boldsymbol{\xi}_0 + t\boldsymbol{\xi}_1.
    \label{eq:flow_interp}
\end{equation}
The "ground truth" conditional vector field that generates this specific straight-line path is simply the difference between the endpoint and the starting point:
\begin{equation}
    u_t(\boldsymbol{\xi}_t | \mathbf{c}) = \boldsymbol{\xi}_1 - \boldsymbol{\xi}_0.
\end{equation}
This provides a direct regression target for our neural network $v_\theta$. The Conditional Flow Matching (CFM) objective is to minimize the expected squared error between the network's prediction and this ground-truth vector field over all possible times, contexts, initial points, and target trajectories:
\begin{dmath}
    \mathcal{L}_{\text{CFM}}(\boldsymbol{\theta}) = \mathbb{E}_{t \sim \mathcal{U}[0,1], \boldsymbol{\xi}_0 \sim q_0(\boldsymbol{\xi}), (\boldsymbol{\xi}_1, \mathbf{c}) \sim q(\boldsymbol{\xi}, \mathbf{c})} \left[ \left\| v_\theta((1-t)\boldsymbol{\xi}_0 + t\boldsymbol{\xi}_1, t, \mathbf{c}) - (\boldsymbol{\xi}_1 - \boldsymbol{\xi}_0) \right\|^2 \right].
    \label{eq:cfm_loss}
\end{dmath}
Here, $q(\boldsymbol{\xi}, \mathbf{c})$ represents the joint distribution of optimal trajectories and their corresponding contexts from our dataset, and $\mathcal{U}(0,1)$ is the uniform distribution over the time interval. The neural network $v_\theta$ is trained to predict the direction $(\boldsymbol{\xi}_1 - \boldsymbol{\xi}_0)$ given the interpolated point $\boldsymbol{\xi}_t$, the time $t$, and the context $\mathbf{c}$.

\subsubsection{Generating Novel Trajectories (Inference)}
After training the neural network $v_\theta$ by minimizing the objective $\mathcal{L}_{\text{CFM}}$, it approximates the true vector field that maps samples from the prior distribution $q_0(\boldsymbol{\xi})$ to the conditional data distribution $q(\boldsymbol{\xi} | \mathbf{c})$. To generate a trajectory for a new context $\mathbf{c}_{\text{new}}$, we first sample an initial point from the prior distribution:
\[
    \boldsymbol{\xi}_0 \sim q_0(\boldsymbol{\xi}).
\]
We then solve the following initial value problem from $t=0$ to $t=1$ using a numerical ODE solver (e.g., Euler or Runge-Kutta methods):
\begin{equation}
    \frac{d\boldsymbol{\xi}_t}{dt} = v_\theta(\boldsymbol{\xi}_t, t, \mathbf{c}_{\text{new}}).
    \label{eq:inference_ode}
\end{equation}
The solution at $t=1$, denoted as $\boldsymbol{\xi}_1 $, is a novel trajectory coefficient sample ${\boldsymbol{\xi}}$ that is conditioned on the provided context $\mathbf{c}_{\text{new}}$:
\begin{equation}
    \boldsymbol{\xi} = \boldsymbol{\xi}_1 = \boldsymbol{\xi}_0 + \int_0^1 v_\theta(\boldsymbol{\xi}_t, t, \mathbf{c}_{\text{new}}) \,dt.
    \label{integral_flow}
\end{equation}

By sampling different initial points $\boldsymbol{\xi}_0$ while keeping $\mathbf{c}_{\text{new}}$ fixed, the model produces a diverse set of trajectories consistent with the specified context. Furthermore, we can generate all these diverse trajectories by evaluating \eqref{integral_flow} in a batch (parallelized) fashion. A typical flow policy trajectory generation is shown in Fig.\ref{flow_denoising}.

\subsubsection{Flow Network Architecture}\label{flow_network} 

Our flow network processes noisy joint trajectories represented as $\boldsymbol{\xi}_{t} \in \mathbb{R}^{n_{\xi} \times n \times n_d}$, where $n_{\xi}$ denotes the order of polynomial basis functions used to represent the trajectories, $n$ is the number of robots, $n_d$ represents the dimension of the workspace (2D or 3D), while $t \in [0, 1]$ indicates the flow timestep. These trajectories are derived through the noise perturbation process described in Eq. \eqref{eq:flow_interp}. The model employs a Diffusion Transformer (DiT) as a backbone.

% We employ a CNN based encoder that takes in $\boldsymbol{\xi}_{t}$ and performs a patchification \cite{Peebles2022ScalableDM} to convert it into a sequence $\boldsymbol{\xi}_{t}^{'} \in \mathbb{R}^{S \times D}$ of length $S$ and dimension $D$. 

\textcolor{black}{
The patchification layer is used to tokenize $\xi_t$ for downstream attention blocks. It consists of a single CNN with a kernel size and stride $(1, n_d)$, taking $n_{\xi}$ input channels and producing $D$ output channels. This operation aggregates all state dimensions of a single robot into one embedding while explicitly avoiding any mixing across robots. Simultaneously, it maps the trajectory coefficient dimension $n_{\xi}$ to the model embedding dimension $D$, producing one token per robot. The output of the convolution is then reshaped and collapsed into $\xi_{t}^{'} \in \mathbb{R}^{S \times D}$. These tokens are then passed to the DiT backbone, where inter-robot interactions are modeled via self-attention layers rather than through the convolution operator. The same structure is used for the initialization network.
}
$S$ and dimension $D$ are tunable model hyperparameters that control the sequence length and the embedding dimension, respectively. We add sinusoidal positional embeddings to $\boldsymbol{\xi}_{t}^{'}$ to preserve spatial-temporal relationships. The scalar timestep $t$ is encoded into a $D$-dimensional vector $\mathbf{t}^{'} \in \mathbb{R}^{D}$ using the same sinusoidal embedding scheme, enabling the model to condition on the flow matching process timing.

% Contextual information is incorporated through two separate PointNet-based networks \cite{Qi2016PointNetDL}. One pointnet-based model for start-goal condition $\mathbf{c}_{sg} \in \mathbb{R}^{n \times (2 \cdot d)}$ representing start and goal positions for each $n$ robot. The second pointnet-based model is for the obstacle condition $\mathbf{c}_{ob} \in \mathbb{R}^{n_{obs} \times (2 \cdot d)}$ representing the obstacle position and velocity and $n_{obs}$ denotes the number of obstacles. They both form the conditioning inputs $\mathbf{c} = [\mathbf{c}_{sg}, \mathbf{c}_{ob}]$ for the flow-matching trajectory prediction network. The two condition networks each produce compact feature representations $\mathbf{c}_{sg}^{'} \in \mathbb{R}^{S \times D}$ and $\mathbf{c}_{ob}^{'} \in \mathbb{R}^{q \times D}$ respectively, where $q$ is a tunable parameter for obstacle features. 

% Contextual information is incorporated through two permutation-invariant PointNet-based networks \cite{Qi2016PointNetDL}. One pointnet-based model for start-goal condition $\mathbf{c}_{sg} \in \mathbb{R}^{n \times (2 \cdot n_d)}$ representing start and goal positions for each $n$ robot. The second pointnet-based model is for the obstacle condition $\mathbf{c}_{ob} \in \mathbb{R}^{n_{obs} \times (2 \cdot n_d)}$ representing the obstacle position and velocity, and $n_{obs}$ denotes the number of obstacles. 

Contextual information is incorporated using two CNN-based networks. The first network encodes the start-goal condition, 
$\mathbf{c}_{\mathrm{sg}} \in \mathbb{R}^{n \times (2 n_d)}$, representing the start and goal positions for each of the $n$ robots. 
The second network is a permutation-invariant PointNet-based model~\cite{Qi2016PointNetDL} that encodes the obstacle condition, 
$\mathbf{c}_{\mathrm{ob}} \in \mathbb{R}^{n_{\mathrm{obs}} \times (2 n_d)}$, representing the positions and velocities of the $n_{\mathrm{obs}}$ obstacles. They together form the conditioning inputs $\mathbf{c} = [\mathbf{c}_{sg}, \mathbf{c}_{ob}]$ for the flow-matching trajectory prediction network. The PointNet based model has $2 \cdot n_d$ input channels and employs 1D convolutions with a kernel size of 1 within its layers. The two condition networks each produce compact feature representations $\mathbf{c}_{sg}^{'} \in \mathbb{R}^{S \times D}$ and $\mathbf{c}_{ob}^{'} \in \mathbb{R}^{\omega \times D}$ respectively, where $\omega$ is a tunable parameter for obstacle features. 

The core DiT block processes the trajectory embedding $\boldsymbol{\xi}_{t}^{'}$ conditioned on both the timestep encoding $\mathbf{t}^{'}$ and start-goal features $\mathbf{c}_{sg}^{'}$, producing an intermediate representation $\boldsymbol{\xi}^{dit}_t \in \mathbb{R}^{S \times D}$. This output then undergoes either self-attention (for obstacle-free cases) or cross-attention with obstacle features $\mathbf{c}_{ob}^{'}$ (when obstacles exist). Finally, a feed-forward network with linear and normalization layers transforms and reshapes these features to generate the output trajectories $v_{\theta}(\boldsymbol{\xi}_t, t, \mathbf{c}) \in \mathbb{R}^{n_{\xi} \times n \times n_d}$ as defined in Eq. \eqref{eq:inference_ode}.

\subsection{Fast Inference-Time Refinement}
\noindent The flow policy described in the previous subsection is trained purely on demonstrations of optimal trajectories. Therefore, it is unaware of the underlying trajectory level constraints. As a result, the predicted trajectories may not be completely feasible. We enforce constraint satisfaction by modifying the flow predicted trajectories through the following optimization problem.

\begin{subequations}
\begin{align}
    \underset{\overline{\boldsymbol{\xi}}}{\text{minimize}} &\hspace{5mm} \frac{1}{2}\left \Vert \overline{\boldsymbol{\xi}}-\boldsymbol{\xi}\right \Vert_2^2 \label{proj_cost} \\ 
    \text{subject to} &\hspace{5mm} \mathbf{A}\, \overline{\boldsymbol{\xi}} = \mathbf{b}  \label{proj_eq} \\
                        &\hspace{5mm} \mathbf{G}\, \overline{\boldsymbol{\xi}} \leq  \mathbf{h} \label{proj_ineq}\\
                      &\hspace{5mm} \mathbf{g}\left(\overline{\boldsymbol{\xi}}\right) \leq \mathbf{0}\label{proj_robot}  
\end{align}
\end{subequations}

Equations \eqref{proj_cost}-\eqref{proj_robot} define a typical projection problem that computes the closest possible trajectory coefficient to the flow prediction $\boldsymbol{\xi}$ that satisfies the constraints. Inference-time refinements are commonly used for diffusion and flow based trajectory predictions \cite{edmp}. Several works formulate this step as a projection optimization problem similar to \eqref{proj_cost}-\eqref{proj_robot} \cite{flow_proj}, \cite{smd}. However, existing works often underemphasize the computational cost of inference-time refinements. For example, these are often several times higher than the computation cost associated with the inference process of diffusion or flow policies \cite{edmp}, \cite{smd}. To address this issue, we adopt a learning-based approach to improve the computational efficiency of inference time refinements. Specifically, we learn a warm-start for \eqref{proj_cost}-\eqref{proj_robot} conditioned on the flow predicted input $\overline{\boldsymbol{\xi}}$. There are two building blocks of our approach. First, we reformulate the underlying quadratic inequalities in constraint function $\mathbf{g}$ (recall \eqref{inter_robot_con}-\eqref{collision_con}) to express the solution process of \eqref{proj_cost}-\eqref{proj_robot} in the form of differentiable fixed-point operations. Second, we leverage the differentiability to develop a pipeline that performs end-to-end self-supervised learning to accelerate the convergence of fixed-point operations. 

\begin{figure*}
    \centering
    \includegraphics[scale=0.39]{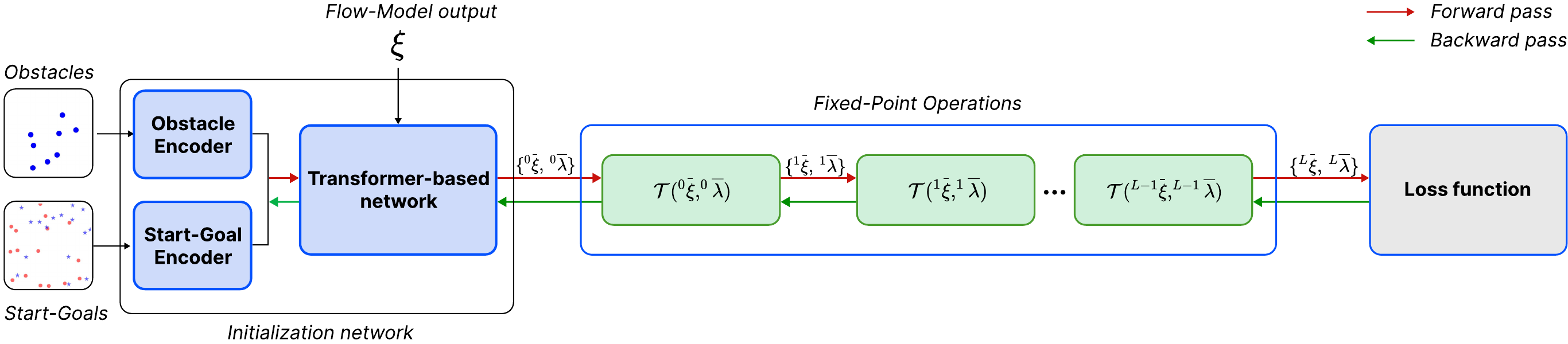}
    % \includesvg[scale=0.5]{figs/unrolling_revised_black.svg}
    \caption{\footnotesize{The training pipeline for learning warm-start for the SF solver. The architecture has two components: a learnable part consisting of a start-goal and obstacle encoder and transformer-based network, and a fixed part that resembles $L$ fixed-point iterations of our SF solver. The loss function is simply made of a fixed-point residual. During training, the gradients flow through the fixed-point layer, ensuring the initialization $({^0}\overline{\boldsymbol{\xi}}, {^0}\overline{\boldsymbol{\lambda}} )$  produced by the networks is used by the downstream SF solver. This in turn requires that the fixed-point iterations are differentiable.}}.
    \label{unroll_learning}
\end{figure*}

\subsubsection{Fixed-Point Representation} We propose a custom SF solver that reduces the solution process of \eqref{proj_cost}-\eqref{proj_robot} into a fixed-point operation $\mathcal{T}$ of the following form, where the left superscript $l$ denotes the iteration number.

\begin{align}
    {^{l+1}}\bar{\boldsymbol{\xi}}, {^{l+1}}\bar{\boldsymbol{\lambda}} = \mathcal{T}({^{l}}\bar{\boldsymbol{\xi}}, {^{l}}\bar{\boldsymbol{\lambda}})
    \label{fixed_point}
\end{align}

\noindent We derive the mathematical structure of $\mathcal{T}$ in the Appendix \ref{Appendix}, where we also define the Lagrange multiplier $\boldsymbol{\lambda}$. But some interesting points are worth pointing out immediately.
\textcolor{black}{
\begin{itemize}
    \item As illustrated in Fig.~\ref{unroll_learning}, the SF solver can be unrolled as a chain of $\mathcal{T}$ operators stacked $L$ times. By ensuring that every computation within $\mathcal{T}$ is differentiable, we enable backpropagation through the entire SF solver. This allows us to compute the gradients of the output $({^L}\bar{\boldsymbol{\xi}}, {^L}\bar{\boldsymbol{\lambda}})$ with respect to the initialization $({^0}\bar{\boldsymbol{\xi}}, {^0}\bar{\boldsymbol{\lambda}})$. This differentiability is central to the self-supervised learning framework discussed later in this section.
    \item In Appendix~\ref{Appendix}, we demonstrate how the quadratic inequalities \eqref{inter_robot_con}-\eqref{collision_con} can be reformulated to ensure that $\mathcal{T}$ relies solely on matrix-matrix products. This formulation is not only differentiable but also facilitates efficient batching and acceleration on GPUs.
\end{itemize}
}

% First, by carefully reformulating the quadratic inequalities \eqref{inter_robot_con}-\eqref{collision_con}, we can ensure that the numerical computations underlying $\mathcal{T}$ only require matrix-matrix products, which can be easily batched and accelerated over GPUs. Secondly, $\mathcal{T}$ involves only differentiable operations, which allows us to compute how changes in the initialization of $\mathcal{T}$ affect the convergence process. This differentiability forms the core of our learning pipeline, designed to accelerate the convergence of fixed-point iteration \eqref{fixed_point}. 

\subsubsection{{Learned Initialization for Fixed-Point Iteration}} Our objective is to learn good initializations that accelerate the convergence of the fixed-point iteration \eqref{fixed_point}. A standard metric for convergence is  the fixed-point residual defined as \cite{sambharya2024learning}

\begin{align}
    \mathcal{L}_{FP} = \left \Vert \begin{bmatrix}
        {^{l+1}}\bar{\boldsymbol{\xi}}\\
        {^{l+1}}\bar{\boldsymbol{\lambda}}
    \end{bmatrix}-\mathcal{T}({^{l}}\overline{\boldsymbol{\xi}}, {^{l}}\bar{\boldsymbol{\lambda})} \right\Vert_2^2
    \label{fixed_point_residual}
\end{align}

A good initialization would ensure that the residual \eqref{fixed_point_residual} quickly converges to zero. To this end, we propose the learning pipeline shown in Fig.\ref{unroll_learning}. It consists of a learnable module followed by an unrolled chain of length $L$ of fixed-point iterations. The learnable part consists of encoders for start and goal pairs and environment obstacles. The embeddings from these encoders are paired with the output from the pre-trained flow model $\boldsymbol{\xi}$ and fed to a transformer that produces the warm-start $({^0}\overline{\boldsymbol{\xi}}, {^0}\boldsymbol{\lambda})$ for the fixed-point iterations. Let $^{L}{\overline{\boldsymbol{\xi}}}, ^{L}\boldsymbol{\lambda}$ be the solution obtained by running the fixed-point iteration for $L$ iterations from the $({^0}\overline{\boldsymbol{\xi}}, {^0}\boldsymbol{\lambda})$. We formulate the following optimization problem to train the learnable part of the SF. 

\begin{align}
    \min_{\boldsymbol{\phi}} \sum_{l=0}^{L-1} \mathcal{L}_{FP}+\left\Vert {^L}\overline{{\boldsymbol{\xi}}}-\boldsymbol{\xi}\right\Vert_2^2,
    \label{NN_loss}
\end{align}

\noindent where $\boldsymbol{\phi}$ contains the weights of the start-goal and obstacle encoders and the transformer network. The first term in \eqref{NN_loss} minimizes the fixed-point residual at each iteration and is responsible for accelerating the convergence of \eqref{fixed_point}  \cite{sambharya2024learning}. The second term ensures that the learned initialization leads to a solution that is minimally displaced from the original flow-predicted trajectory coefficient. During training, the gradient of the loss function is traced through the stacked layers of $\mathcal{T}$ to the learnable parts. This ensures that the neural network layers are aware of how its predictions are leveraged by the downstream solver and leads to highly effective warm-start for the fixed-point solver. It is worth pointing out that the loss function \eqref{NN_loss} does not require the ground-truth solution of the fixed-point iteration \eqref{fixed_point}. In that sense, our learning process is self-supervised and is guided by the fixed-point residual itself.

\subsubsection{Architecture of Initialization Network}

The initialization model consists of two components: two CNN-based models and a lightweight transformer model. The CNN encoders, architecturally identical to those in the flow network, generate condition embeddings: $\mathbf{c}_{sg}^f \in \mathbb{R}^{S \times 2 \cdot n_d}$ from start-goal states and $\mathbf{c}_{ob}^f \in \mathbb{R}^{\omega \times 2 \cdot n_d}$ from obstacles, where $S, \omega$ are tunable and $n_d$ denotes the workspace dimension (2D/3D). The transformer module refines the input flow-generated trajectories $\boldsymbol{\xi}$ conditioned on $\mathbf{c}_{sg}^f$ and $\mathbf{c}_{ob}^f$. 

The trajectories are first patchified via CNN model into $\boldsymbol{\xi}^{'} \in \mathbb{R}^{S \times D}$ with positional embeddings added. This CNN model used for the patchification is also similar to the one used in the flow model. Within the transformer, $\boldsymbol{\xi}^{'}$ is processed with $\mathbf{c}_{sg}^f$ via self-attention, and optionally with $\mathbf{c}_{ob}^f$ through cross-attention when obstacles are present. The output provides warm-start values: initial Lagrangian multipliers $^0\overline{\boldsymbol{\lambda}}$ and near-feasible trajectories $^0\overline{\boldsymbol{\xi}} \in \mathbb{R}^{n_{\xi} \times n \times n_d}$ (where $n_{\xi}$ is the polynomial basis order) for the fixed-point operation $\mathcal{T}$ defined in Eq. \eqref{fixed_point}.

\section{Validation and Benchmarking}\label{val}
The objectives of this section are threefold. 
\begin{itemize}
    \item Demonstrate that our approach can produce smooth multi-robot trajectories in a scalable manner. 
    \item Show improvement in trajectory quality, success rate, and computation time compared to existing model-based and data-driven methods.
    \item Analyze the robustness of our approach to generalize to both in-distribution and out-of-distribution test cases.
\end{itemize}

\subsection{Implementation Details} 
\noindent The flow network Fig.\ref{fig:flow_model}, initialization network Fig.\ref{unroll_learning} and the SF solver were all implemented in JAX \cite{jax} using Equinox \cite{kidger2021equinox} as the neural training library. Network parameters are detailed in the Appendix (Section \ref{Appendix}, Table \ref{tab:flow_model_parameters}, \ref{tab:init_model_parameters}). \textcolor{black}{All benchmarking experiments were conducted on an RTX 5090 desktop with an Intel i9 processor and 32GB of RAM.}

\subsubsection{Data Collection}The flow policy was trained on a dataset of over $20,000$ multi-robot trajectories, with start and goal configurations sampled within a rectangular workspace centered at the origin. To generate this data, we extended the method proposed in \cite{Rastgar2020GPUAC} to support affine workspace constraints. Furthermore, we developed a batched implementation of the solver, enabling parallel processing of multiple problem instances. This optimization significantly scaled up data collection, allowing the full dataset to be generated in approximately 10 minutes on an NVIDIA RTX 5090. During testing, new start and goal configurations are sampled from the same distribution as the training data.
\textcolor{revision_color}{Empirical evidence confirming that the test set is in-distribution yet contains no near-duplicates of training instances is provided in Appendix~\ref{appendix:train_test_novelty}.}

\textcolor{black}{
\subsubsection{Metrics} We compare our approach with different baselines using the following three metrics}\label{metrics}

\textcolor{black}{
\begin{itemize}
\item \textbf{Success Rate (SR) and Computational Efficiency:} We measure the robustness of each planner by calculating the Success Rate—the percentage of trials where a collision-free solution is found. Additionally, we analyze computational efficiency by comparing the average wall-clock time for successful queries.
\item \textbf{Trajectory Quality and Optimality:} To assess physical feasibility and efficiency of the motions, we compute:
\begin{enumerate}
\item \textbf{Path Length:} The Euclidean arc-length of the generated trajectory.
\item \textbf{Smoothness Cost:} The sum of the squared acceleration norm ($\sum_k ||\ddot{\mathbf{p}}_{i, k}||^2$), averaged across all the robots. This metric serves as a proxy for trajectory smoothness
\end{enumerate}
All quality metrics are averaged across successful runs over multiple random seeds and environment configurations.
\end{itemize}
}
\begin{figure*}
    \centering
    \includegraphics[scale=0.37]{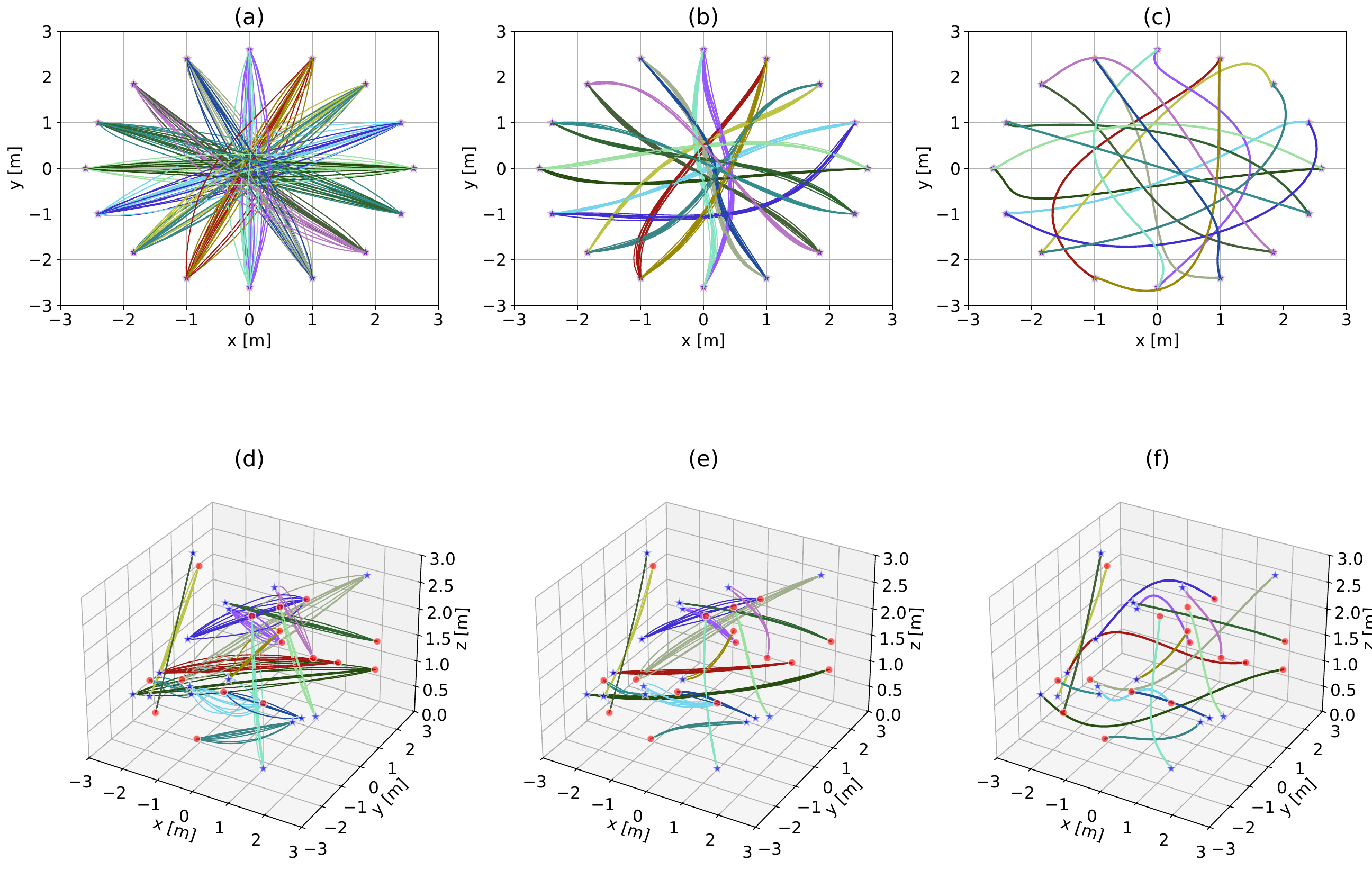}
    \caption{\footnotesize{The three stages of our trajectory planning pipeline. We first sample a large number of trajectories ($\approx 256$) from the trained flow policy (Fig.(a), (d)). We then sort the sampled trajectories based on constraint satisfaction and choose the top 10 with the lowest residual (Fig.(b)-(e)). Finally, these trajectories are refined through SF, and the feasible trajectory with the lowest smoothness cost is output as the optimal solution (Fig.(c)-(f)).}}
    \label{qual_result_16}
\end{figure*}

\begin{figure}[!h]
    \centering
    \includegraphics[scale=0.28]{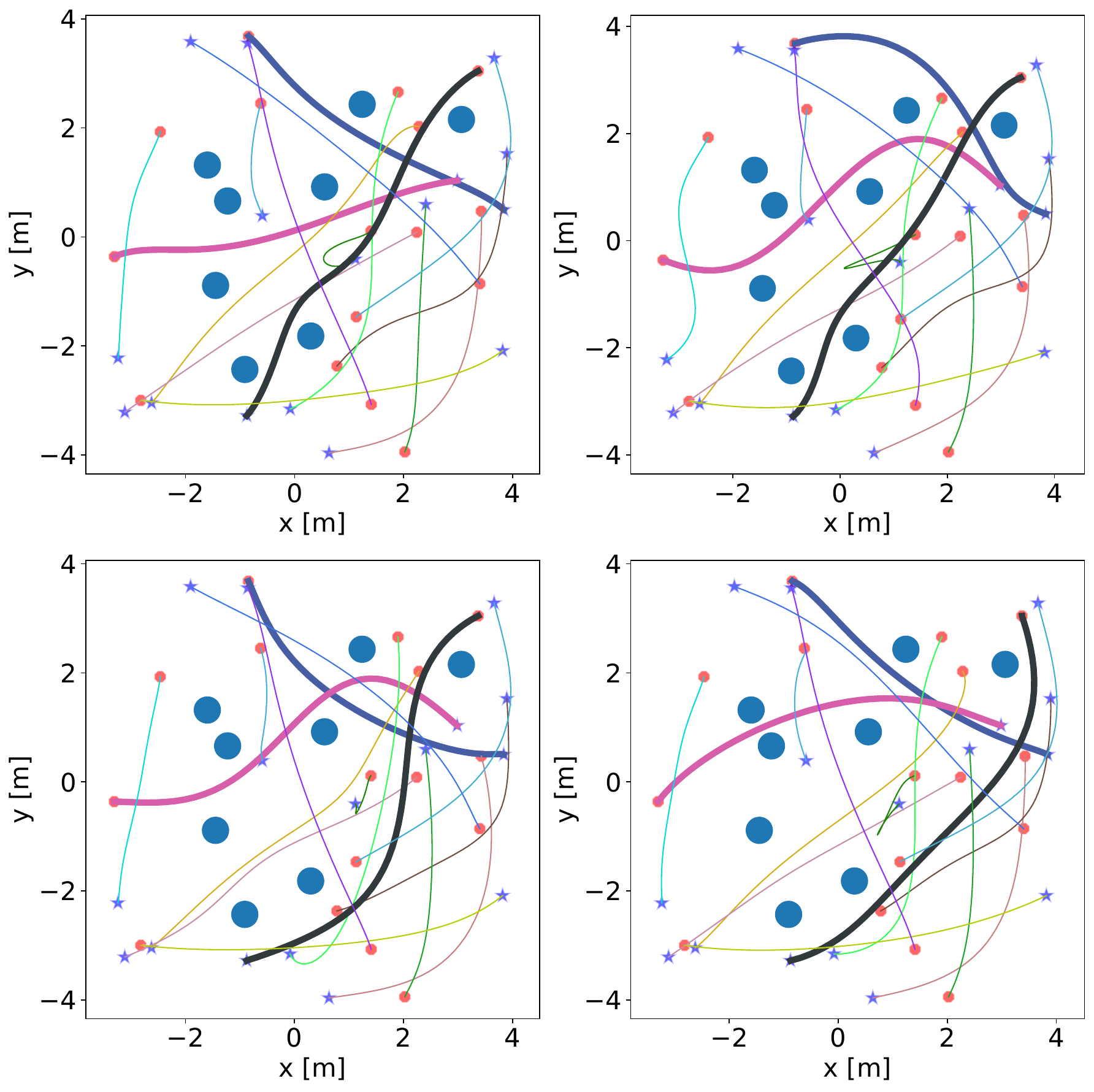}
    \caption{\footnotesize{Demonstration of diverse collision avoidance behaviors for 16 robots in obstacle-filled environments obtained with our approach. As can be seen, the same robot can choose different strategies to avoid static obstacles and accordingly also its strategy to avoid other robots. Some of the trajectories showing pronounced diversity are highlighted.} }
    \label{fig:diversity}
\end{figure}

\begin{figure}[!h]
    \centering
    \includegraphics[width=8.5cm]{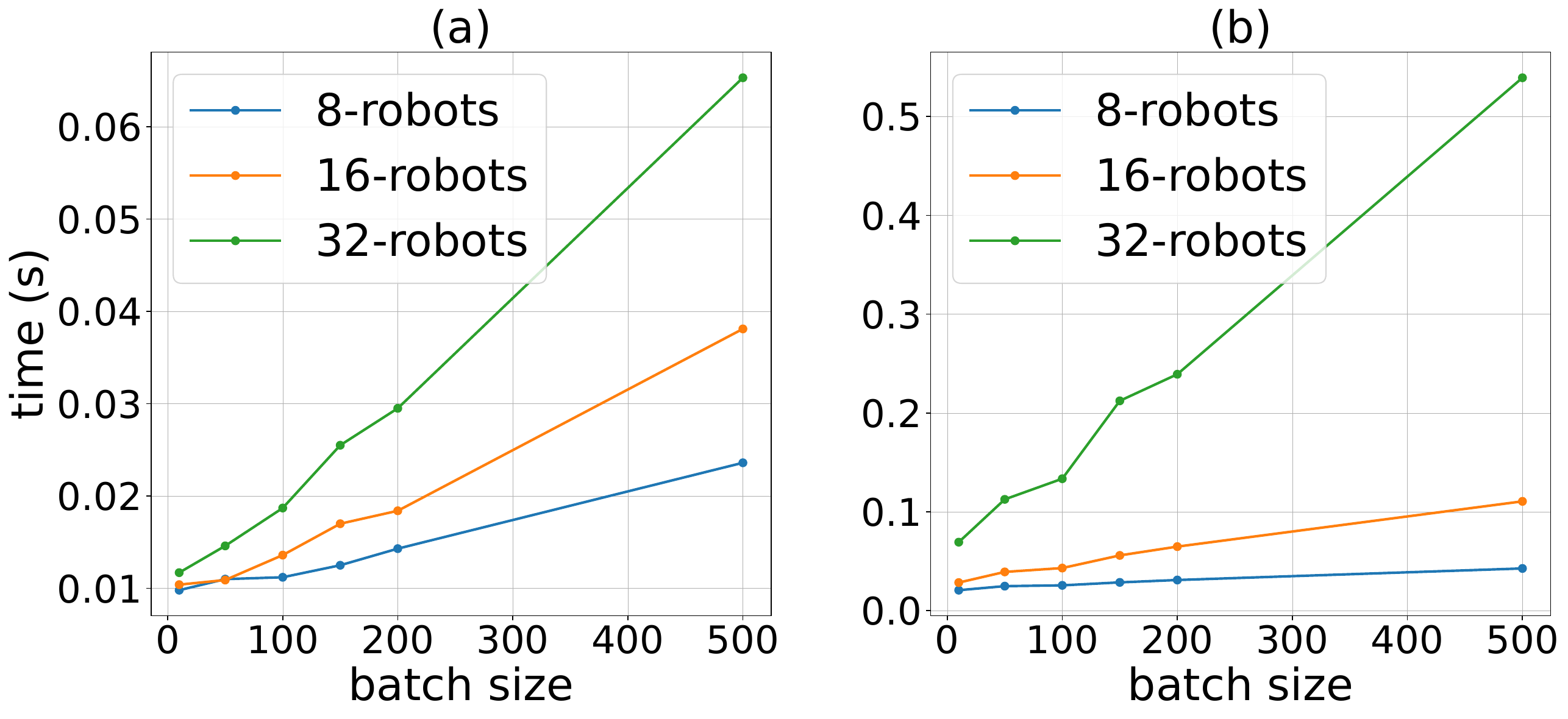}
    \caption{\footnotesize{Fig.(a) shows the computation time required to sample a batch of trajectories from the flow model for different number of robots. Fig.(b) shows the computation time to perform 500 iterations of SF to refine a batch of trajectories. As can be seen, due to GPU acceleration/parallelization, both timings scale almost linearly with batch size.}}
    \label{fig_scalability}
\end{figure}

% \begin{figure}
%     \centering
%     \includegraphics[width=5cm]{figs/tracking_error_box.pdf}
%     \caption{\footnotesize{\textcolor{blue}{Tracking error using a simple PID controller in PyBullet simulator}}  }
%     \label{tracking_error}
% \end{figure}

\subsection{Qualitative Validation }
\noindent In this subsection, we validate the individual components of our pipeline Fig.\ref{fig:pipeline} using two simple scenarios involving 16 robots (Fig.\ref{qual_result_16}). As described earlier, trajectories are first sampled from the trained flow policy and subsequently refined using the SF solver. We break down the sampling process into two parts. We first sample a large number of trajectories from the flow policy (Fig.\ref{qual_result_16}(a),(d)). We then select the top 10 trajectories that show the least constraint violation (Fig.\ref{qual_result_16} (b), (e)) and these are then passed through the SF. The trajectories obtained from the SF are ranked based on the constraint residual (see \eqref{primal_residual}) and smoothness (acceleration norm) and then the best trajectory is returned as the optimal solution (Fig.\ref{qual_result_16} (c), (f)). It is worth pointing out that our flow policy was trained on trajectories between randomly sampled start and goal pairs. Yet, it could generalize to the unique case where the robots are placed on the perimeter of a circle and have to move to their antipodal position.

\begin{figure*}[t]
    \centering
    \subfloat[$t=0.00$]{%
        \includegraphics[width=0.19\textwidth]{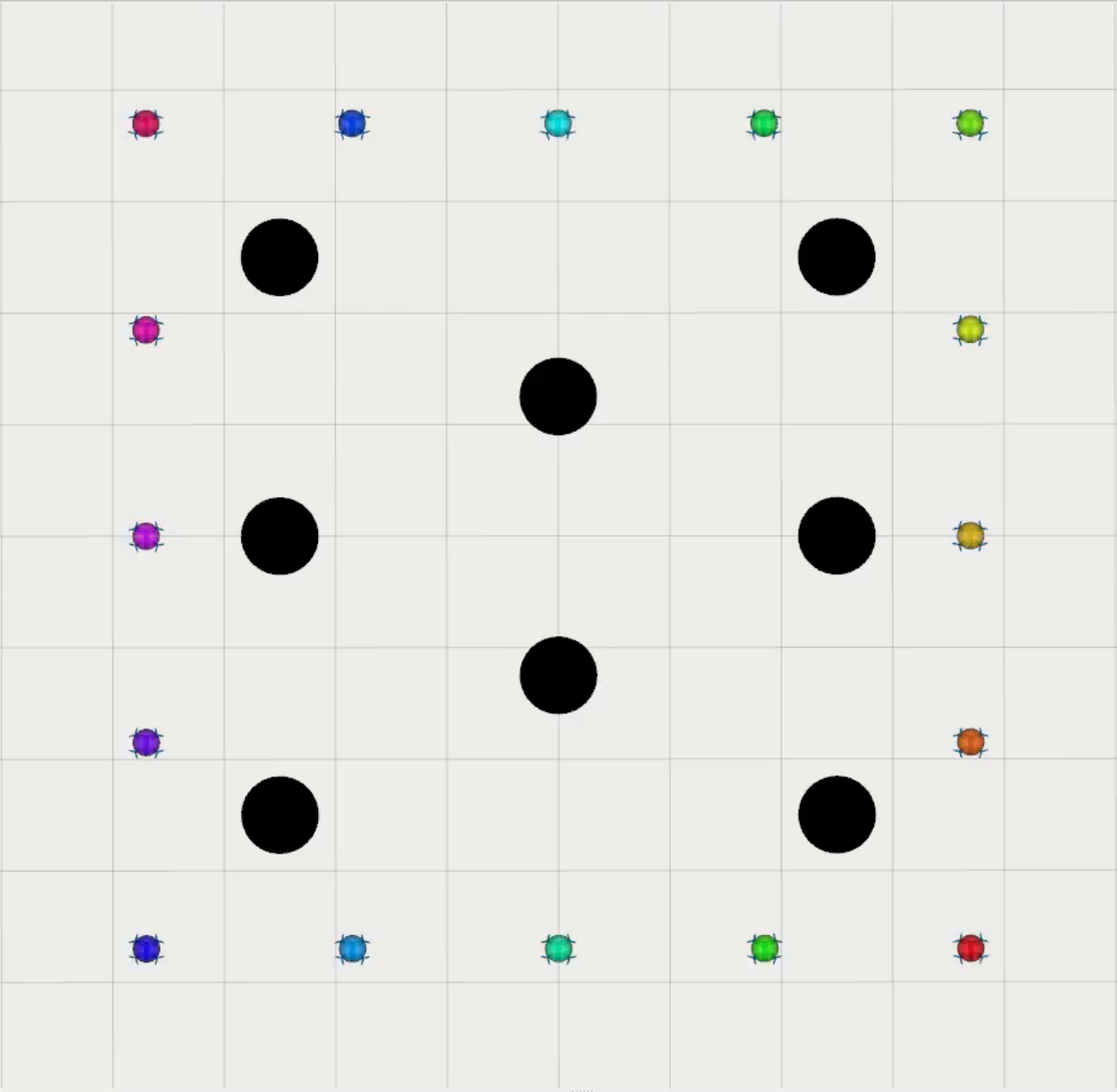}
    }%
    \subfloat[$t=1.25$]{%
        \includegraphics[width=0.19\textwidth]{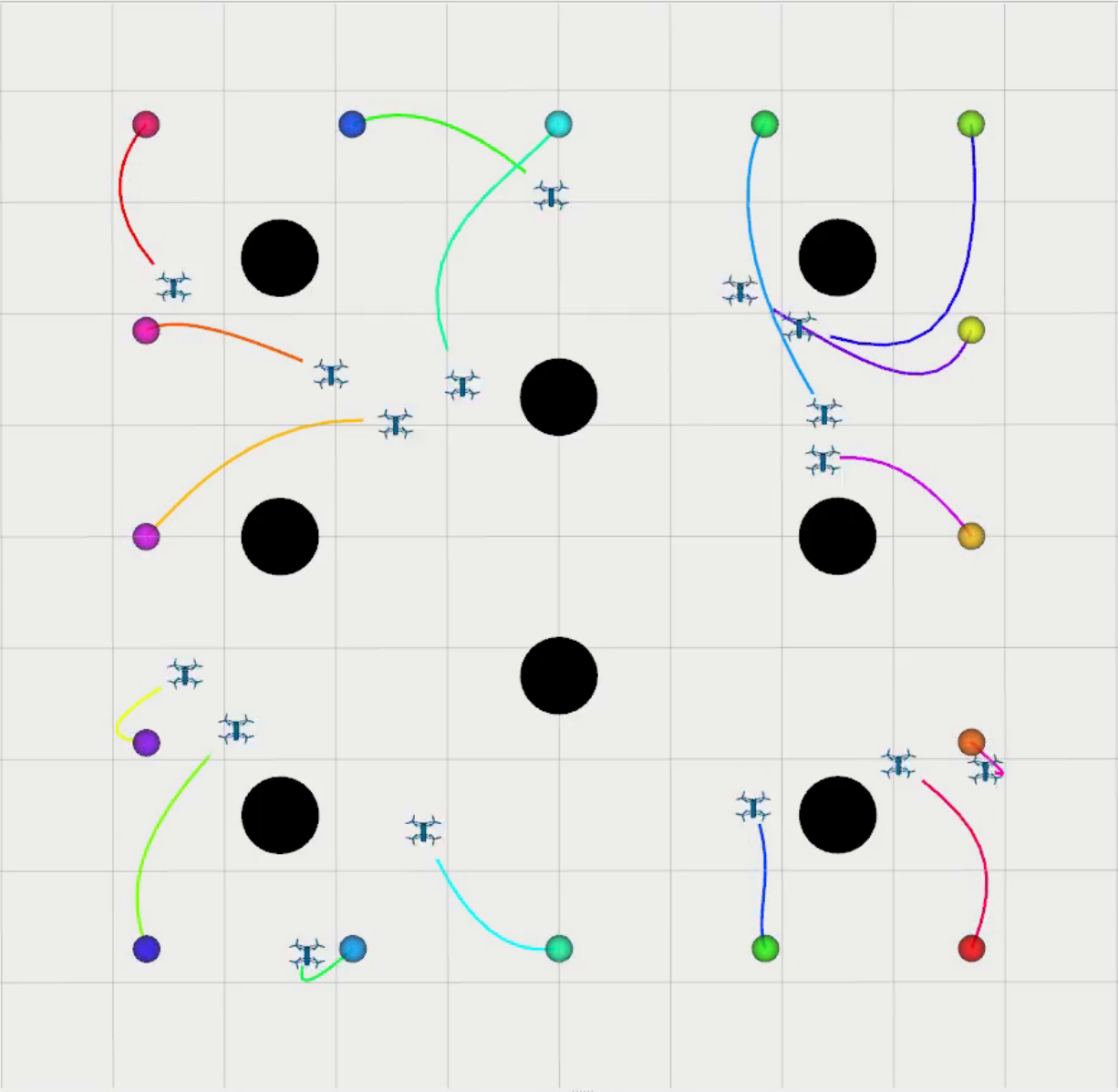}
    }%
    \subfloat[$t=2.50$]{%
        \includegraphics[width=0.19\textwidth]{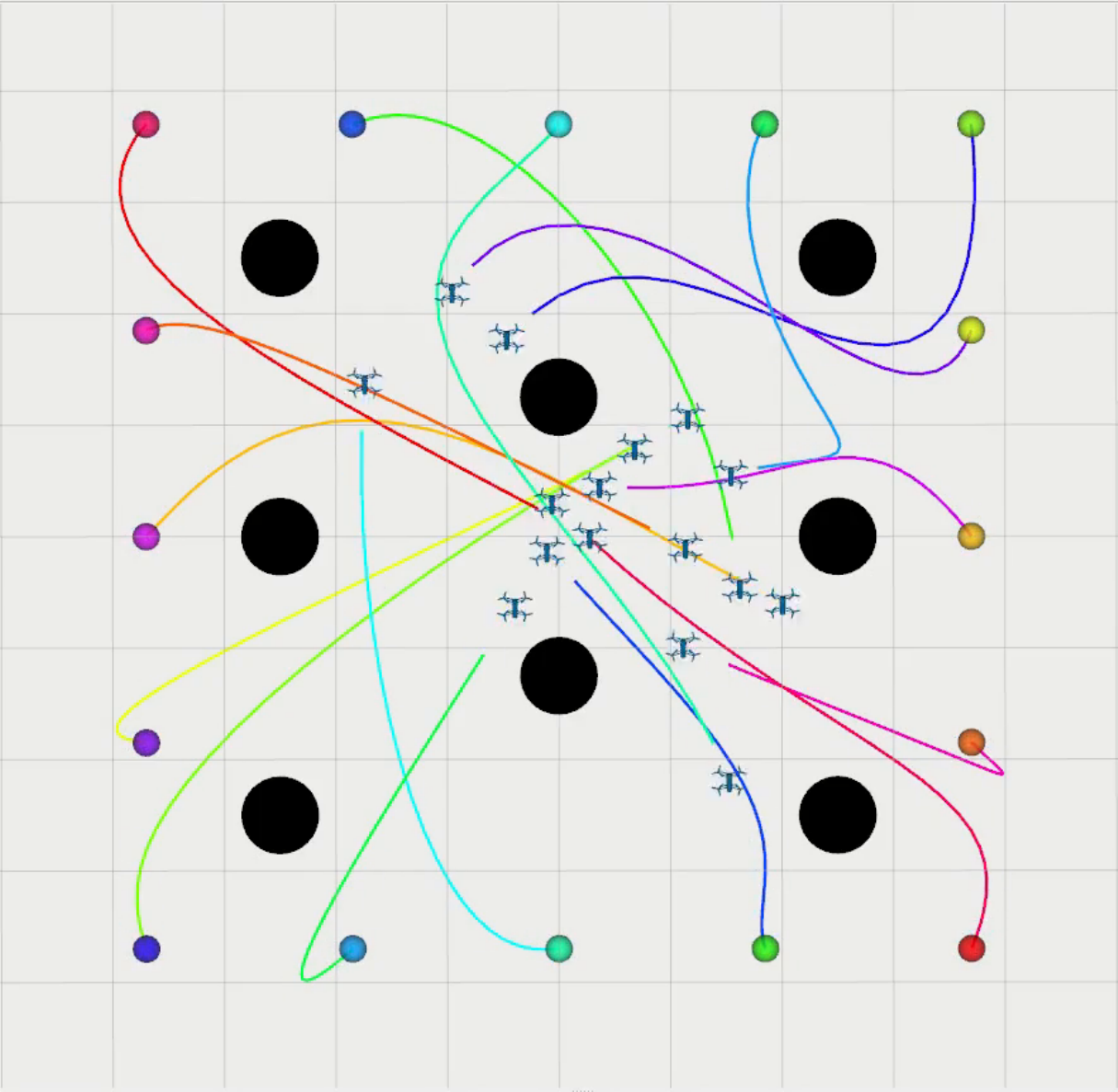}
    }%
    \subfloat[$t=3.75$]{%
        \includegraphics[width=0.19\textwidth]{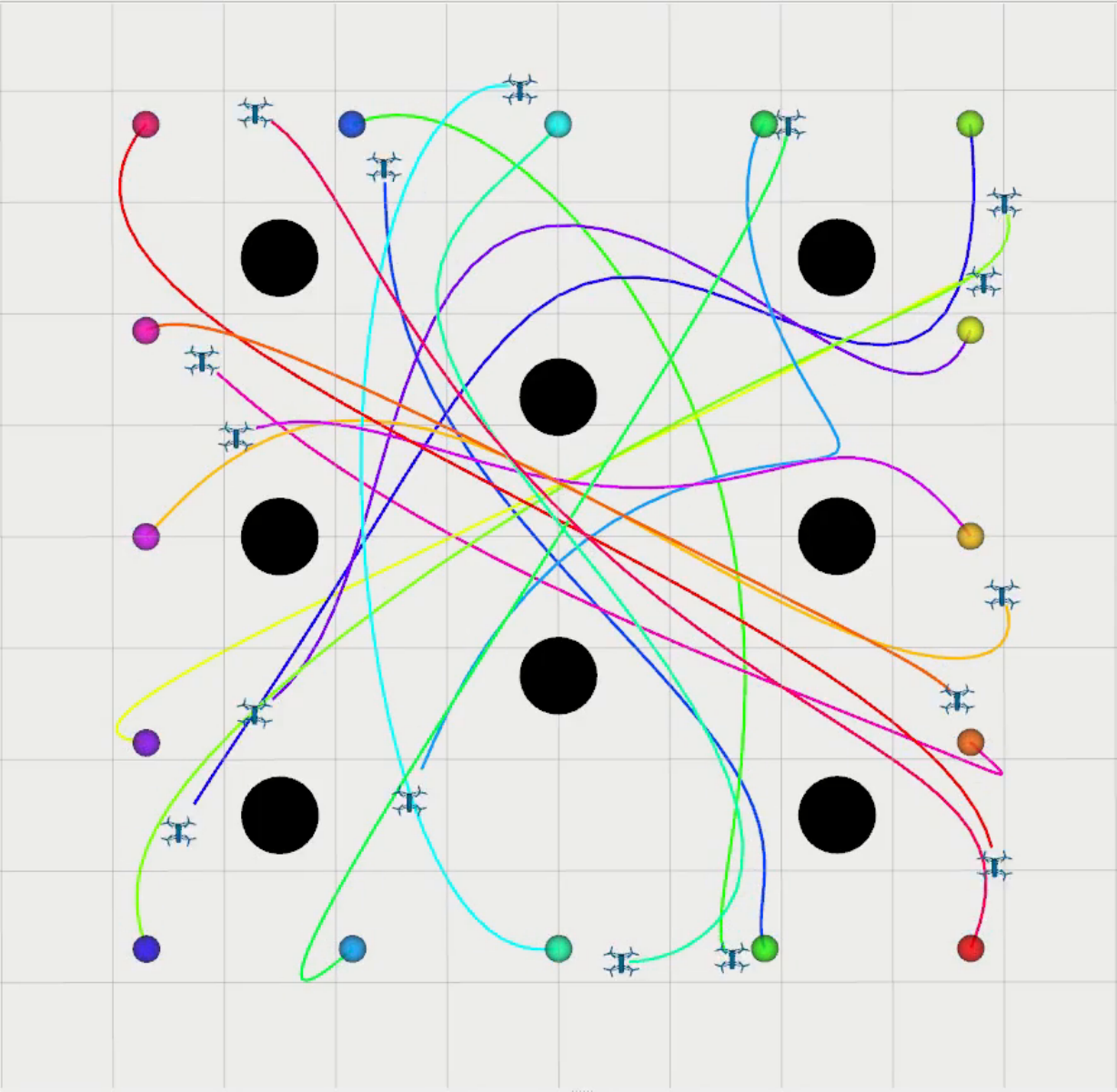}
    }%
    \subfloat[$t=5.00$]{%
        \includegraphics[width=0.19\textwidth]{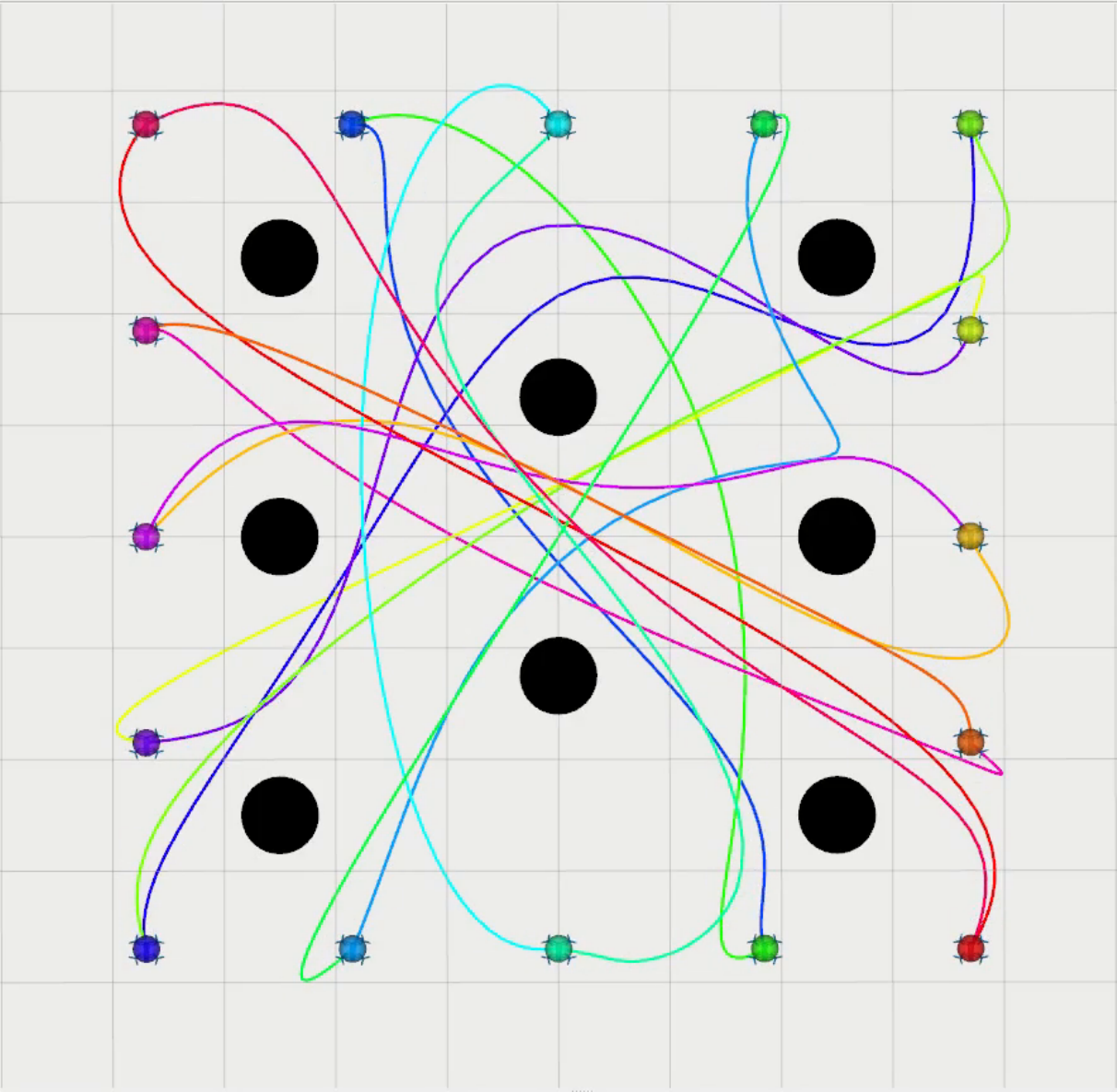}
    }
    \caption{Snapshots of multiple robots navigating in a cluttered 2D environment with static obstacles.}
    \label{fig:rviz_obs}
\end{figure*}

\begin{figure*}[t]
    \centering
    \subfloat[$t=0.00$]{%
        \includegraphics[width=0.19\textwidth]{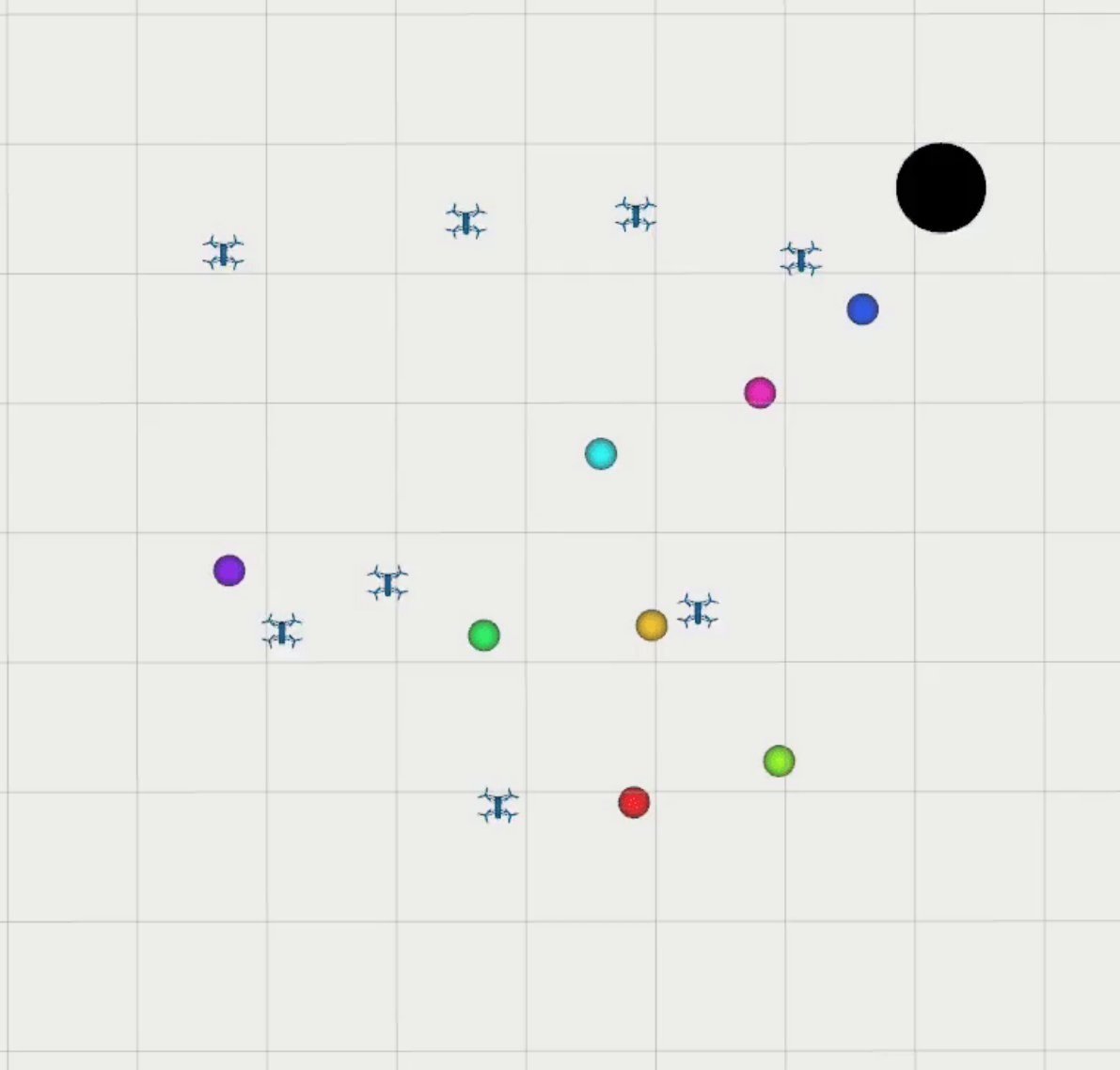}
    }%
    \subfloat[$t=1.25$]{%
        \includegraphics[width=0.19\textwidth]{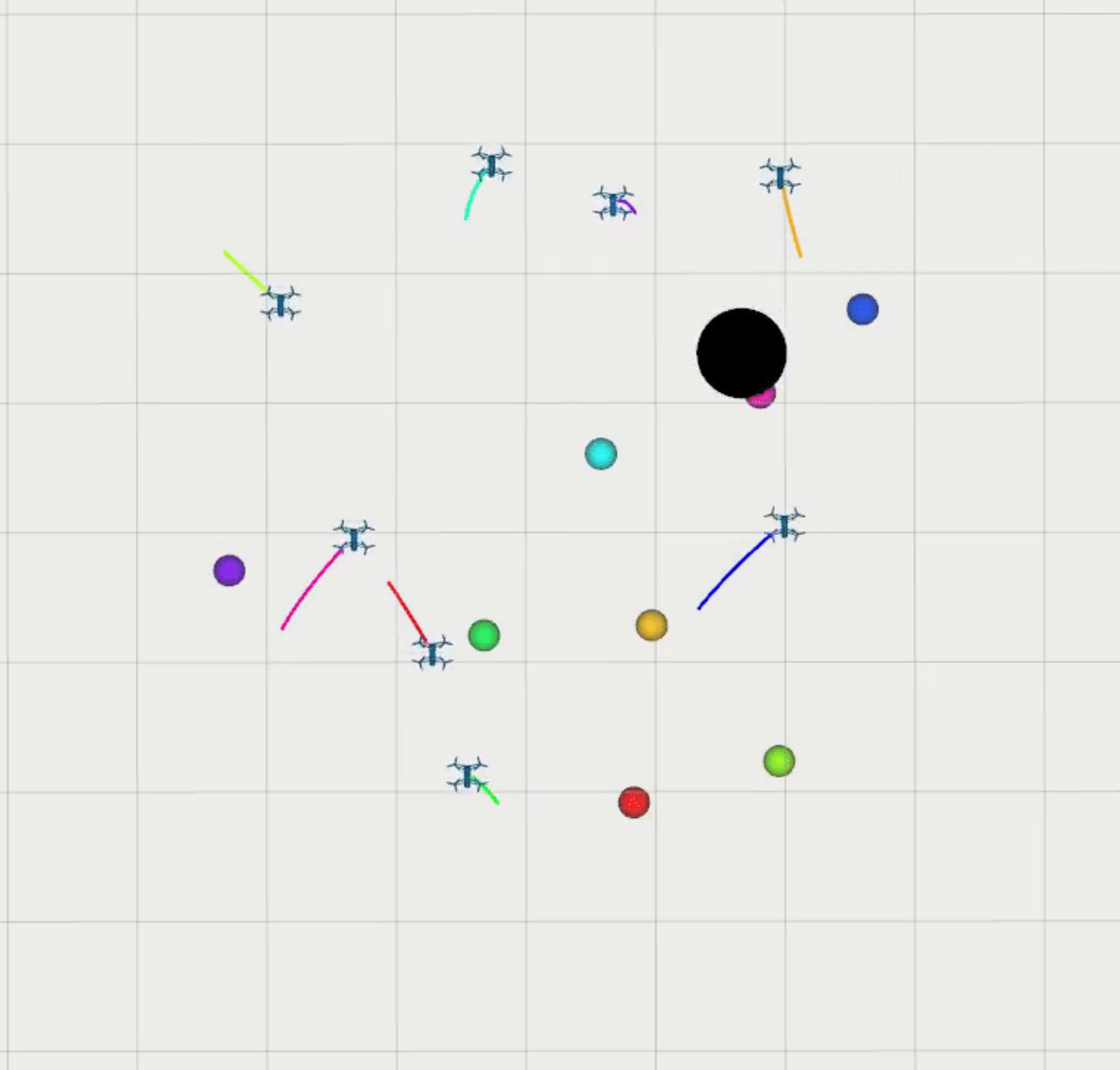}
    }%
    \subfloat[$t=2.50$]{%
        \includegraphics[width=0.19\textwidth]{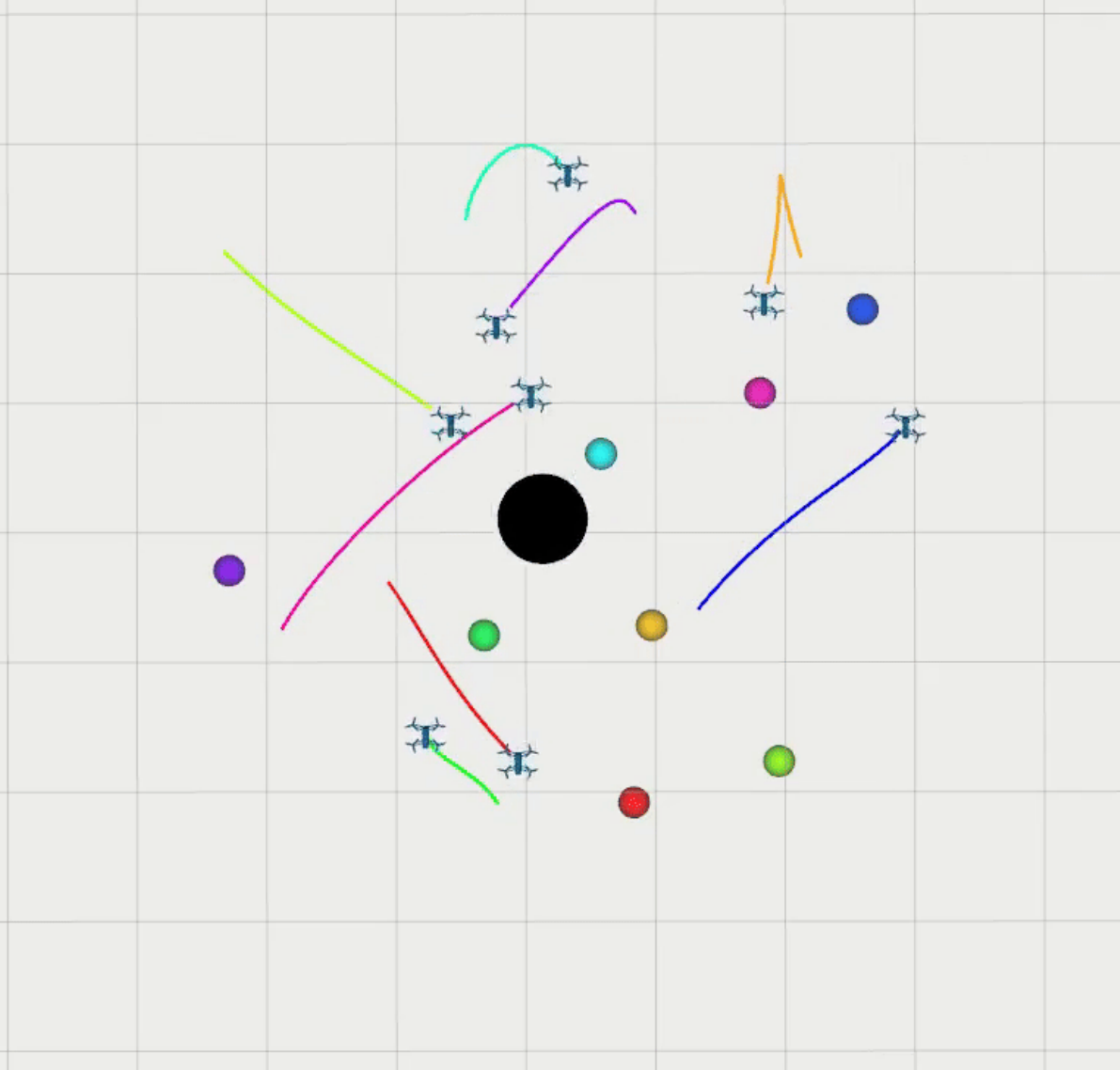}
    }%
    \subfloat[$t=3.75$]{%
        \includegraphics[width=0.19\textwidth]{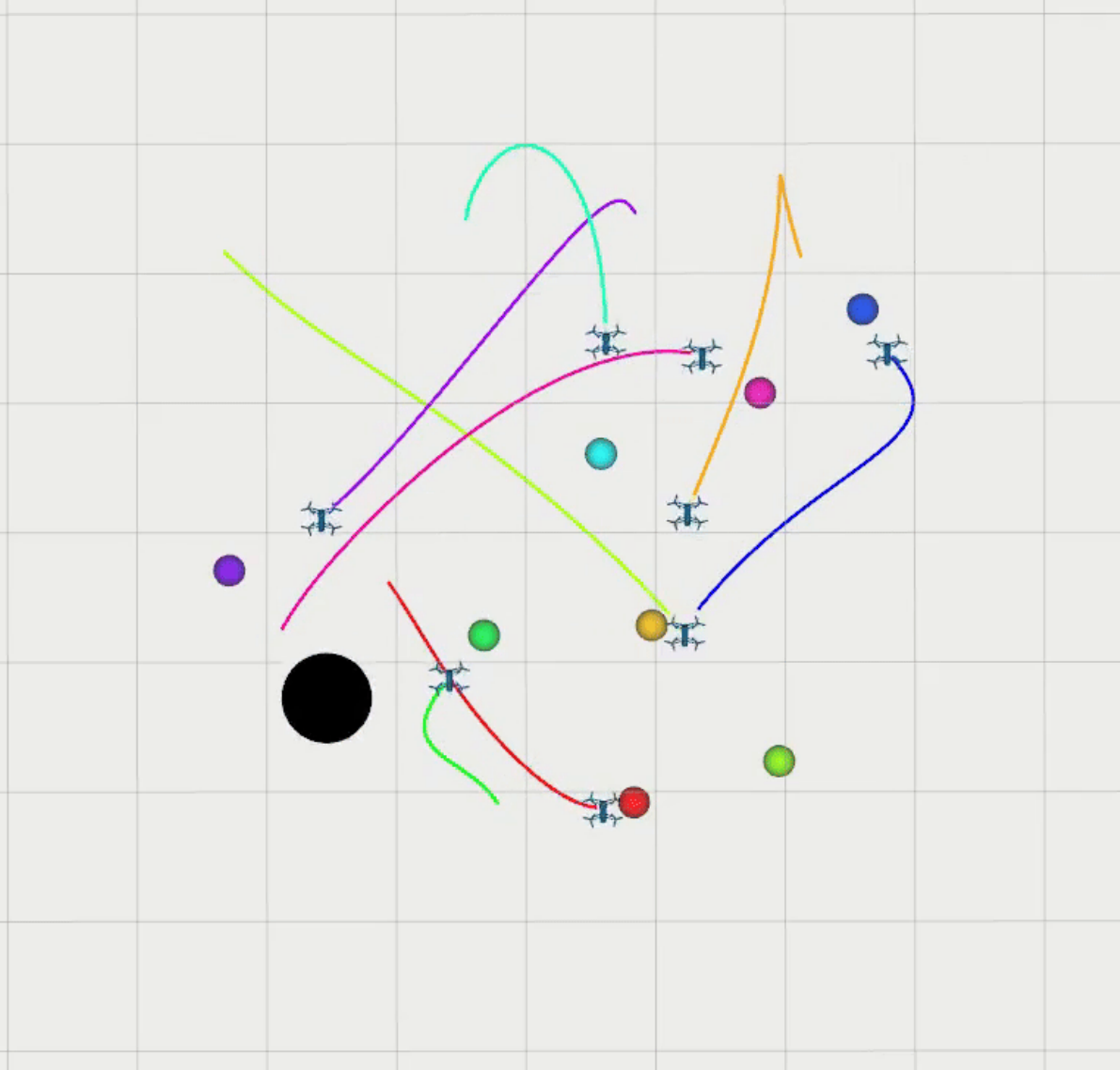}
    }%
    \subfloat[$t=5.00$]{%
        \includegraphics[width=0.19\textwidth]{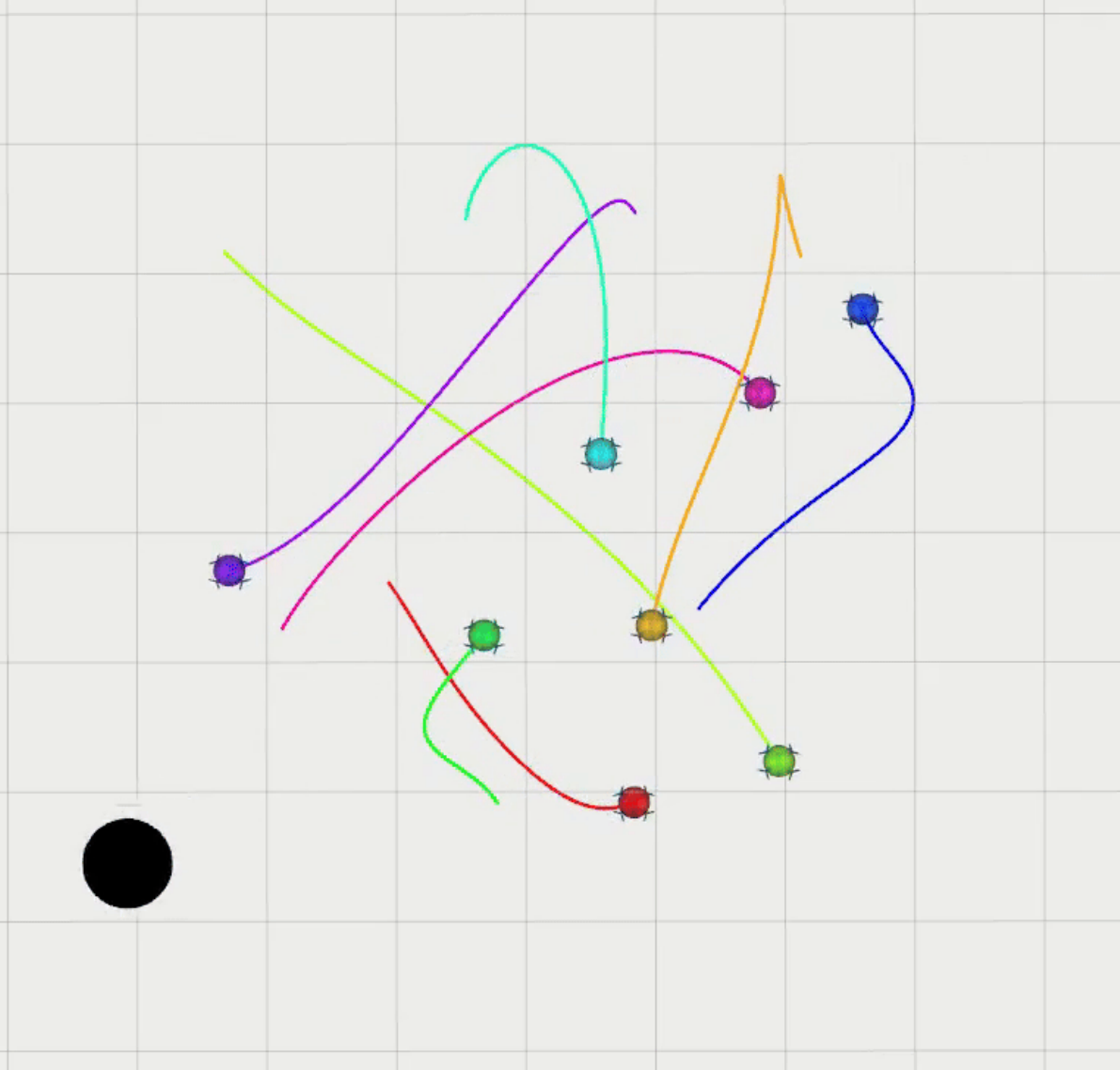}
    }
    \caption{Snapshots of eight robots navigating in the presence of a dynamic obstacle (black circle).}
    \label{fig:rviz_obs_dynamic}
\end{figure*}

\textcolor{black}{Fig.\ref{fig:rviz_obs} presents snapshots of robots navigating in cluttered 2D environments. Fig.\ref{fig:rviz_obs_dynamic} illustrates a dynamic scenario with eight robots and one moving obstacle. Several robots have start or goal positions located along or near the obstacle’s trajectory. The robots exhibit adaptive behavior by yielding to the approaching obstacle and, when necessary, temporarily deviating from a direct path to their goals before proceeding once the obstacle has passed.}

A key strength of flow matching is its ability to learn multi-modal distributions. In our context, this manifests as diverse collision avoidance behaviors, reflected in the variation in both speed and path choice among the robots. This trajectory diversity is particularly evident in environments with obstacles. Fig.\ref{fig:diversity} shows one such result for a scenario where 16 robots move between their assigned start and goal in a cluttered environment, highlighting trajectories that circumvent obstacles in different ways while adapting strategies to avoid other robots. 

% In other words, having a dedicated initialization network for warm-starting the SF solver allows us to reach a particular residual threshold at lower iterations than what can be achieved with naively using the flow output

\noindent \emph{Importance of Diversity:} One envisioned application of our approach is in generating data-driven simulation environments for training navigation policies \cite{8865441, Mavrogiannis2020BGAPBS}, \cite{nair2022dynabarn}. The diversity in multi-robot trajectories enables the simulation of rich and realistic crowd behaviors, providing a robust dataset for policy learning.

\subsection{Scalability}

\noindent A key advantage of our framework is its exceptional computational efficiency, which enables high-throughput trajectory planning by leveraging massive parallelism. We validate this performance in Figure \ref{fig_scalability}, which analyzes the scalability of our pipeline's two core components.

First, the learned flow policy provides rapid inference, enabling extensive exploration of the solution space. As shown in Figure \ref{fig_scalability}(a), the policy can generate hundreds of candidate trajectories in tens of milliseconds, even for large systems. This rapid generation is crucial to our approach, as it provides a rich set of high-quality candidates for the subsequent refinement stage.

Second, the SF solver is designed to exploit this parallelism. Figure \ref{fig_scalability}(b) demonstrates that its computation time scales nearly linearly with the number of trajectories refined simultaneously, a direct benefit of our custom GPU-vectorized solver (see \ref{Appendix}).

The combination of fast sampling and parallel refinement makes the pipeline highly effective for concurrent problem solving. For example, our system can process fifty distinct planning problems at once. Generating 10 candidate trajectories for each problem (500 total) takes less than 60 ms. Refinement of the top 5 candidates from each problem (250 total) takes less than 300 ms. Consequently, our framework can deliver high-quality, refined solutions for all fifty problems in well under a second, illustrating its strong potential for real-time, multi-task applications.

\subsection{Feasibility}

\begin{figure}[!t]
    \centering
    \subfloat[]{%
        \includegraphics[width=0.48\columnwidth]{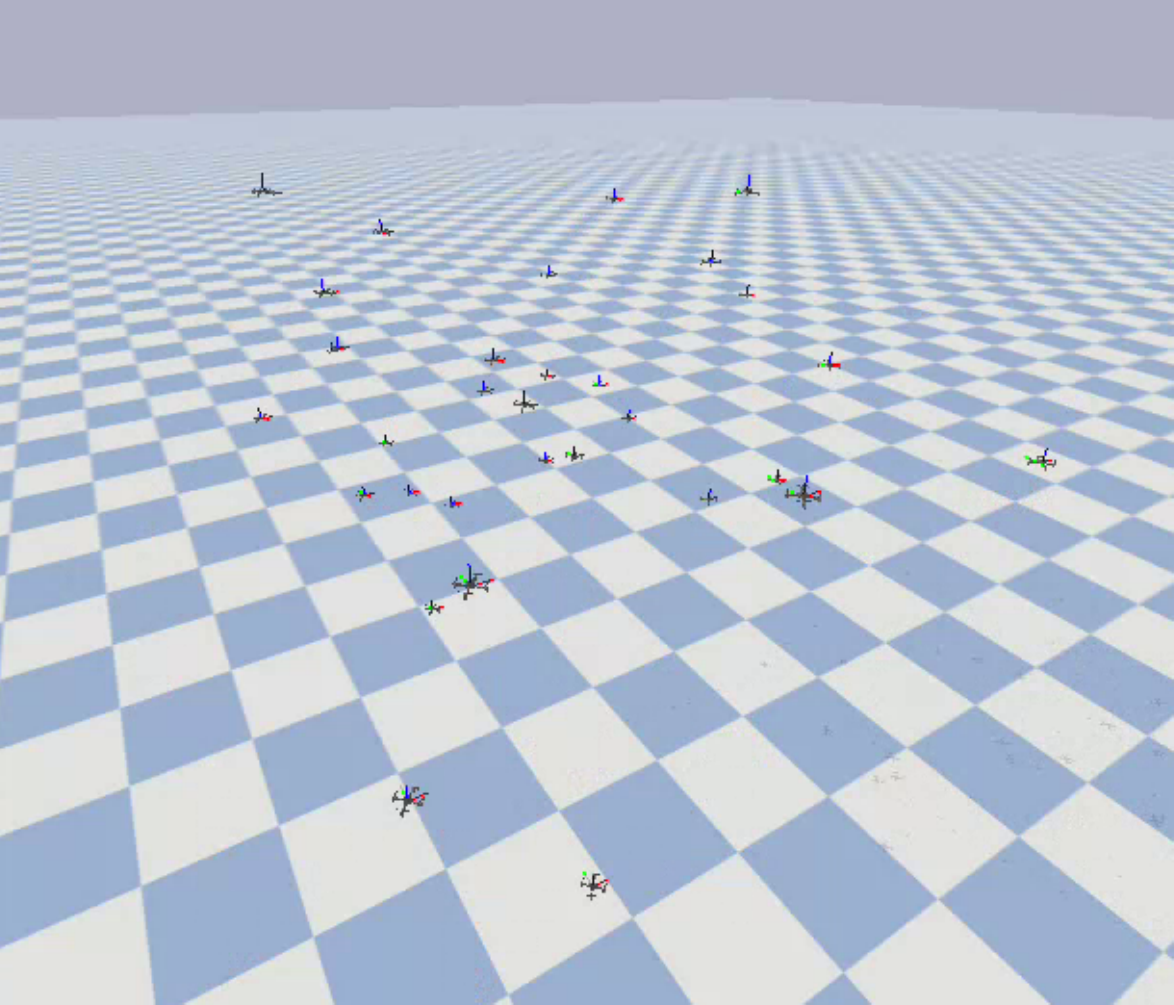}
        \label{fig:pybullet}
    }\hfill
    \subfloat[]{%
        \includegraphics[width=0.48\columnwidth]{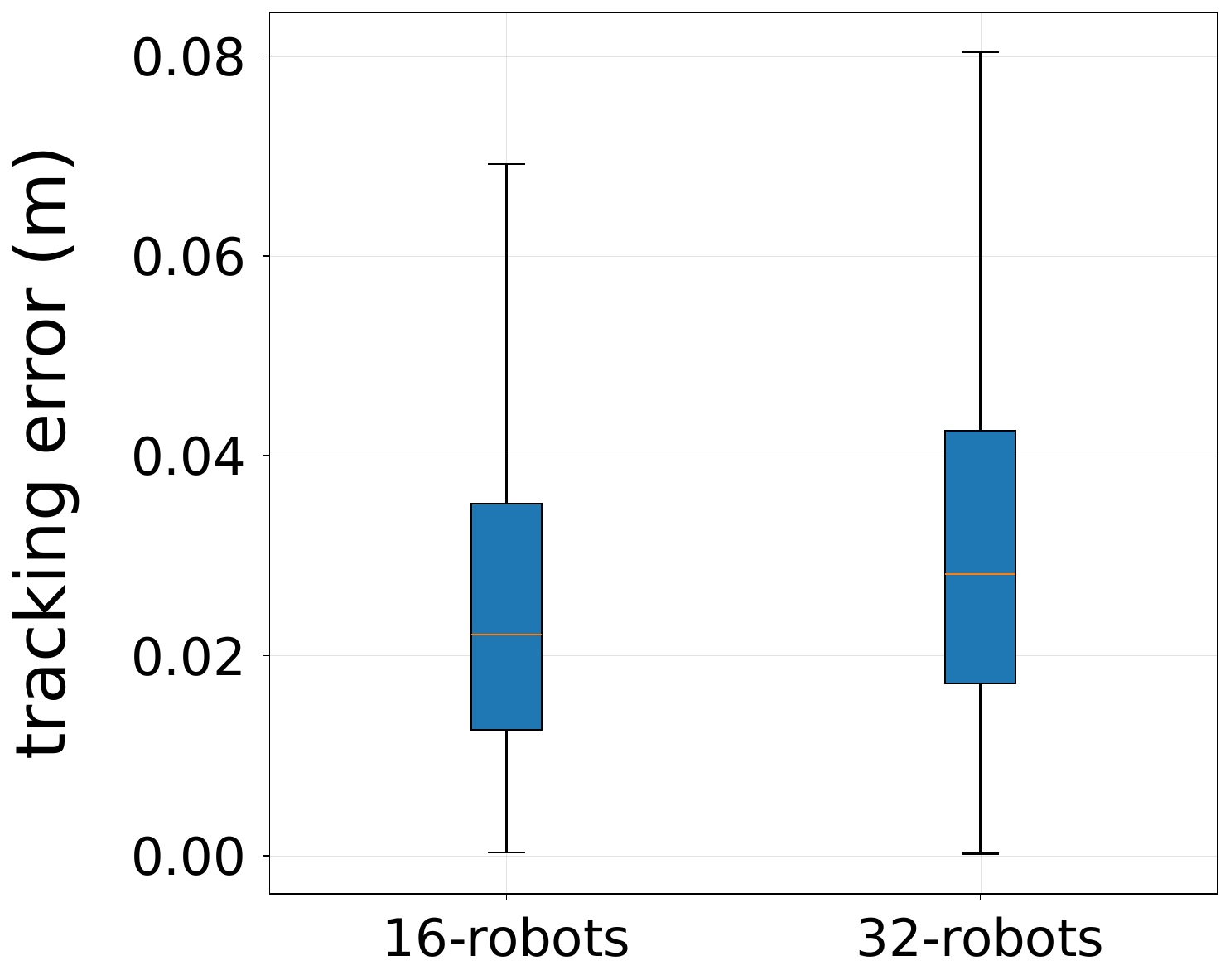}
        \label{fig:pybullet_error}
    }
    \caption{Simulation-based validation of the proposed method: (\ref{fig:pybullet}) representative PyBullet snapshot and (\ref{fig:pybullet_error}) corresponding position tracking error distribution.}
    \label{fig:pybullet_tracking}
\end{figure}

\textcolor{black}{The feasibility of the generated trajectories is validated in a PyBullet simulation environment (Fig.~\ref{fig:pybullet}) for scenarios involving 16 and 32 robots. Each robot is controlled in closed loop using a PID tracking controller. Position tracking errors are recorded at every time step of the simulation, and the resulting error distributions are summarized using box plots Fig.~\ref{fig:pybullet_error}. This provides a quantitative assessment of how accurately the generated trajectories can be followed in practice.}

\begin{figure}[!h]
    \centering
    \includegraphics[width=8.5cm]{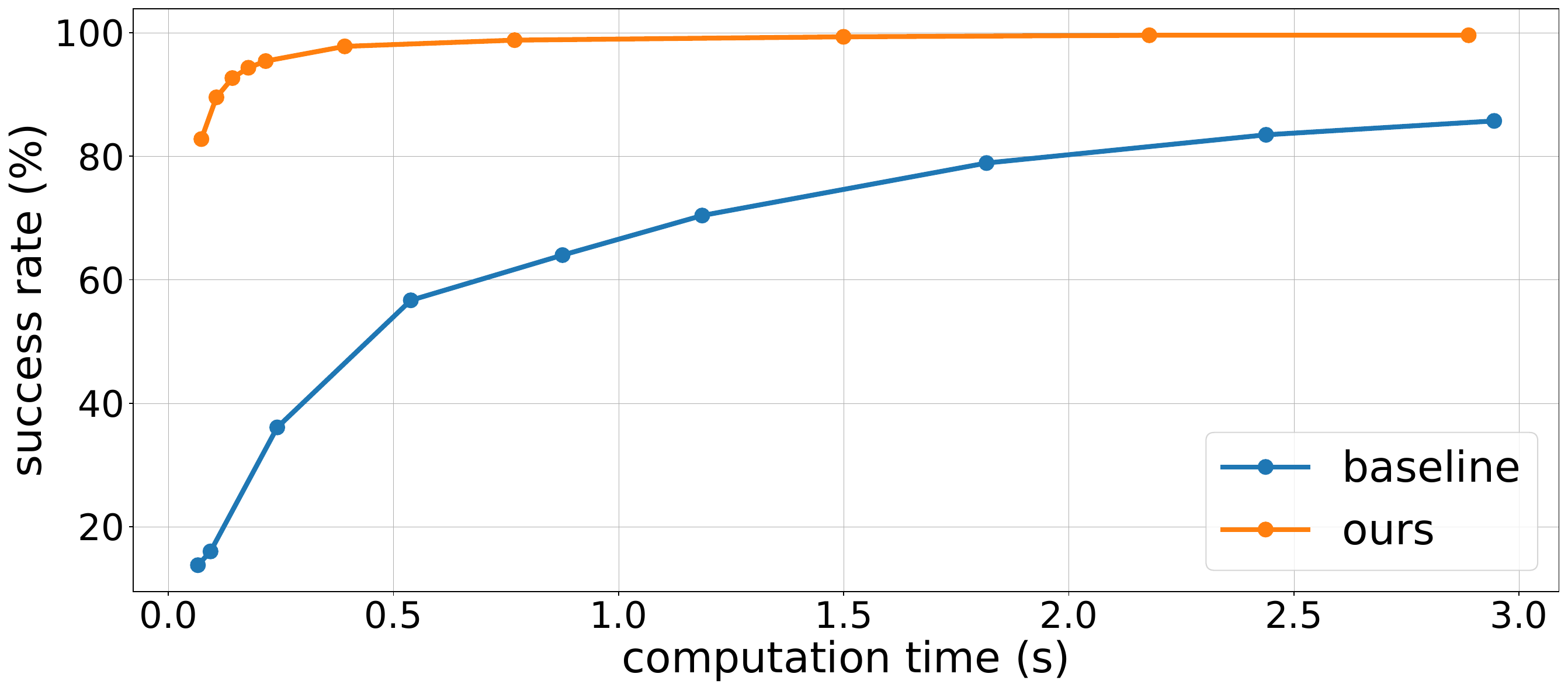}
    \caption{\footnotesize{\textcolor{black}{Success rate as a function of computation time. Our learning-based approach (orange) is compared against pure optimization baseline \cite{Rastgar2020GPUAC} (blue) in a cluttered environment with 16 robots. Our approach demonstrates rapid convergence, achieving near $100\%$ success rate within 0.5s, while the baseline exhibits slower growth, reaching approximately 85$\%$ after 3.0 s.}}}
    \label{rate_time_obs}
\end{figure}

\begin{figure*}
    \centering
    \includegraphics[scale=0.3]{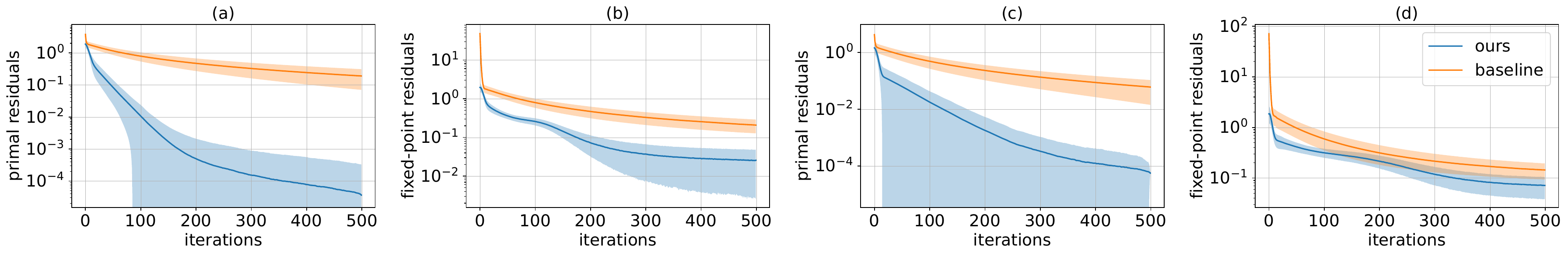}
    \caption{\footnotesize{Comparison of primal and fixed-point residuals obtained with baseline \cite{Rastgar2020GPUAC} and our approach. Fig.(a)-(b) summarizes the trends for 16 robot scenarios in 2D, while Fig.(c)-(d) reproduces the same for a 3D setting. The shaded region presents the confidence interval, while the solid lines depict the mean trend.}  }
    \label{16_robot_res}
\end{figure*}

\begin{figure*}
    \centering
    \includegraphics[scale=0.3]{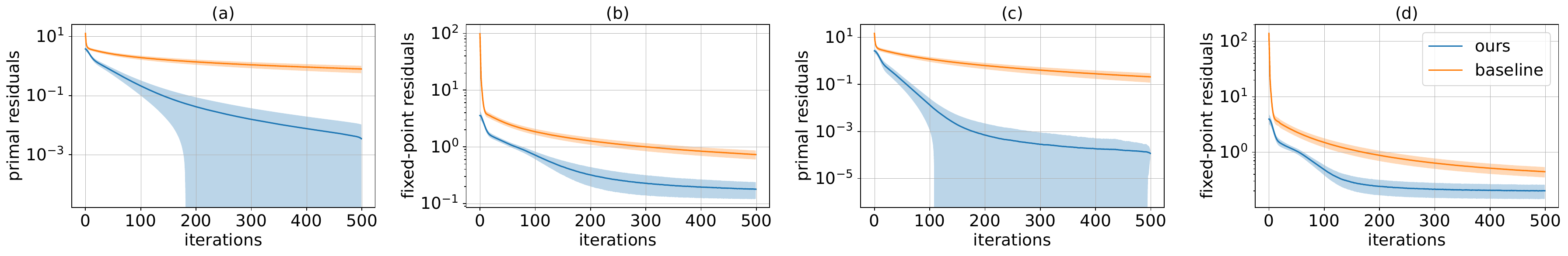}
    \caption{\footnotesize{The same results as Fig.\ref{16_robot_res} but reproduced for 32 robot setting in 2D ( Fig.(a)-(b) ) and 3D (Fig.(c)-(d)). The baseline is \cite{Rastgar2020GPUAC}.  }}
    \label{32_robot_res}
\end{figure*}

\begin{table*}[h]
    \caption{Comparison of Primal Results for Baseline \cite{Rastgar2020GPUAC} and Ours-Init Methods}
    \label{tab:baseline_vs_init}
    \renewcommand{\arraystretch}{1.2}
    \setlength{\tabcolsep}{7pt} 
    \centering
    % You need to have \usepackage{multirow} in your preamble for this to work
    \begin{tabular}{|c|c|c|c|c|c|c|c|}
    \hline
    
    % --- HEADER ---
    \multirow{2}{*}{Scenario} & \multirow{2}{*}{Primal} & \multicolumn{3}{c|}{Baseline} & \multicolumn{3}{c|}{Ours-Init} \\
    \cline{3-8}
    & & mean & max & min & mean & max & min \\
    \hline \hline

    % --- DATA for 16 robots 2D ---
    \multirow{2}{*}{16 robots 2D} & 0.01 & 2606 & 10987 & 340 & \textbf{82} & \textbf{464} & \textbf{10} \\
    \cline{2-8}
    & 0.001 & 4941 & 14827 & 667 & \textbf{110} & \textbf{499} & \textbf{10} \\
    \hline

    % --- DATA for 16 robots 3D ---
    \multirow{2}{*}{16 robots 3D} & 0.01 & 1102 & 7921 & 4 & \textbf{46} & \textbf{228} & \textbf{1} \\
    \cline{2-8}
    & 0.001 & 2099 & 9999 & 4 & \textbf{61} & \textbf{485} & \textbf{1} \\
    \hline

    % --- DATA for 32 robots 2D ---
    \multirow{2}{*}{32 robots 2D} & 0.01 & 9747 & 29758 & 1991 & \textbf{271} & \textbf{500} & \textbf{84} \\
    \cline{2-8}
    & 0.001 & 17189 & 29956 & 3787 & \textbf{338} & \textbf{500} & \textbf{102} \\
    \hline

    % --- DATA for 32 robots 3D ---
    \multirow{2}{*}{32 robots 3D} & 0.01 & 2280 & 9304 & 694 & \textbf{100} & \textbf{353} & \textbf{24} \\
    \cline{2-8}
    & 0.001 & 4190 & 9999 & 1289 & \textbf{137} & \textbf{479} & \textbf{27} \\
    \hline
    
    \end{tabular}
\end{table*}

\begin{table*}[h]
    \renewcommand{\arraystretch}{1.2}
    \setlength{\tabcolsep}{7pt}
    \centering
    \caption{Combined performance comparison of Baseline \cite{Rastgar2020GPUAC} and Our method in 2D and 3D planning scenarios.}
    \label{tab:combined_results}
    \begin{tabular}{|c|c|c|c|c|c|c|}
    \hline
    
    % Header
    \multirow{2}{*}{Dimension} & \multirow{2}{*}{\begin{tabular}[c]{@{}c@{}}\# of\\ robots\end{tabular}} & \multirow{2}{*}{Method} & \multicolumn{4}{c|}{Metrics} \\
    \cline{4-7}
    & & & time (s) & success rate & smoothness cost ($[m/s^2]$) & arc length ($[m]$) \\
    \hline \hline
    
    % 2D Data
    \multirow{6}{*}{2D} & \multirow{2}{*}{8} & Baseline \cite{Rastgar2020GPUAC} & 0.218 & 100 & 0.2646 & 1.2966 \\
    \cline{3-7}
    & & Ours & \textbf{0.0515} & 100 & \textbf{0.2141} & \textbf{1.0489} \\
    \cline{2-7}
    & \multirow{2}{*}{16} & Baseline \cite{Rastgar2020GPUAC} & 0.5927 & 99.9 & 0.278 & 1.3624 \\
    \cline{3-7}
    & & Ours & \textbf{0.0654} & \textbf{100} & \textbf{0.2398} & \textbf{1.1748} \\
    \cline{2-7}
    & \multirow{2}{*}{32} & Baseline \cite{Rastgar2020GPUAC} & 3.4068 & 99.61 & \textbf{0.2027} & \textbf{0.9934} \\
    \cline{3-7}
    & & Ours & \textbf{0.2966} & \textbf{100} & 0.241 & 1.1809 \\
    \hline \hline
    
    % 3D Data
    \multirow{6}{*}{3D} & \multirow{2}{*}{16} & Baseline \cite{Rastgar2020GPUAC} & 0.1482 & 100 & 0.6313 & 3.0932 \\
    \cline{3-7}
    & & Ours & \textbf{0.069} & 100 & \textbf{0.6272} & \textbf{3.0731} \\
    \cline{2-7}
    & \multirow{2}{*}{32} & Baseline \cite{Rastgar2020GPUAC} & 0.999 & 99.81 & \textbf{0.6375} & \textbf{3.1237} \\
    \cline{3-7}
    & & Ours & \textbf{0.1511} & \textbf{100} & 0.7002 & 3.4309 \\
    \cline{2-7}
    & \multirow{2}{*}{64} & Baseline \cite{Rastgar2020GPUAC} & 33.7 & 99.71 & \textbf{0.7143} & \textbf{3.5003} \\
    \cline{3-7}
    & & Ours & \textbf{1.02} & \textbf{99.93} & 0.8065 & 3.9518 \\
    \hline
    \end{tabular}
    \label{baseline_vs_ours_full}
\end{table*}

\begin{figure*}
    \centering
    \includegraphics[scale=0.3]{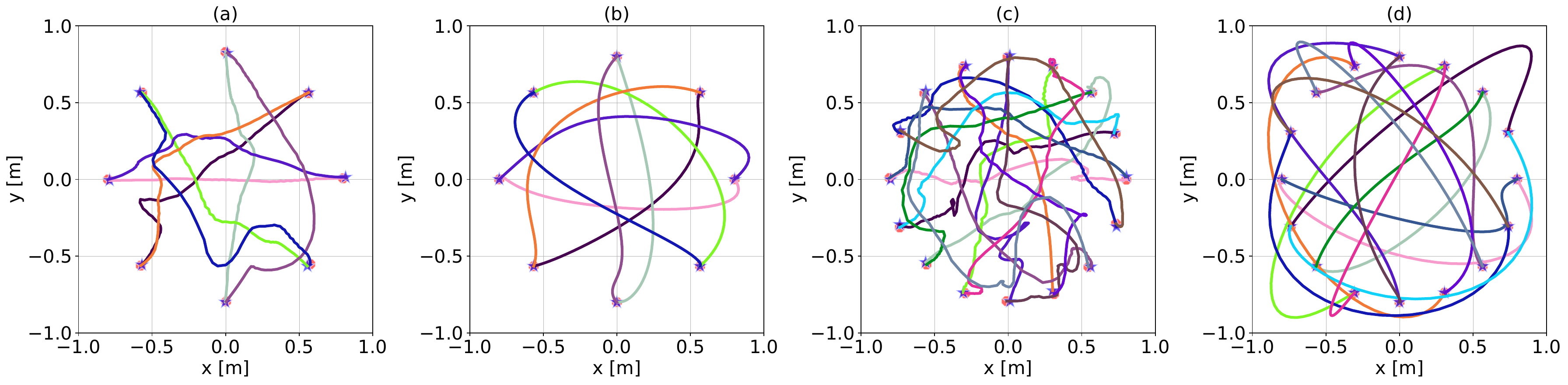}
    \caption{\footnotesize{Comparison between our approach and a diffusion-based multi-robot planner MMD \cite{mmd}. Fig.(a)-(b) shows MMD and our approach in a 8 robot scenario respectively. Fig.(c)-(d) repeats the comparison for 16 robots with Fig.(c) being the MMD and Fig.(d) depicting our approach. As can be seen, our approach models the global interaction between the robots better leading to smoother de-conflicting maneuvers.}  }
    \label{ours_vs_mmd}
\end{figure*}

\begin{table*}[h]
    \caption{Comparison of MMD \cite{mmd} and Our Method on 2D Scenarios}
    \label{tab:mmd_vs_ours_2d}
    \renewcommand{\arraystretch}{1.2}
    \setlength{\tabcolsep}{7pt} 
    \centering
    % You need to have \usepackage{multirow} in your preamble for this to work
    \begin{tabular}{|c|c|c|c|c|c|}
    \hline
    Method & Num. of Robots & Time (s) & Success Rate & Smoothness cost ($[m/s^2]$) & Arc Length ($[m]$) \\
    \hline \hline
    
    % --- MMD Data ---
    \multirow{3}{*}{MMD} & 8 & 5.3706 & 100 & \textbf{0.2063} & \textbf{1.0155} \\
    \cline{2-6}
    & 16 & 18.0 & 100 & \textbf{0.2393} & 1.1779 \\
    \cline{2-6}
    & 32 & 48.1758 & 100 & 0.2487 & 1.2239 \\
    \hline
    
    % --- Ours Data ---
    \multirow{3}{*}{Ours} & 8 & \textbf{0.0515} & 100 & 0.2141 & 1.0489 \\
    \cline{2-6}
    & 16 & \textbf{0.0654} & 100 & 0.2398 & \textbf{1.1748} \\
    \cline{2-6}
    & 32 & \textbf{0.2966} & 100 & \textbf{0.2410} & \textbf{1.1809} \\
    \hline
    
    \end{tabular}
    \label{mmd_vs_ours_table}
\end{table*}

\subsection{Comparison with \cite{Rastgar2020GPUAC} }
% \noindent We compare our hybrid approach, which combines aspects of data-driven reasoning and trajectory optimization, to a centralized baseline that solves the problem from scratch without any learning. To this end, we select \cite{Rastgar2020GPUAC} as the baseline due to its close relation to our method. In fact, the SF solver described in Appendix \ref{Appendix} extends \cite{Rastgar2020GPUAC} by incorporating additional constraints and augmenting an initialization network. Our analysis focuses on the metrics presented in Section \ref{metrics}. \textcolor{black}{In particular, we compare how fast \cite{Rastgar2020GPUAC} and our approach is able to converge to a trajectory with low fixed-point and primal residuals. The latter is formulated in Eqn. \eqref{primal_residual} and serves as a key metric for quantifying the satisfaction of critical constraints: workspace limits \eqref{workspace_con}, inter-robot separation \eqref{inter_robot_con}, and obstacle avoidance \eqref{collision_con}.}

\noindent We compare our hybrid approach, which combines aspects of data-driven reasoning and trajectory optimization, to a centralized baseline that solves the problem from scratch without any learning. To this end, we select \cite{Rastgar2020GPUAC} as the baseline due to its close relation to our method. In fact, the SF solver described in Appendix \ref{Appendix} extends \cite{Rastgar2020GPUAC} by incorporating additional constraints and augmenting an initialization network. Our analysis focuses on the metrics presented in Section \ref{metrics}. \textcolor{black}{In particular, we evaluate the convergence of trajectories in terms of fixed-point and primal residuals, with the latter formulated in Eq.\eqref{primal_residual}, which quantify the satisfaction of key constraints: workspace limits \eqref{workspace_con}, inter-robot separation Eqn.\eqref{inter_robot_con}, and obstacle avoidance Eqn.\eqref{collision_con}.}

% \begin{itemize}
%     \item \textbf{Convergence:} How fast each approach is able to converge to a trajectory with low fixed-point and primal residuals. The latter is formulated in Eqn. \eqref{primal_residual}, serves as a key metric, quantifying the satisfaction of critical constraints: workspace limits \eqref{workspace_con}, inter-robot separation \eqref{inter_robot_con}, and obstacle avoidance \eqref{collision_con}.
%     \item \textbf{Success-Rate and Computation Time:} Given an upper-bound of $10k$ iterations, what is the percentage of problems solved by our approach and \cite{Rastgar2020GPUAC} and the corresponding computation time?
%     \item \textbf{Trajectory Quality:} The smoothness cost (trajectory acceleration norm) and arc-length associated with trajectories obtained with both methods.
% \end{itemize}

Fig.\ref{16_robot_res}-\ref{32_robot_res} shows the decay of the primal and fixed-point residuals across iterations for different problem scenarios. As can be seen, our approach delivers substantially faster residual decay, especially at low iteration counts, by leveraging learning on prior collected datasets. To further quantify the performance gain provided by our approach, Table \ref{tab:baseline_vs_init} reports the number of iterations \cite{Rastgar2020GPUAC} and our approach require to achieve primal residual values below $0.01$ and $0.001$. This statistic is crucial as it directly corresponds to the speed at which each method can converge to a (approximately) feasible trajectory. Consider the 2D scenario with 16 robots: the mean iteration number needed by our approach to reach a residual of $0.01$ is 82. This is 30 times less than the mean number of iterations required by \cite{Rastgar2020GPUAC}. The performance gap increases further when considering a primal residual threshold of $0.001$. Similar trends persist for other planning scenarios summarized in Table \ref{tab:baseline_vs_init}.

Table \ref{baseline_vs_ours_full} compares \cite{Rastgar2020GPUAC} and our approach in terms of computation time, success rate, and trajectory quality\footnote{To give the best possible chance to \cite{Rastgar2020GPUAC}, we gave an upper bound of 10K iterations to it. }. While both methods achieve comparable trajectory quality and success rates, our method achieves a significant reduction in computation time. For instance, in 8-robot planning problems, our approach is four times faster than \cite{Rastgar2020GPUAC}. The performance gap grows further with an increase in the number of robots. In 16-robot scenarios, our approach is up to an order-of-magnitude faster in 2D setting and more than two times faster in 3D setting. The 2D setting is particularly challenging due to the reduced maneuvering space available to each robot. In this difficult setting, our approach provides almost an order of magnitude speed-up over \cite{Rastgar2020GPUAC}.

\textcolor{black}{We also evaluate our approach against \cite{Rastgar2020GPUAC} in static-obstacle environments with 16 robots. The results are summarized in Fig. \ref{rate_time_obs}, which illustrates the relationship between success rate and computation time. As shown, our approach achieves a near $100\%$ success rate in under 0.5 s, whereas the baseline fails to reach this threshold even after 3 s. Moreover, our method attains over $80\%$ success rate within just 50 ms, demonstrating how the proposed learning-based formulation significantly accelerates the trajectory planning process.}

\begin{remark}
\textcolor{black}{In summary, our learning-based approach provides excellent speed-up over \cite{Rastgar2020GPUAC}, improved success rate in cluttered environments, while producing trajectories of comparable quality.  }    
\end{remark}

\subsection{Comparison with Diffusion Based Multi-Robot Planner MMD \cite{mmd}}
\noindent We compare our approach with MMD \cite{mmd}, which leverages diffusion policies and combines with graph search methods to plan multi-robot trajectories. A qualitative comparison is shown in Fig.\ref{ours_vs_mmd} for a scenario in which 16 robots are placed on the perimeter of a circle and have to move to their antipodal position. Both approaches produce distinctively different kinds of trajectories. The MMD trajectories direct robots toward the center of the circle, which is the main conflict point in this scenario. Subsequently, the robots coordinate to navigate safely towards the goal. While MMD successfully produced safe trajectories, it does not start de-conflicting until the robots are very close to each other. This is because the underlying diffusion policy does not capture the multi-robot interactions as it was trained on the dataset of just single-robot motions \cite{mmd}. In contrast, our flow policy can capture global interactions between the robots and chooses a de-conflicting maneuver that results in more clearance between the robots. 

\begin{figure}
    \centering
    \includegraphics[width=7cm]{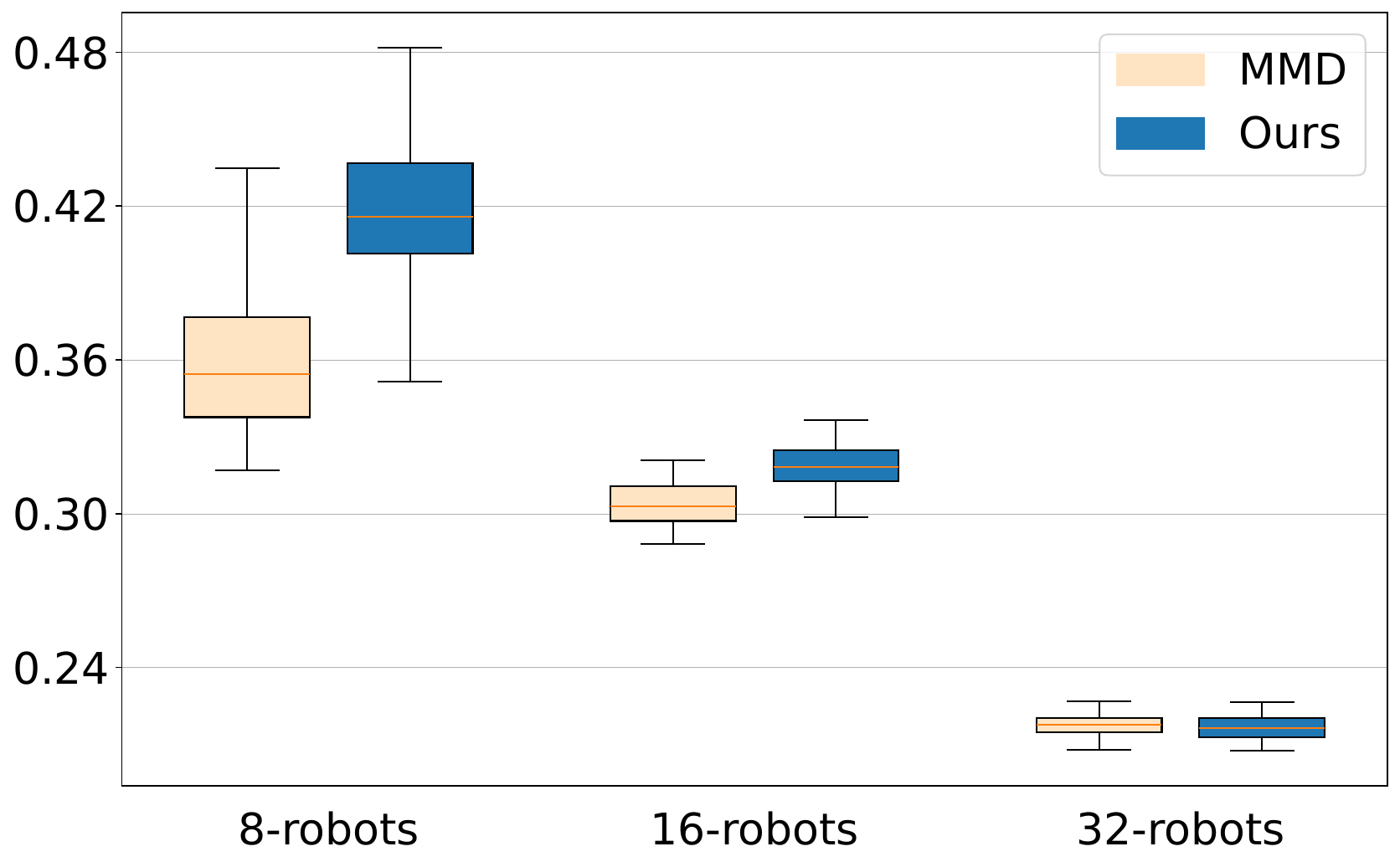}
    \caption{\footnotesize{Comparison of inter-robot distances resulting from MMD \cite{mmd} and our approach. If free space is available, as in the case of 8 and 16 robot planning problems, our approach can find trajectories with higher clearance between the robot trajectories. As the number of robots increases and space becomes restrictive, both approaches converge to similar behavior.}  }
    \label{clearance_mmd_ours}
\end{figure}

To further quantify the benefits that our approach provides in terms of inter-robot clearance, we considered 1000 random 2D planning scenarios with varying numbers of robots. Start and goal positions were sampled from a rectangular area of $[-1, 1]\times [-1, 1]$, with each robot having a circular footprint of radius $0.1m$. Fig.\ref{clearance_mmd_ours} summarizes the average pairwise robot distances for MMD and our approach. As can be seen, in 8-robot planning scenarios, our approach produces close to 16 $\%$  improvement in average inter-robot distances. Marginal improvements are observed for 16 robots, and these diminish further as the number of robots increases. Both methods converge to the same level in terms of inter-robot clearance due to reduced maneuvering space. Nevertheless, it can be concluded that in the presence of more maneuvering space, our approach is more likely to leverage it than MMD.

% The space becomes too cluttered, and both methods eventually provide the same performance. Nevertheless, it can be concluded that if there is more maneuvering space, our approach is more likely to leverage it than MMD.

Table \ref{mmd_vs_ours_table} compares the overall performance of our approach and MMD in terms of the three metrics mentioned in Section \ref{metrics}. Both approaches achieve comparable average trajectory quality (smoothness cost, arc-length) and success rates across 1000 different problem instances, each for 8, 16, and 32 robots. However, the computation-time trends are strikingly different. The slow diffusion policy underlying MMD and computationally expensive conflict-search results in substantially higher computation time than our approach. For example, in the 8-robot scenario, our approach is an order of magnitude faster than MMD, and the gap grows to $160$ times for the 32-robot planning instances. 

\begin{remark}
\textcolor{black}{In summary, our approach exploits free space more efficiently than MMD, producing collision-free trajectories significantly faster while maintaining comparable trajectory quality and success rates.}    
\end{remark}

\subsection{Comparison with \cite{park2020efficient}}
\noindent In this subsection, we benchmark our approach against \cite{park2020efficient}, which employs a batch-sequential planning strategy. This method first partitions the robots into distinct groups. For each group, it performs joint trajectory optimization, considering all robots within that specific group simultaneously. Following this initial grouping and intra-group planning, the approach transitions to a sequential planning phase to coordinate the actions between the different groups. This means that the groups have a predetermined order, and the trajectory planning for each group is carried out one by one. A significant characteristic of this method is that each group must treat all previously planned groups as dynamic obstacles.

\begin{figure*}
    \centering
    \includegraphics[scale=0.3]{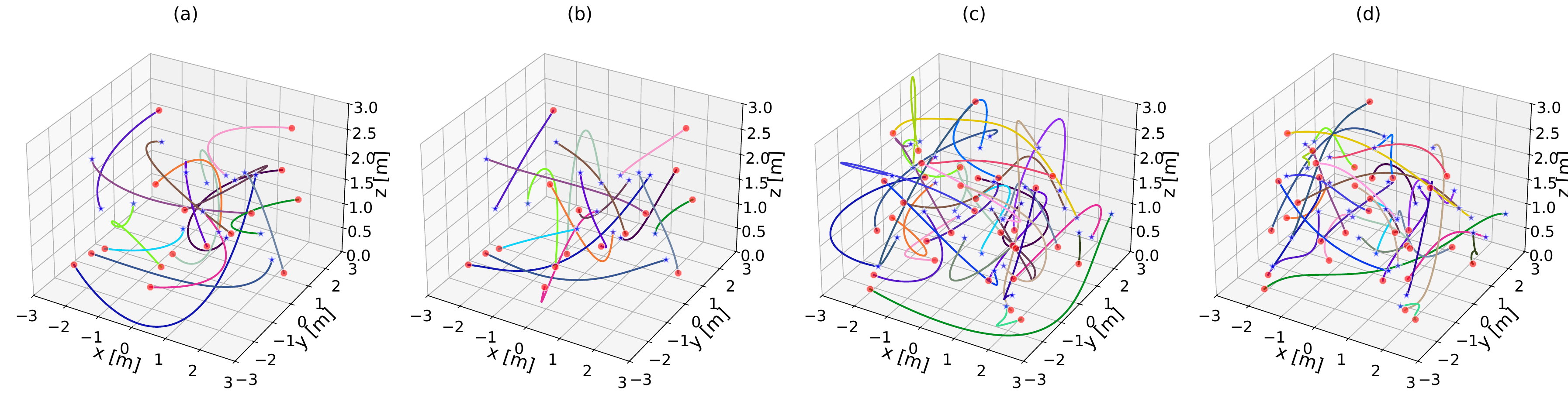}
    \caption{\footnotesize{Comparison between our approach and \cite{park2020efficient}. Fig.(a)-(b) considers the 3D planning scenario involving 16 robots with (a) being \cite{park2020efficient} and Fig.(b) depicting our approach. Similarly, Fig.(c) and (d) show \cite{park2020efficient}'s and our performance, respectively, while planning for 32 robots in 3D.  It can be seen that our approach produces less curved trajectories, which follow a shorter path between the start and goal.}  }
    \label{park_vs_ours}
\end{figure*}

A qualitative comparison between our approach and \cite{park2020efficient} is shown in Fig.\ref{park_vs_ours}. We consistently observed that \cite{park2020efficient} produced longer and more curved trajectories. This is because the sequential approach of Fig.\ref{park_vs_ours} restricts the potential for more complex and cooperative group interactions. For instance, a group planned later in the sequence cannot influence the trajectory of a group planned earlier, even if a slight adjustment could lead to a more globally optimal or efficient solution for the entire multi-robot team. In contrast, our approach considers the inter-robot interactions more efficiently by searching in the joint feasible space of all the robots simultaneously.

Table \ref{table_park_vs_ours} further reinforces the trend observed in Fig.\ref{park_vs_ours}. Across 16 and 32 robot planning instances, our approach achieved shorter trajectories by $33 \%$ and $29\%$, respectively. The restricted planning space available to \cite{park2020efficient} also affects its success rate, which is $89 \%$ and $69\%$ for 16 and 32 robot planning problems, respectively. In contrast, our approach successfully solves all the problems in this benchmark. Our approach is also almost an order of magnitude faster than \cite{park2020efficient} in all problem instances \footnote{The substantially lower success rate of \cite{park2020efficient} observed here is attributed to the tighter workspace constraints imposed in our evaluation}. 

\begin{remark}
\textcolor{black}{In summary, our approach outperforms \cite{park2020efficient} across all metrics discussed in Section \ref{metrics}, including trajectory length, success rate, and computation time.}    
\end{remark}

\begin{table*}[h]
    \renewcommand{\arraystretch}{1.2}
    \setlength{\tabcolsep}{7pt} 
    \centering
    \caption{Performance comparison of our proposed method against \cite{park2020efficient} in 3D environments for 16 and 32 robots.}
    \label{table_park_vs_ours}
    \begin{tabular}{|c|c|c|c|c|c|c|}
    \hline
    & \multicolumn{3}{|c|}{\cite{park2020efficient} on 3D} & \multicolumn{3}{|c|}{Ours on 3D} \\
    \hline
    Num. of Robots & Time (s) & Success Rate (\%) & Arc Length ($[m]$) & Time (s) & Success Rate (\%) & Arc Length ($[m]$)\\
    \hline
    16 & 0.54 & 89.89 & 4.0669 & \textbf{0.069} & \textbf{100} & \textbf{3.0731} \\
    \hline
    32 & 1.72 & 63.12 & 4.4407 & \textbf{0.1511} & \textbf{100} & \textbf{3.4309} \\
    \hline
    \end{tabular}
\end{table*}

\subsection{Ablations and Additional Results}

\subsubsection{Importance of an Initialization Network in SF Convergence} 
\begin{figure*}
    \centering
    \includegraphics[scale=0.30]{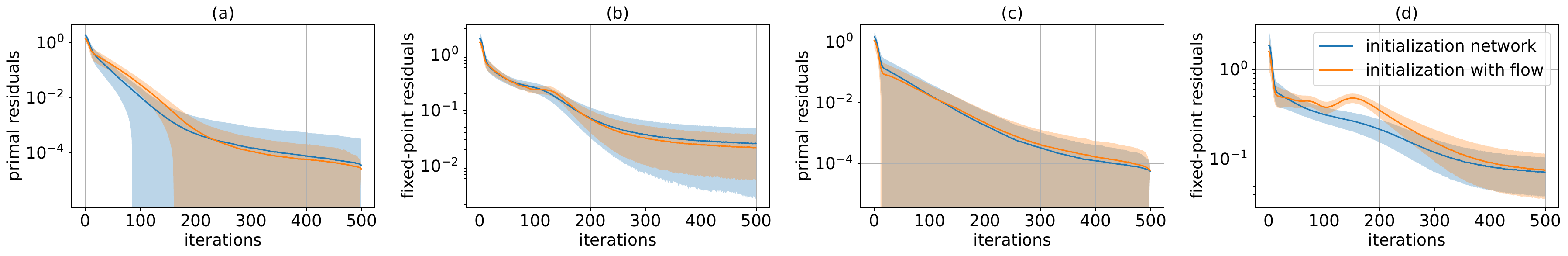}
    \caption{\footnotesize{Primal and fixed-point residuals for the SF solver for 16 robot planning problem in (2D) (Fig.(a)-(b)) and 3D (Fig. (c)-(d)). We show the trend for two different initialization strategies: one where the SF solver is directly initialized with the flow policy output and in the second, it is warm-started using the initialization network conditioned on the flow policy output. It can be seen that having a dedicated initialization network leads to a faster decay of primal and fixed-point residuals.} }
    \label{res_16}
\end{figure*}

\begin{figure*}
    \centering
    \includegraphics[scale=0.30]{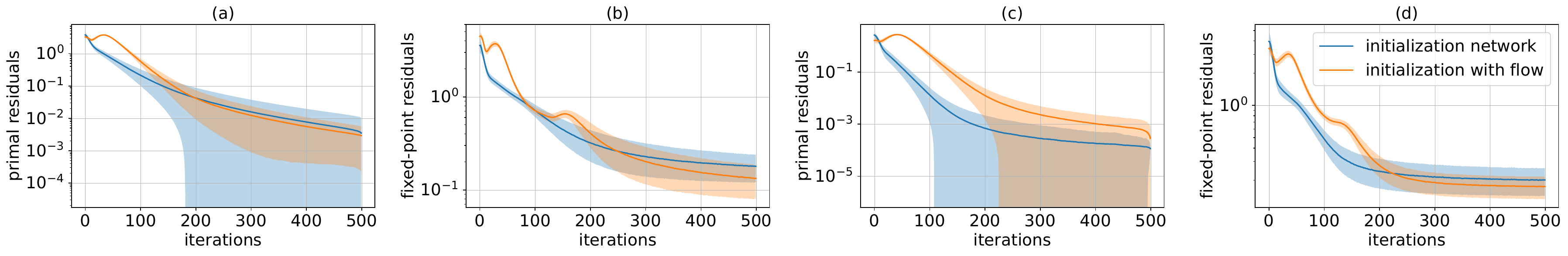}
    \caption{\footnotesize{Primal and fixed-point residuals for the SF solver for 32 robot planning problem in (2D) (Fig.(a)-(b)) and 3D (Fig. (c)-(d)). The results carry similar meaning as Fig.\ref{res_16}.}}
    \label{res_32}
\end{figure*}

We now analyze the critical role of our flow-conditioned initialization network in warm-starting the SF solver. In an ablation study, we bypassed the initialization network and initialized the SF solver directly with outputs from the flow policy. This tests whether the near-optimal trajectories produced by the flow policy alone suffice for robust convergence.
This direct approach is significantly less effective as shown in Figures \ref{res_16} and \ref{res_32}. The decay of both primal and fixed-point residuals is substantially slower than our full model, including the initialization network. This outcome highlights a key design insight: the flow policy and the SF solver have distinct objectives. The flow policy is trained solely to imitate expert data and is therefore agnostic to the optimization landscape of the downstream solver. In contrast, the initialization network is explicitly trained to bridge this divide. It learns to refine the flow policy's output, transforming it into a starting point that is not just near-optimal in trajectory space, but is also a more effective initial guess for the SF solver's specific optimization process.

% In this subsection, we analyze the importance of having a network conditioned on the flow policy (recall Fig.\ref{unroll_learning}) output for warm-starting the SF. To this end, we adopt an alternate approach of initializing the SF directly with the flow policy output and by-passing the initialization network. The hypothesis that we intend to test is that since flow produced trajectories are close to optimality, does initializing the SF with them could still produce a good convergence trend. The results are summarized in Fig.\ref{res_16}-\ref{res_32}. We can clearly observe that directly initializing the SF with the flow output does not produce as good a primal and fixed-point residual decay as that obtained with the initialization network. One reason for this is the flow policy is trained purely to mimic expert trajectories. During training time, the flow model is not aware of its initialization is going to be used by the downstream SF solver; a gap which is filled by having a dedicated flow-conditioned initialization network.

\begin{figure*}
    \centering
    \includegraphics[scale=0.30]{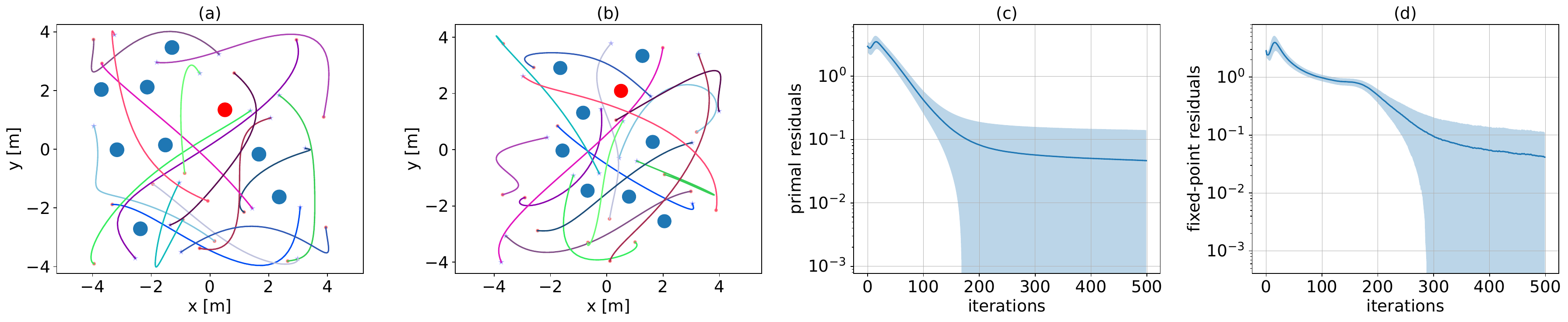}
    \caption{\footnotesize{Robustness of the SF solver to a minor out-of-distribution perturbation in obstacle count. The initialization network is trained with a fixed number of obstacles (eight, shown in cyan); the flow model generates trajectory candidates conditioned on this count and does not observe the extra obstacle. At test time, one additional static obstacle is introduced before the SF optimizer runs, adding an extra set of collision-avoidance constraints to the problem. As seen from Fig.(a)--(b), the SF successfully produces smooth and feasible trajectories despite this perturbation. Fig.(c)--(d) further corroborates this; the primal and fixed-point residual decay remains comparable to the unperturbed setting shown in Fig.\ref{16_robot_res}. }   }
    \label{extra_obs_results}
\end{figure*}

% \subsubsection{Adaptation to Perturbation in Problem Parameters} Our initialization network is trained on environments with a fixed number of obstacles, raising a critical question: can it generalize to novel problem configurations at inference time? To evaluate this robustness, we designed test scenarios in which we introduced an additional obstacle that was not present during the training phase (Figures \ref{extra_obs_results}(a)-(b)).
% Despite this out-of-distribution perturbation, our SF solver competes strongly, successfully computing smooth and feasible trajectories. This qualitative success is corroborated by quantitative analysis, as shown in Figures \ref{extra_obs_results}(c)-(d). The rapid decay of the primal and fixed-point residuals confirms that the solver maintains its efficient convergence, underscoring its robustness to unforeseen environmental changes.

\subsubsection{Adaptation to Perturbation in Problem Parameters}: \textcolor{revision_color}{Because our initialization network is trained on environments with a fixed number of obstacles, a natural question arises: can it still provide an effective warm-start when the SF encounters a structurally modified problem at inference time? To investigate this, we designed test scenarios where a single additional static obstacle---unseen during training---is introduced after the flow model and initialization network generate their trajectory candidates, but before the SF optimizer is invoked (Figures~\ref{extra_obs_results}(a)-(b)). Consequently, while the generative models operate on the original obstacle configuration, the SF optimizer must accommodate the new obstacle via an additional set of collision-avoidance constraints.}

\textcolor{revision_color}{Despite this out-of-distribution perturbation, the SF consistently recovers smooth and feasible trajectories. This is corroborated quantitatively in Figures~\ref{extra_obs_results}(c)-(d): across 1000 perturbed scenarios, the primal and fixed-point residual statistics decay rapidly and remain comparable to the previously reported unperturbed baseline. This confirms that the initialization network yields a warm-start that remains valid even when the downstream optimizer must satisfy additional constraints. Furthermore, because the extra obstacle is placed randomly within a tight workspace, it frequently escalates the problem's difficulty. For instance, in Figures~\ref{extra_obs_results}(a)-(b), the added obstacle directly obstructs the nominal path between the start and goal for several robots. We note that this test has a limited scope but it does sheds some light into the robustness of our approach.}

\subsubsection{Asymptotic Optimality}
\begin{figure}[!t]
    \centering
    \includegraphics[width=5cm]{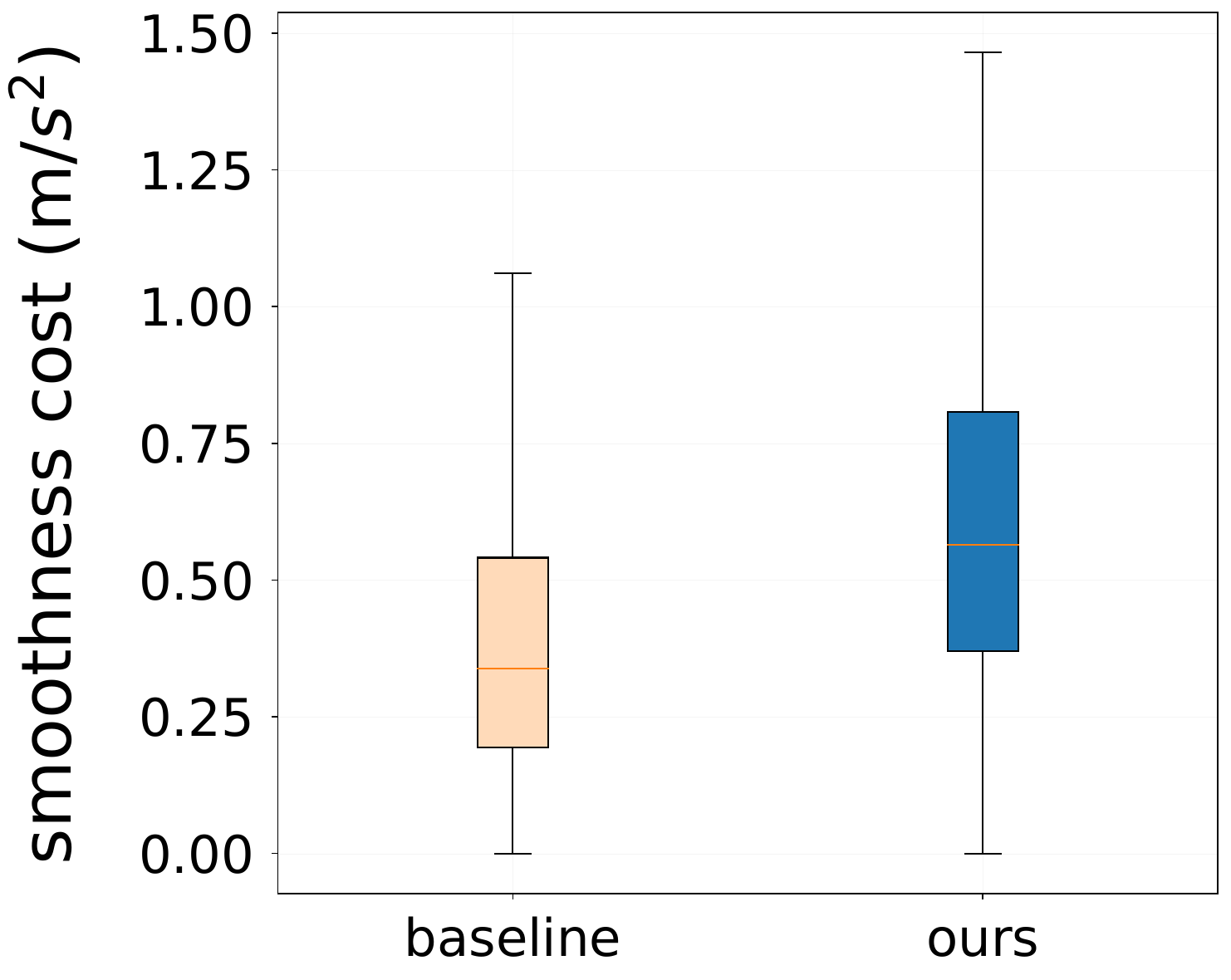}
    \caption{\footnotesize{\textcolor{black}{Optimality analysis comparing the smoothness of trajectories generated by our method versus the asymptotic solution of the baseline. \cite{Rastgar2020GPUAC}. While the baseline achieves near-optimal smoothness (median cost $< 0.33\,m/s^2$), our learning-based approach exhibits a higher smoothness cost (median $\approx 0.56\,m/s^2$). This gap represents the trade-off accepted to achieve the massive speed-up in computation time.}}  }
    \label{smoothness_box}
\end{figure}

\textcolor{black}{Figure~\ref{smoothness_box} illustrates the trade-off between solution optimality and computational speed. We compared the smoothness cost (sum of square accelerations of all the robots, see \eqref{cost}) of our approach to a baseline allowed to run to near-convergence. As expected, the baseline yields smoother trajectories, achieving a median cost of $\approx 0.33\,{m/s^2}$ compared to $\approx 0.56\,{m/s^2}$ for our method. Although our approach incurs a higher smoothness cost—indicating a larger deviation from the global optimum—this trade-off is acceptable given the massive speed-up in computation time detailed in Table \ref{baseline_vs_ours_full} and  Fig.~\ref{rate_time_obs}.}

%%%%%%%%%%%%%%%%%%%%%%%%%%%%%%%%%%%%%%%%%%%%%%%%%%%%%%%%%%%%%%%%%%%%%%%%%%%%%%%
% NEW RESULTS

\subsection{Real-World Experiments}
\begin{figure*}[t]
    \centering
    \subfloat[$t=2.00$]{%
        \begin{minipage}{0.25\textwidth}
            \centering
            \includegraphics[width=\linewidth]{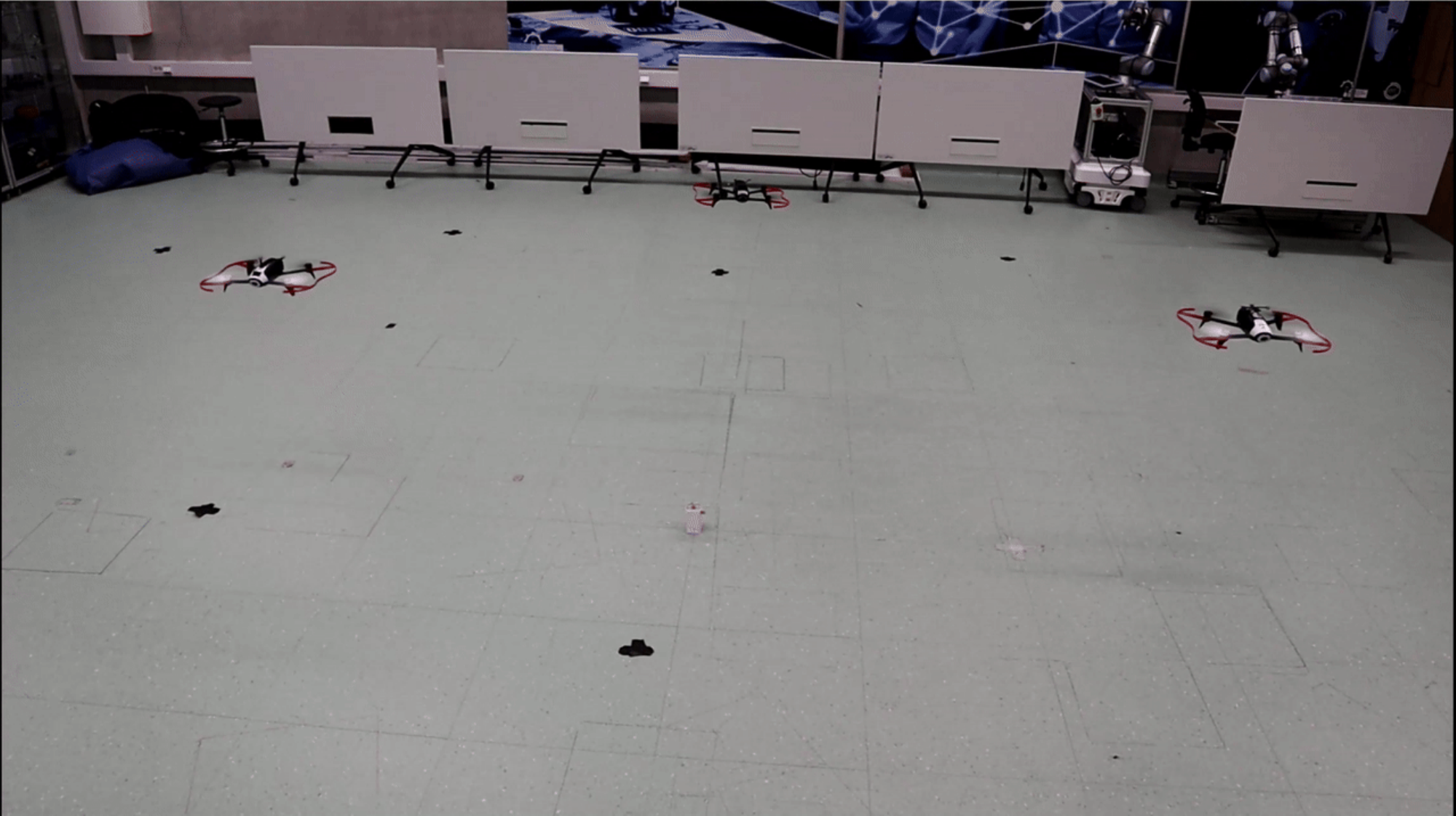}\\[3pt]
            \includegraphics[width=\linewidth]{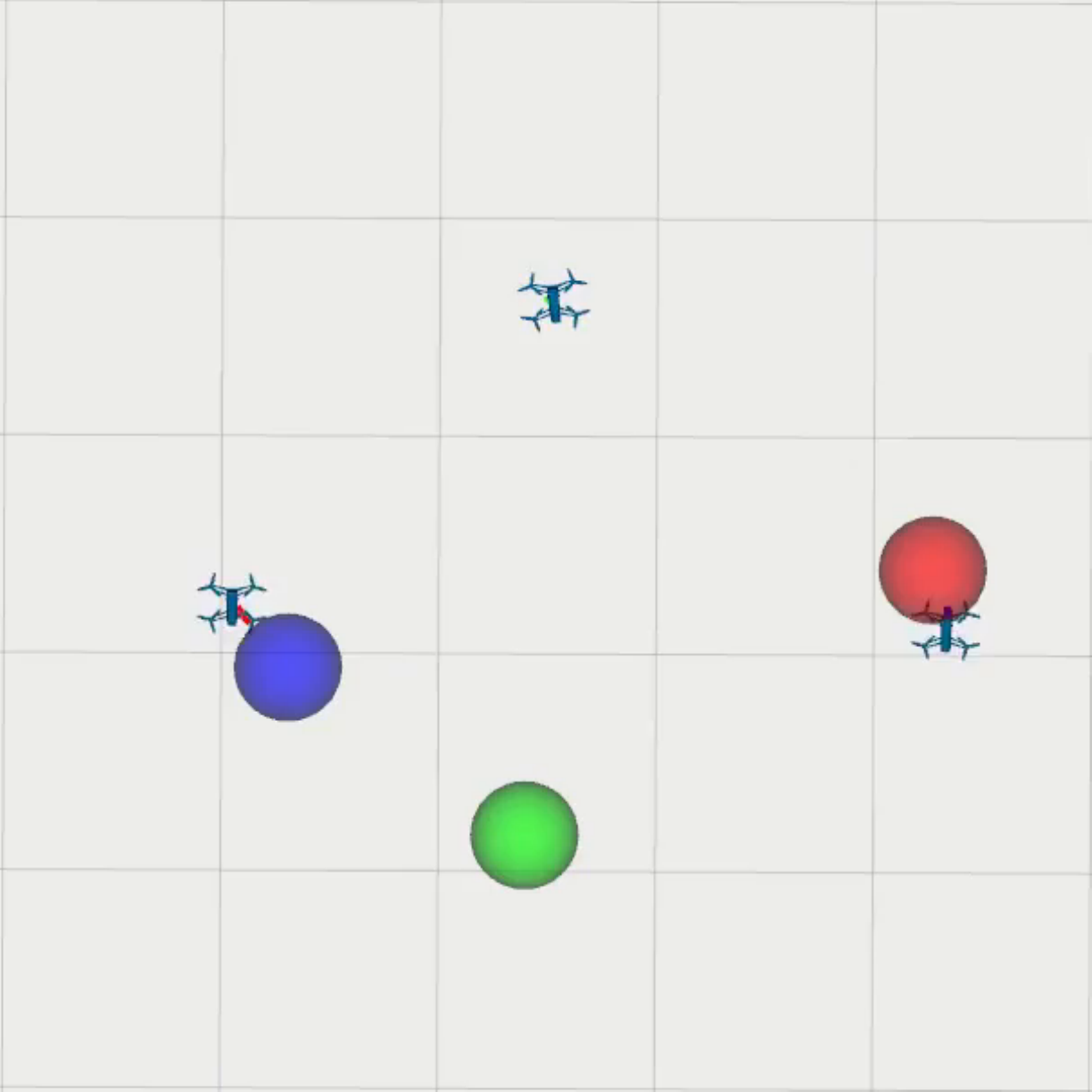}
        \end{minipage}
    }%
    \subfloat[$t=8.33$]{%
        \begin{minipage}{0.25\textwidth}
            \centering
            \includegraphics[width=\linewidth]{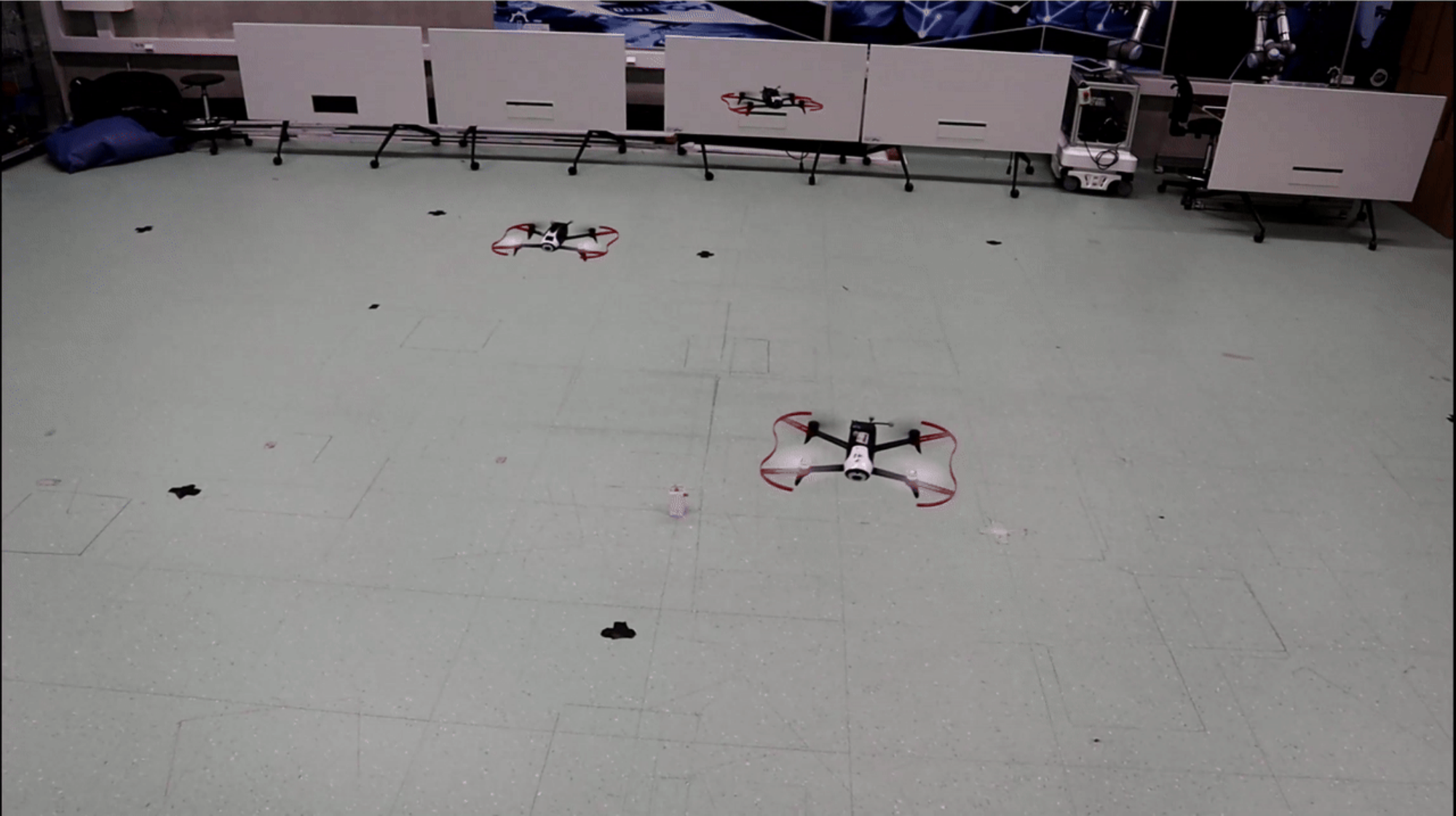}\\[3pt]
            \includegraphics[width=\linewidth]{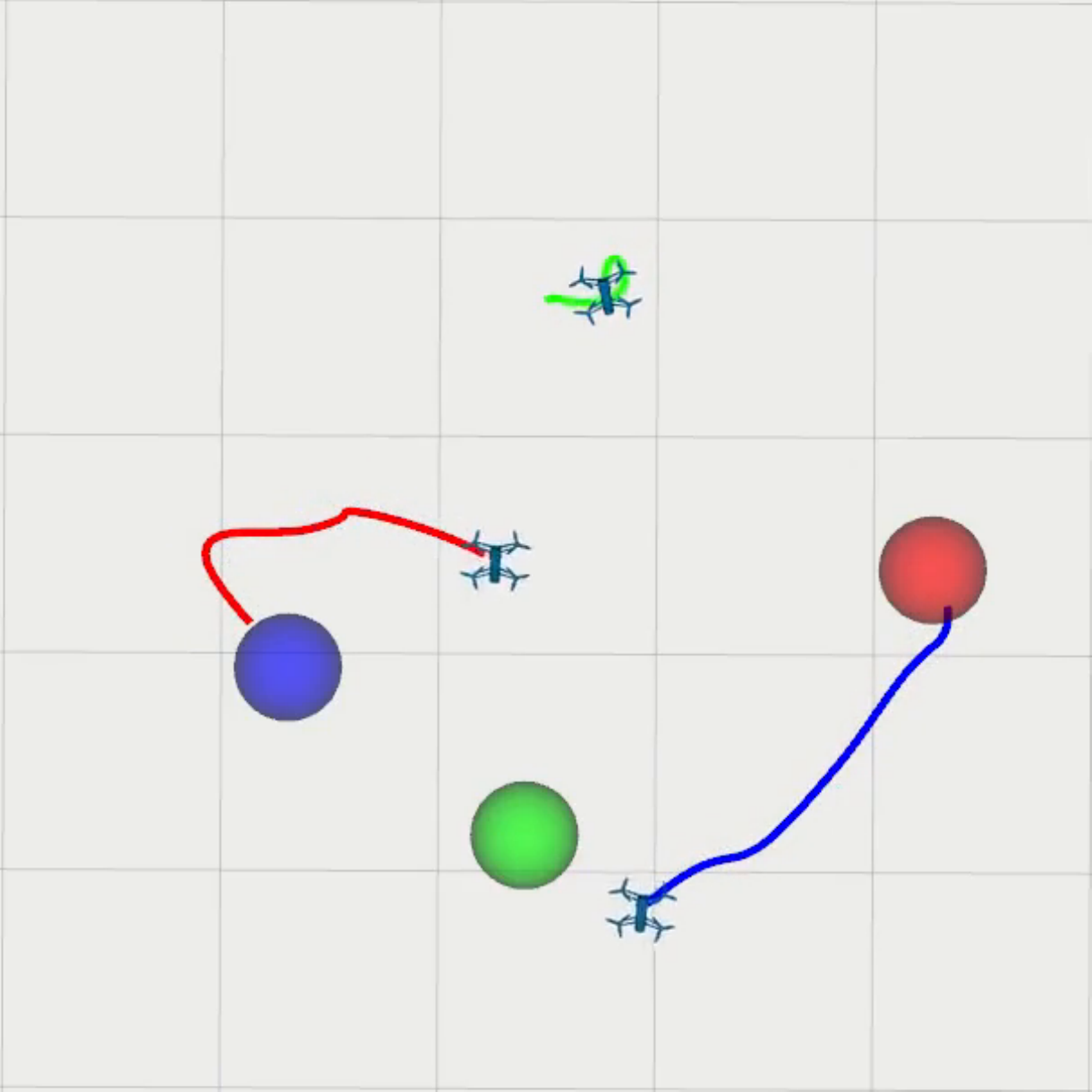}
        \end{minipage}
    }%
    \subfloat[$t=11.66$]{%
        \begin{minipage}{0.25\textwidth}
            \centering
            \includegraphics[width=\linewidth]{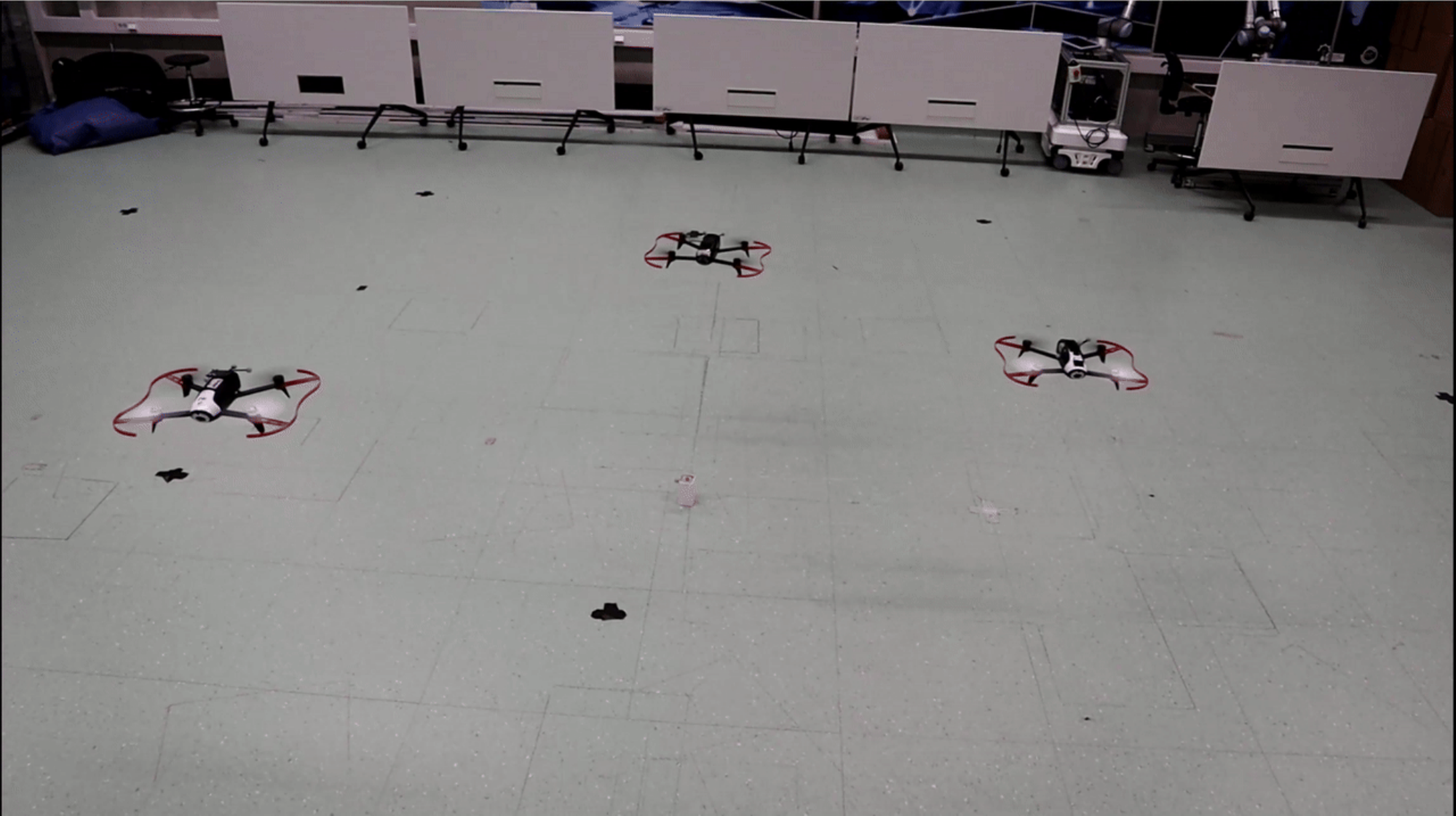}\\[3pt]
            \includegraphics[width=\linewidth]{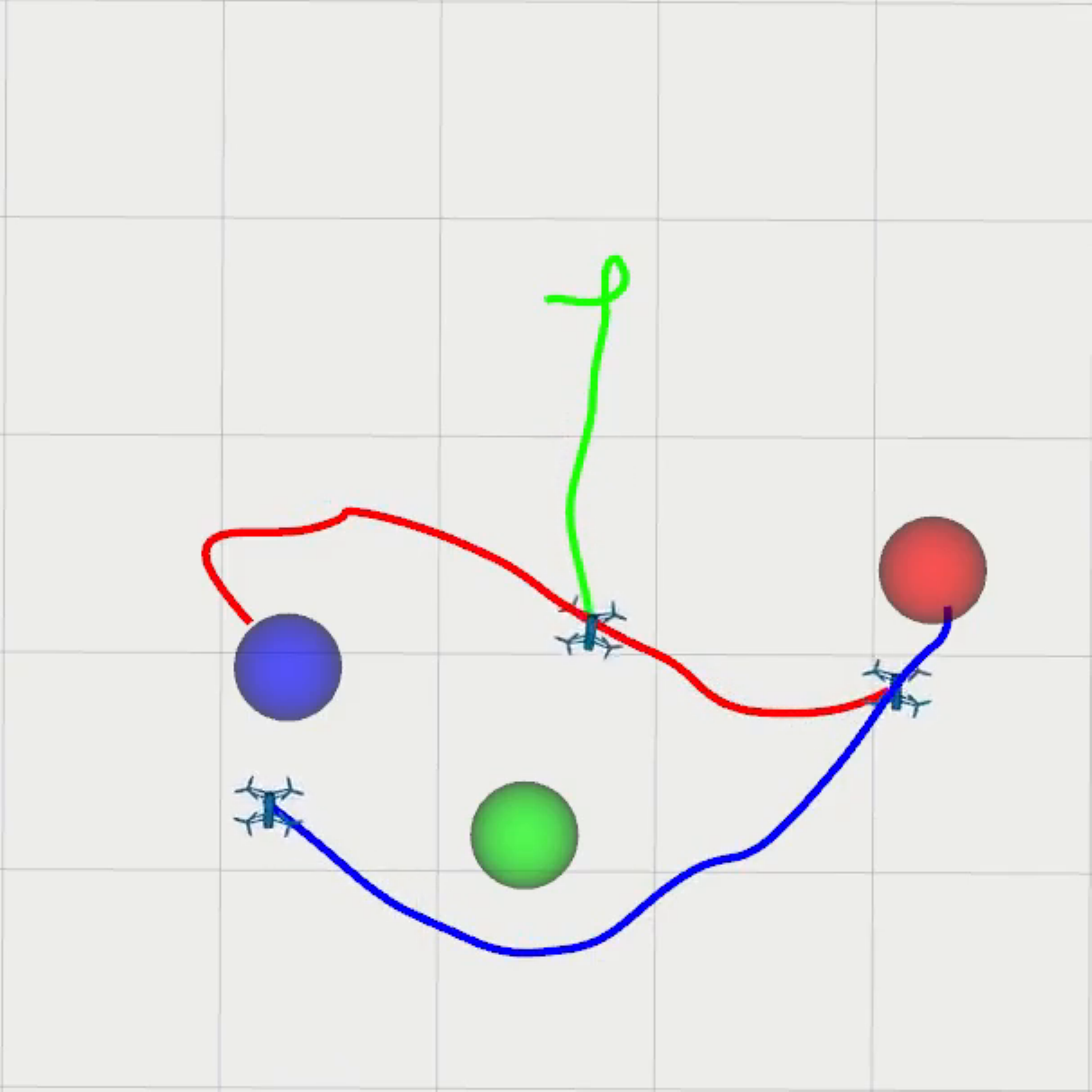}
        \end{minipage}
    }%
    \subfloat[$t=15.00$]{%
        \begin{minipage}{0.25\textwidth}
            \centering
            \includegraphics[width=\linewidth]{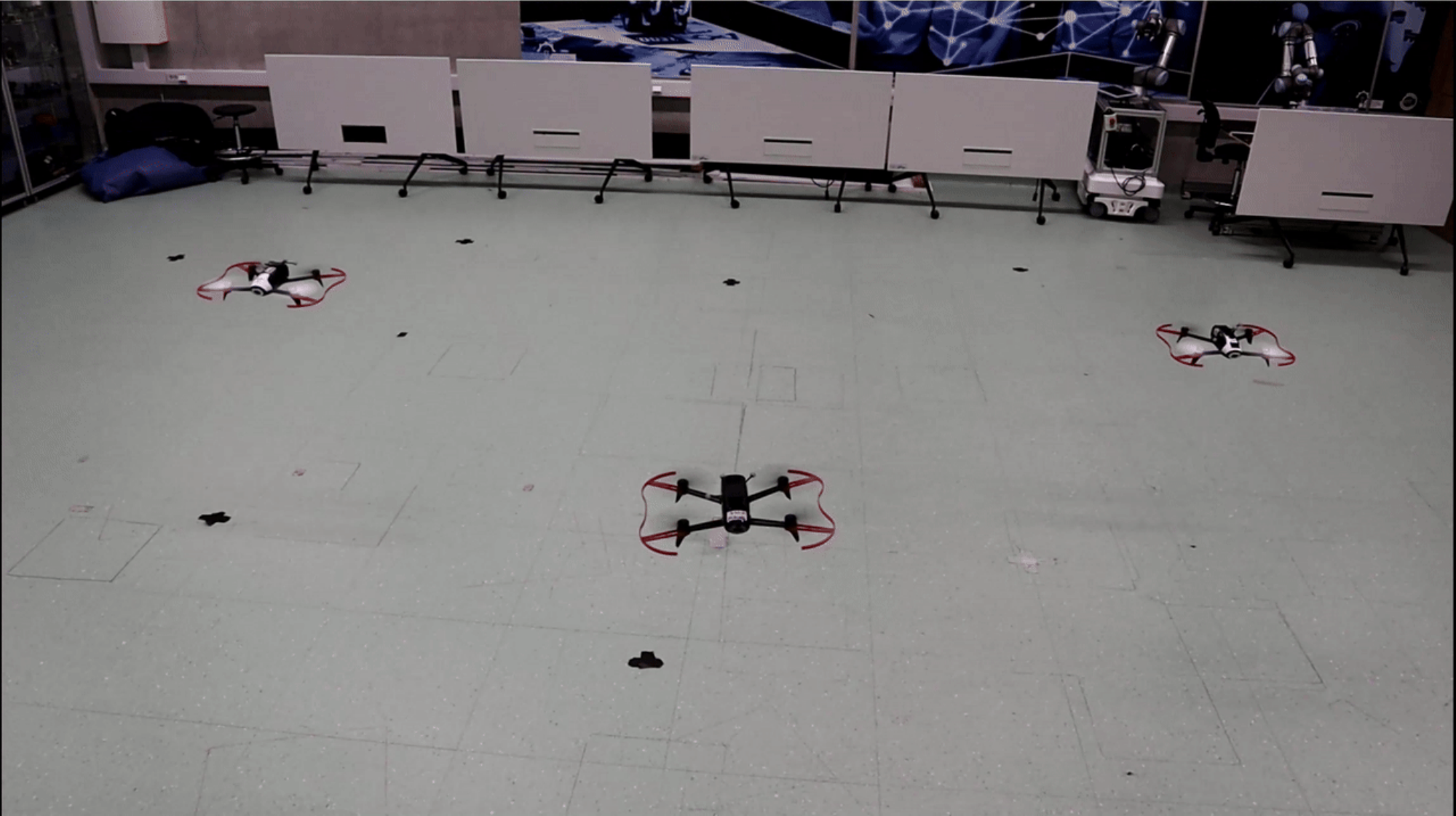}\\[3pt]
            \includegraphics[width=\linewidth]{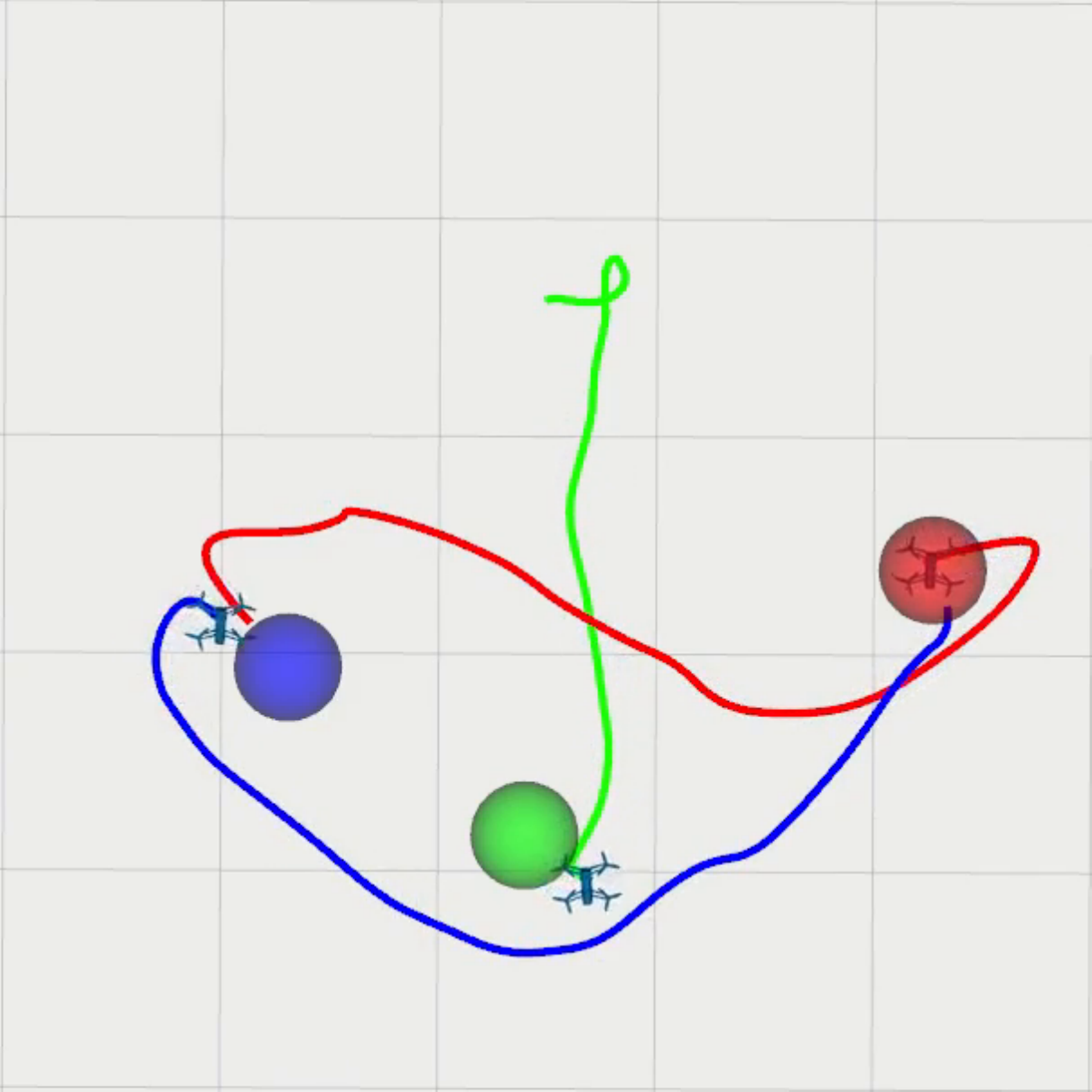}
        \end{minipage}
    }
    \caption{Snapshots of multiple Parrot Bebop quadrotors executing coordinated trajectories in the indoor motion-capture environment. Hardware (top) and trajectory visualization (bottom).}
    \label{fig:drone_columns}
\end{figure*}

\begin{figure*}[t]
    \centering
    \subfloat[$t=0.00$]{%
        \begin{minipage}{0.20\textwidth}
            \centering
            \includegraphics[width=\linewidth]{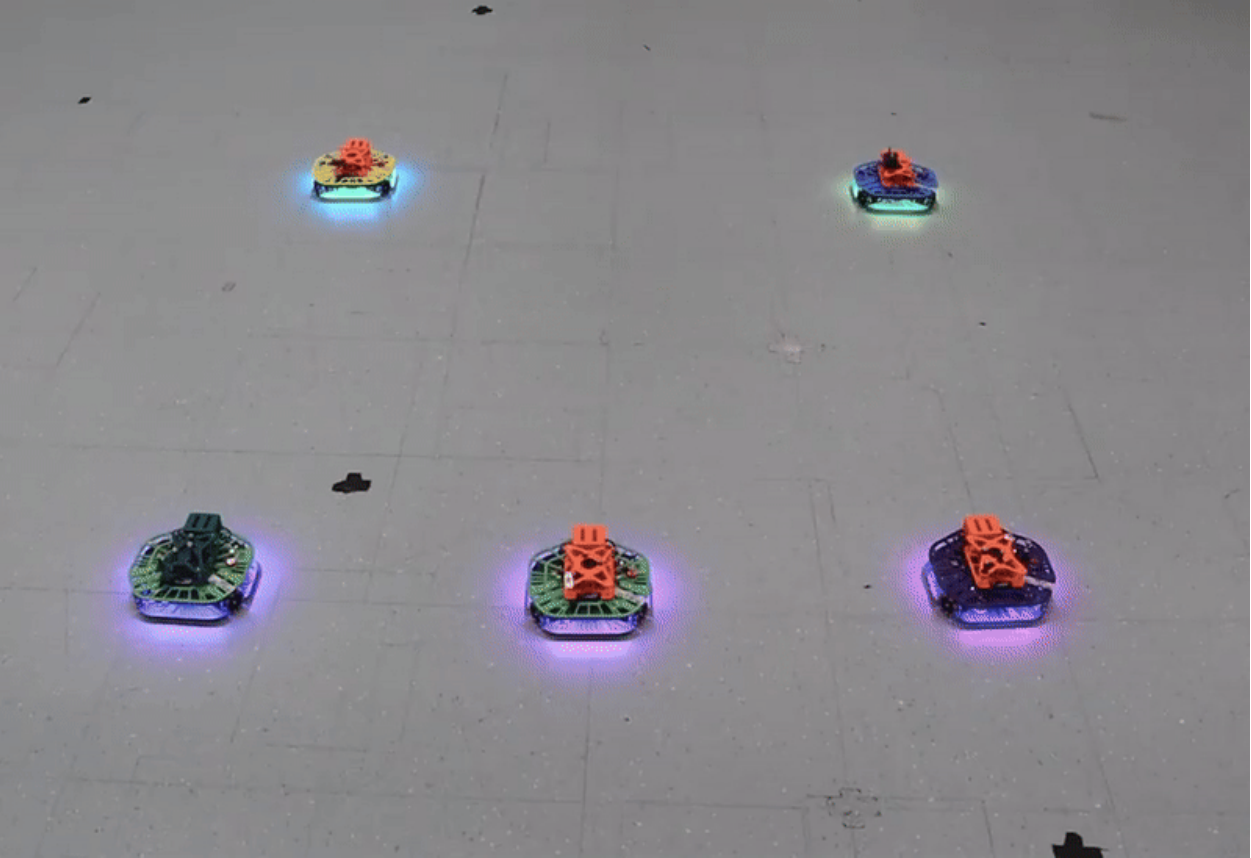}\\[3pt]
            \includegraphics[width=\linewidth]{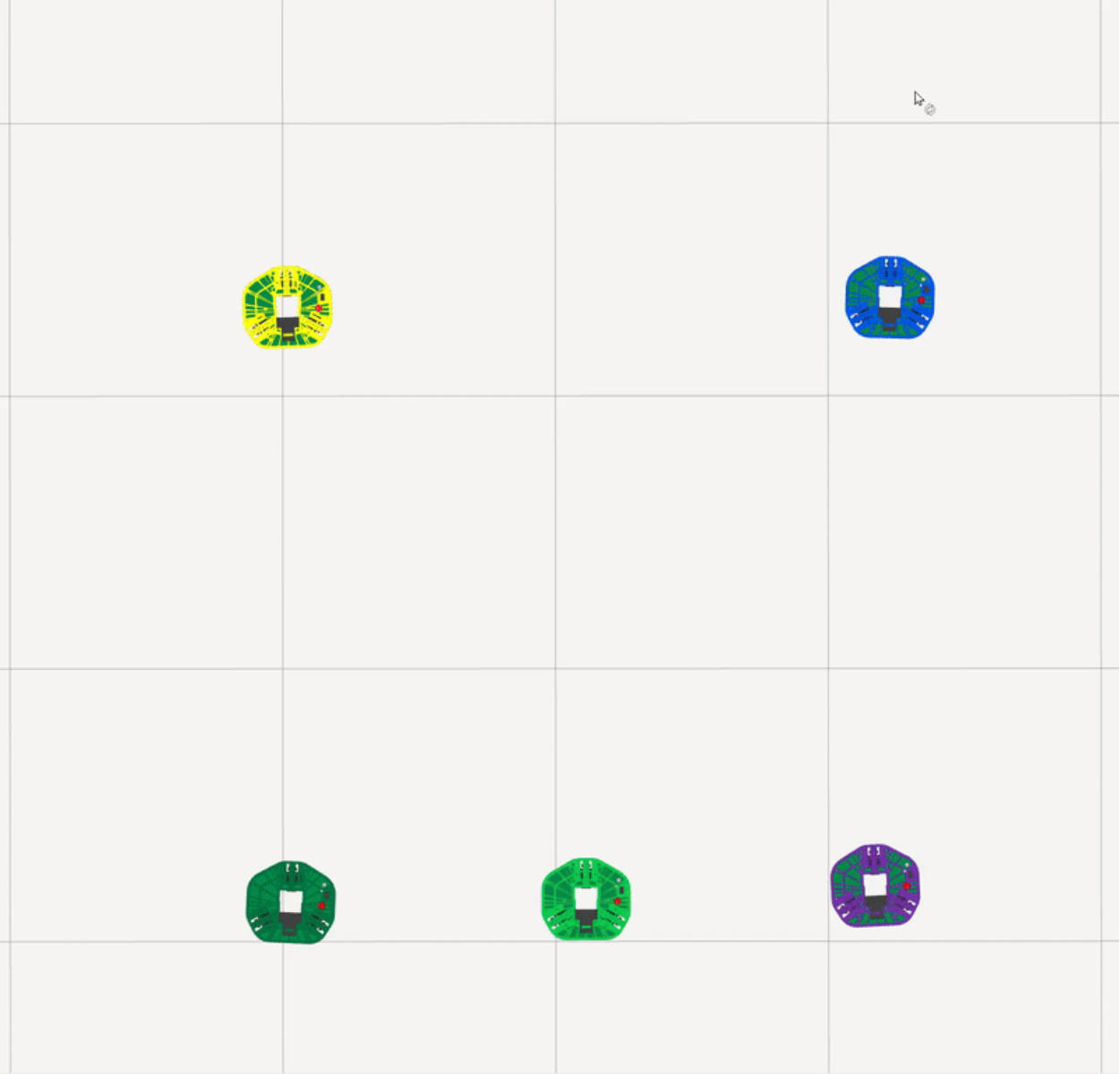}
        \end{minipage}
    }%
    \subfloat[$t=5.00$]{%
        \begin{minipage}{0.20\textwidth}
            \centering
            \includegraphics[width=\linewidth]{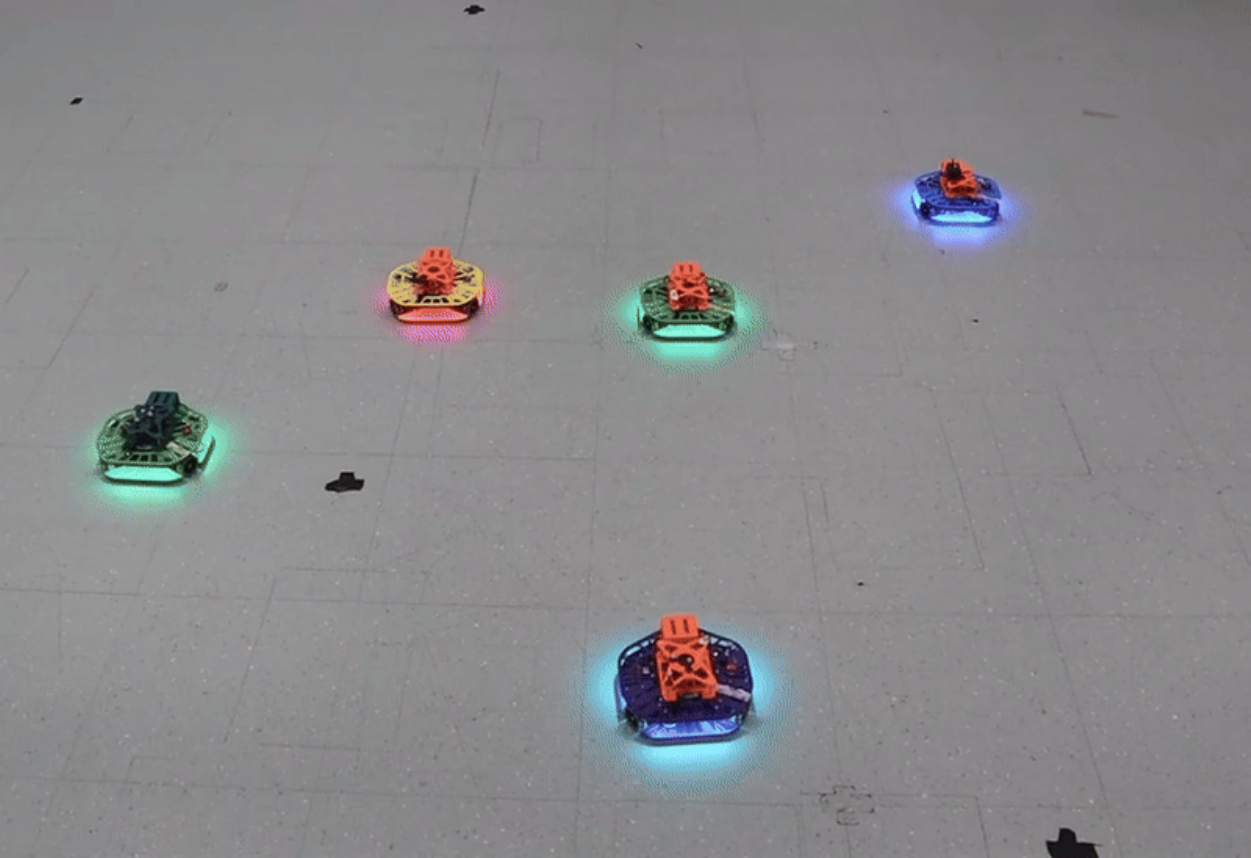}\\[3pt]
            \includegraphics[width=\linewidth]{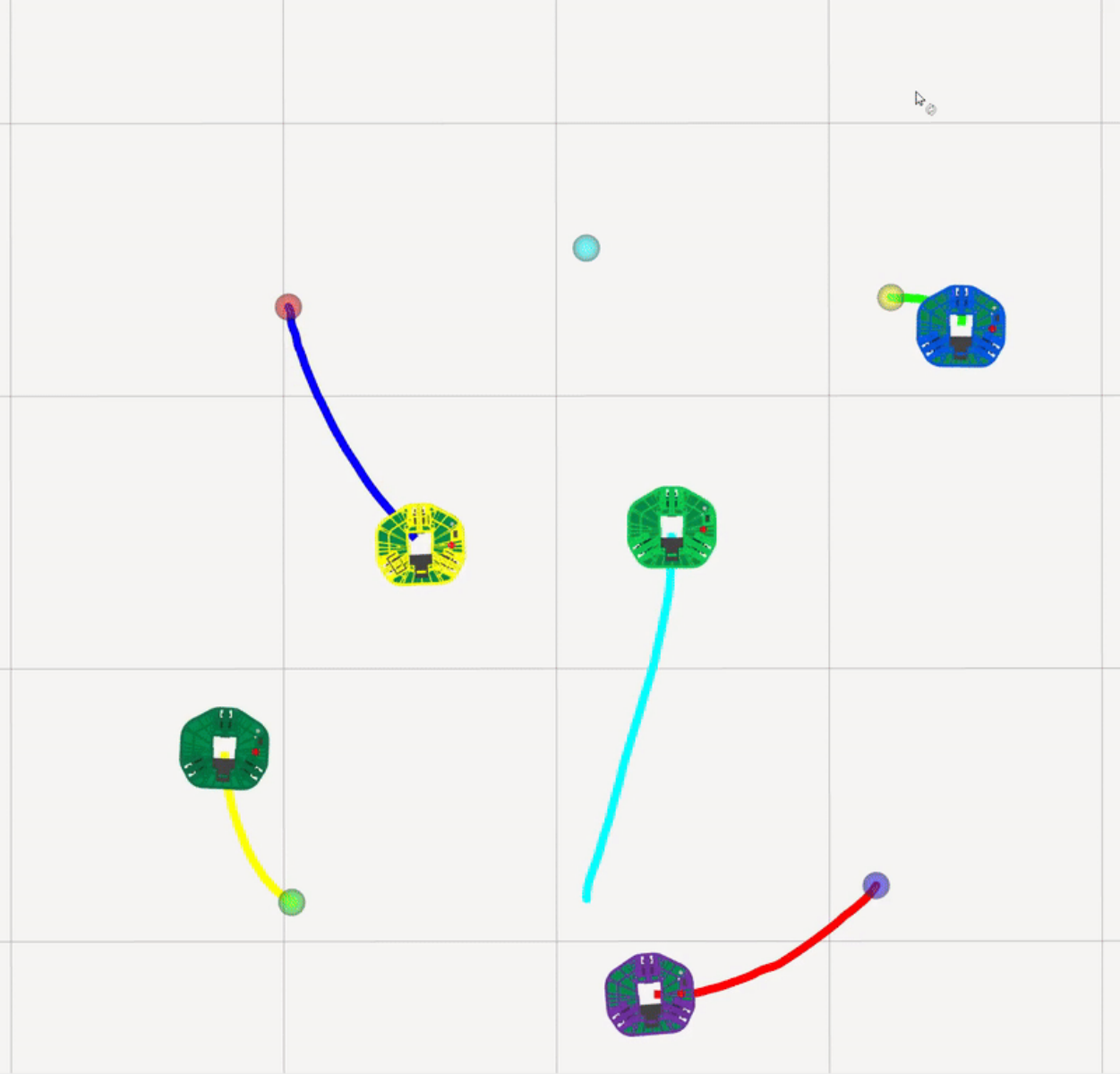}
        \end{minipage}
    }%
    \subfloat[$t=10.00$]{%
        \begin{minipage}{0.20\textwidth}
            \centering
            \includegraphics[width=\linewidth]{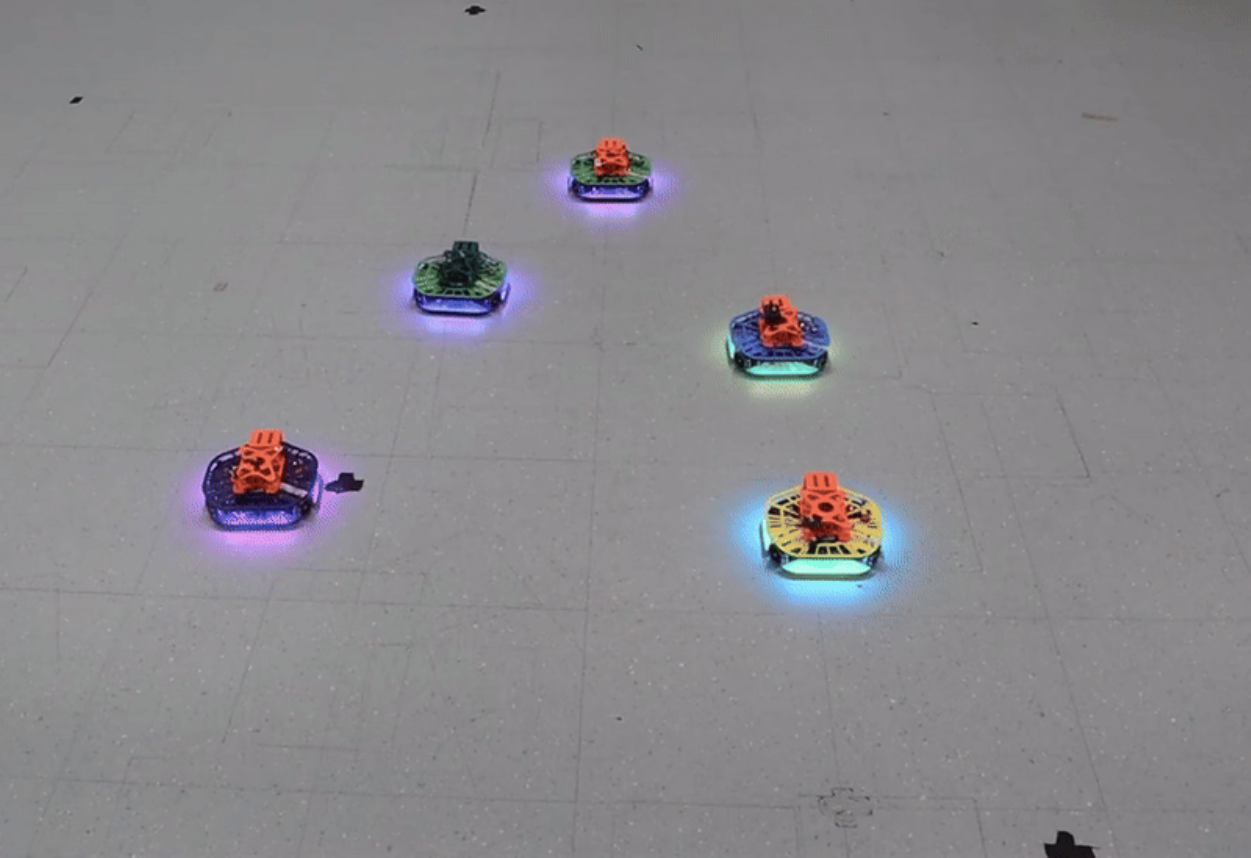}\\[3pt]
            \includegraphics[width=\linewidth]{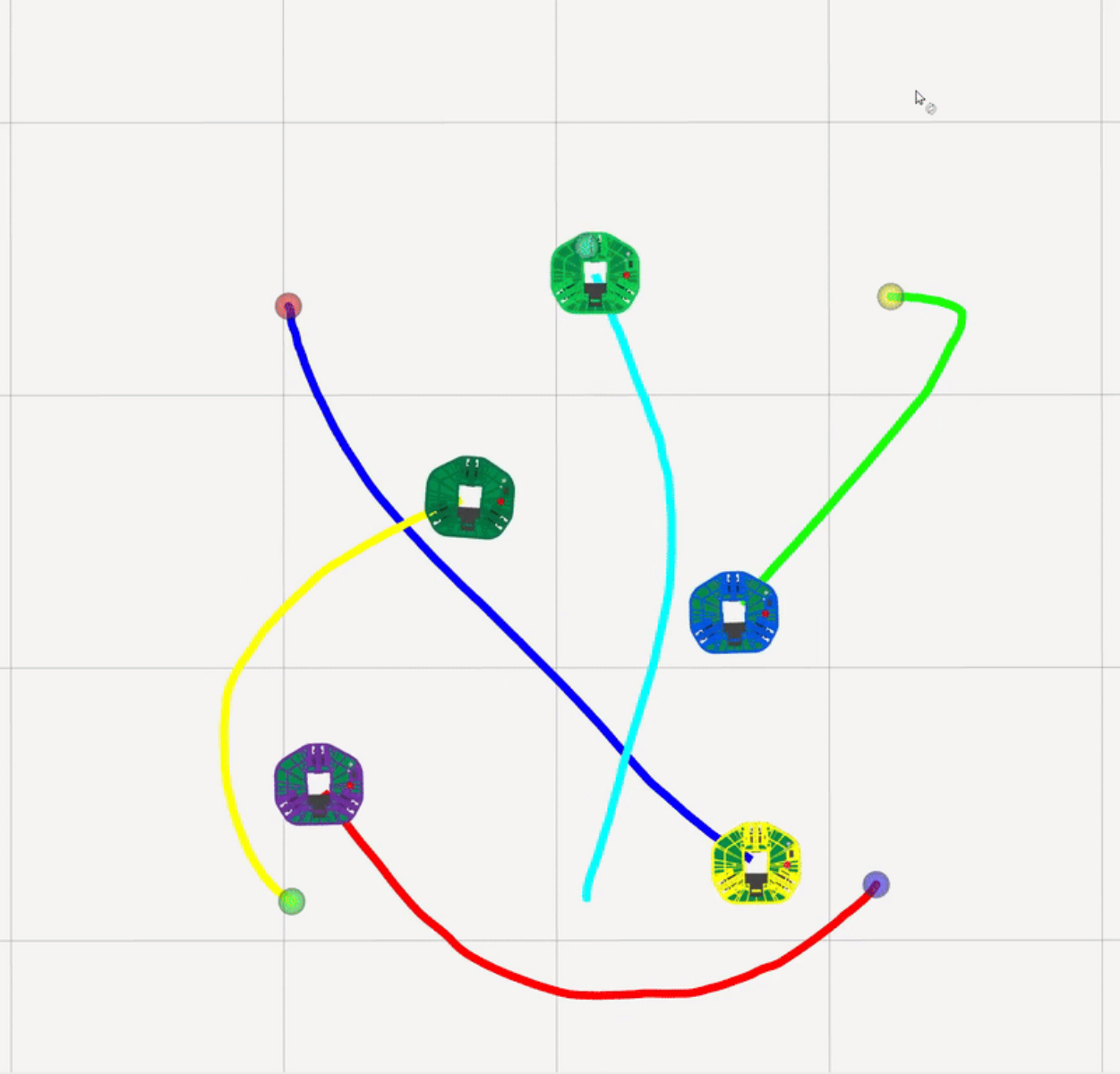}
        \end{minipage}
    }%
    \subfloat[$t=15.00$]{%
        \begin{minipage}{0.20\textwidth}
            \centering
            \includegraphics[width=\linewidth]{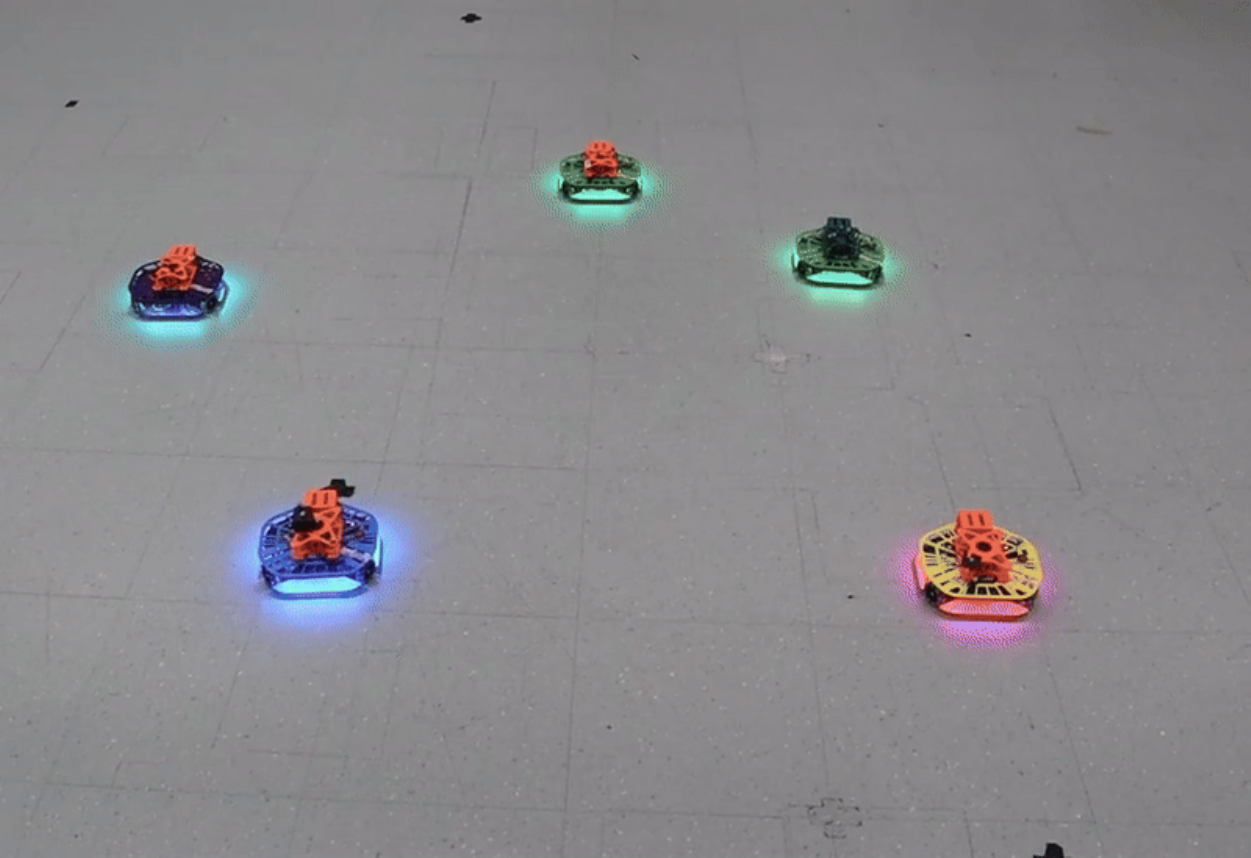}\\[3pt]
            \includegraphics[width=\linewidth]{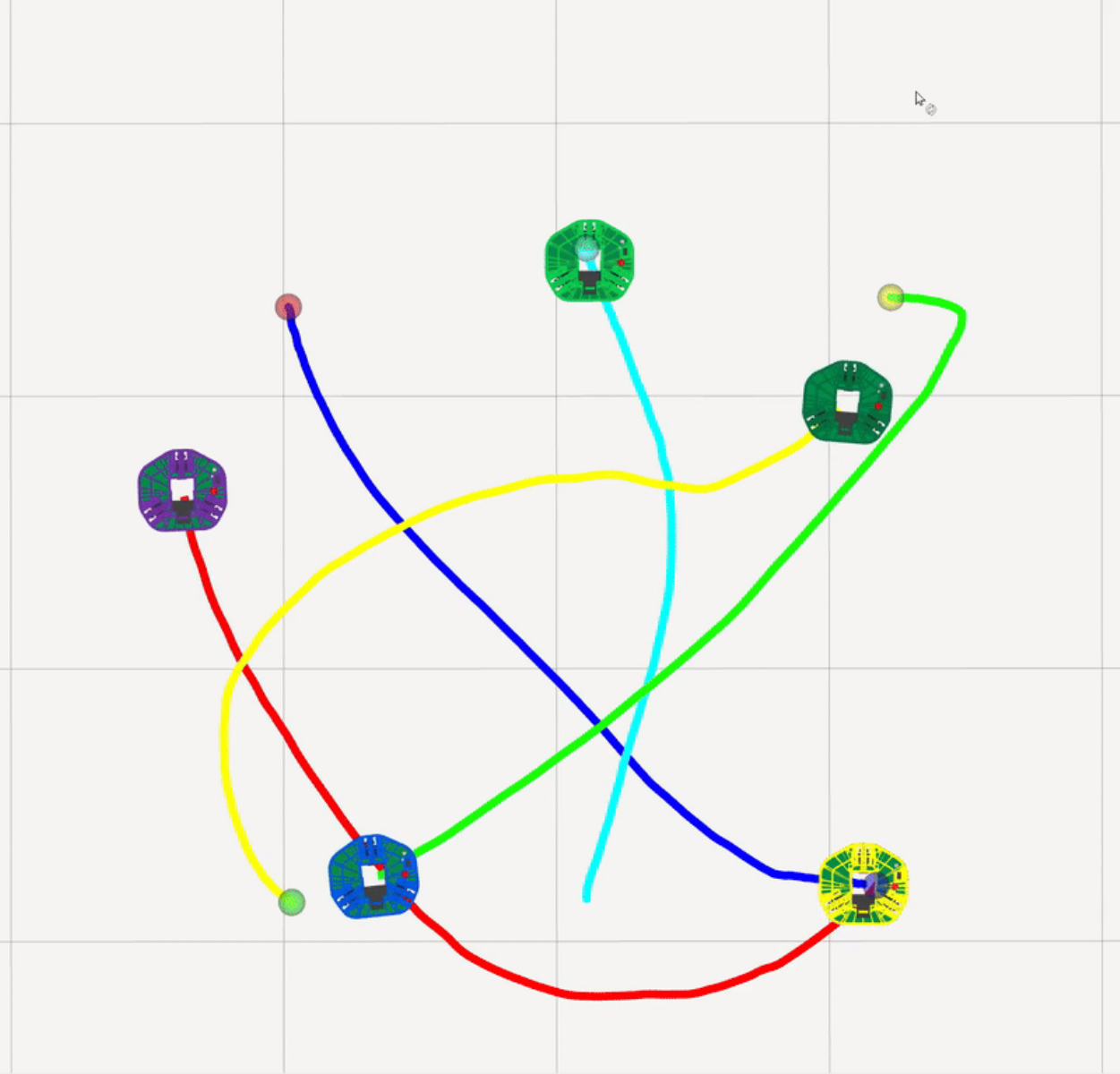}
        \end{minipage}
    }%
    \subfloat[$t=20.00$]{%
        \begin{minipage}{0.20\textwidth}
            \centering
            \includegraphics[width=\linewidth]{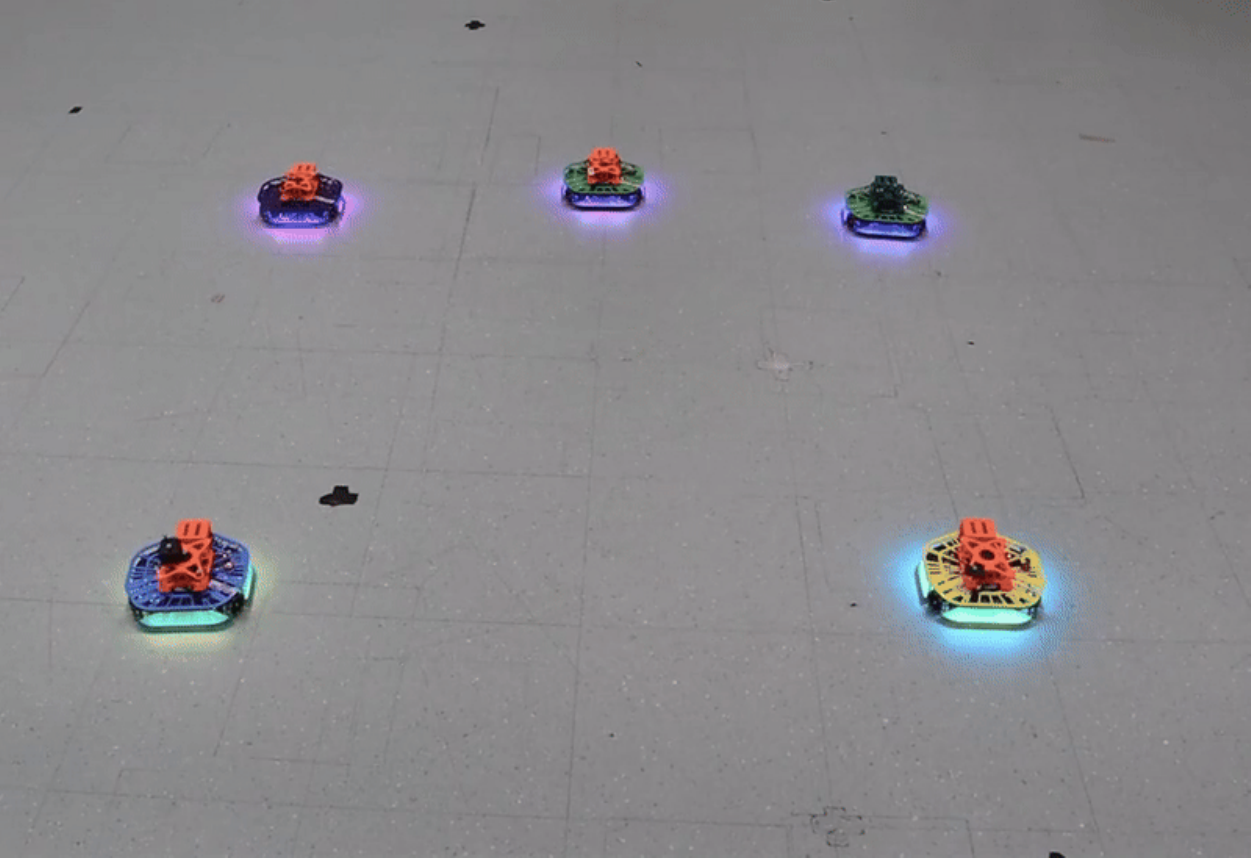}\\[3pt]
            \includegraphics[width=\linewidth]{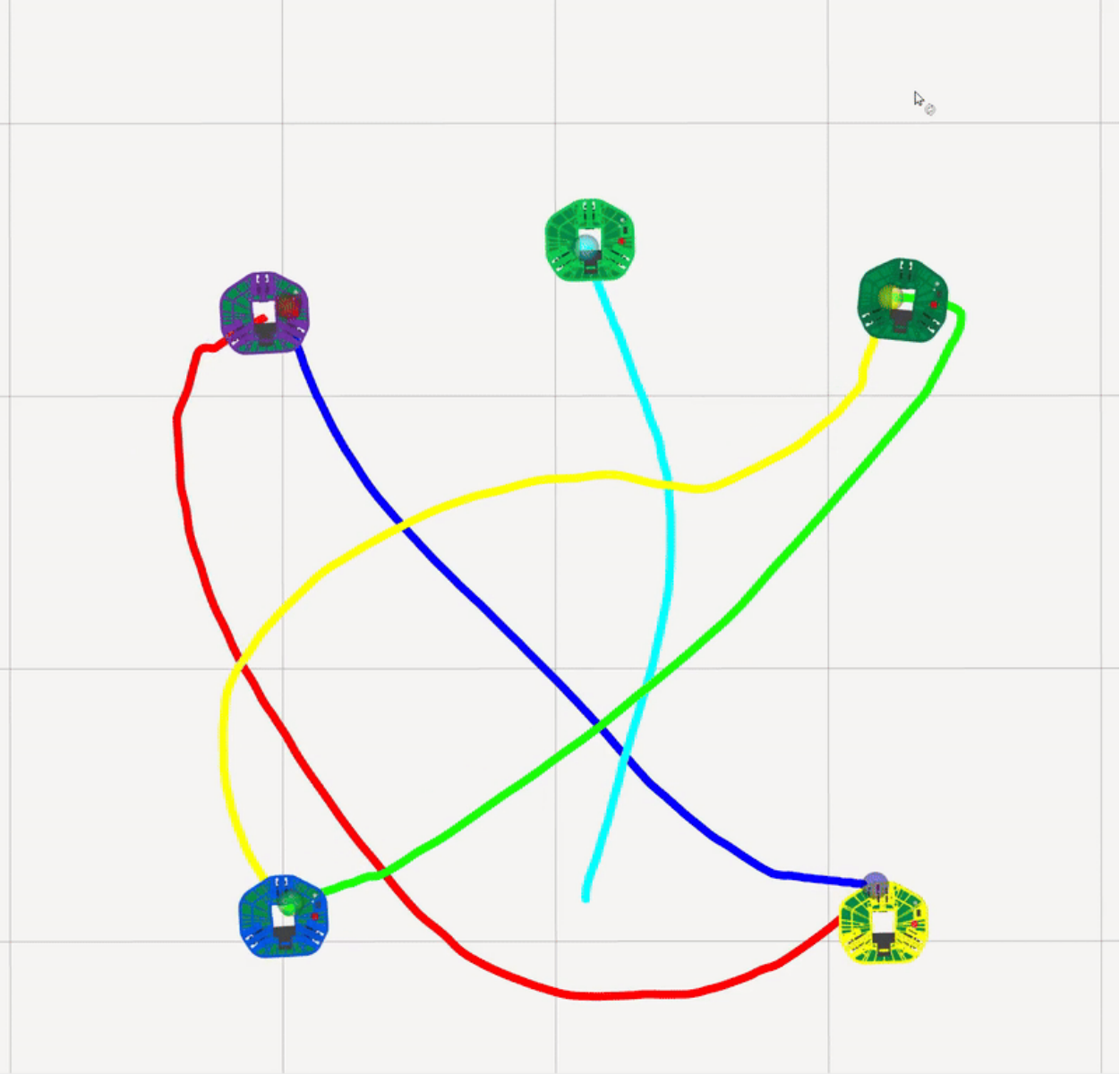}
        \end{minipage}
    }
    \caption{Snapshots of multiple Robotont \cite{motsharg2024robotont} robots performing coordinated navigation in the shared workspace. Hardware (top) and trajectory visualization (bottom).}
    \label{fig:robotonts_columns}
\end{figure*}

\textcolor{black}{We validated our approach through real-world multi-robot experiments conducted in an indoor environment equipped with an OptiTrack motion-capture system. The motion-capture setup provided real-time pose measurements for all robots, which were streamed to a central laptop executing the complete planning pipeline. Experiments were performed on two distinct hardware platforms: Parrot Bebop quadrotors for aerial evaluation (Fig.~\ref{fig:drone_columns}) and Robotont \cite{motsharg2024robotont} mobile robots for ground navigation (Fig.~\ref{fig:robotonts_columns}). All robots and the motion-capture system were connected to a single central computer via a local wireless network, where joint trajectory optimization and control command computation were carried out.}

% Paragraph 2: Method Adaptation (Velocity Conditioning & Replanning)
\textcolor{black}{For these hardware experiments, we extended the encoders of the Flow Matching network (Fig.~\ref{fig:flow_model}) and the SF initialization network (Fig.~\ref{unroll_learning}) to be additionally conditioned on the robots' initial velocities. The pipeline was retrained to generate trajectories from arbitrary initial positions and velocities to a set of goal configurations. We leveraged this capability to perform on-the-fly replanning during hardware demonstrations, which proved crucial for compensating for tracking errors, communication delays, and minor external disturbances. At each control cycle, updated position measurements were used to estimate current velocities, after which the centralized planner generated refined trajectories and transmitted velocity commands to the robots.}

% Paragraph 3: Results & Generalization
Across multiple trials on both platforms, the robots successfully executed coordinated motions within a shared workspace without collisions. The system consistently maintained smooth trajectories while satisfying inter-robot separation constraints, demonstrating that Flow-Opt can be reliably deployed on physical robotic systems and generalizes effectively across heterogeneous platforms. \textcolor{revision_color}{A supplementary video included in the additional material provides representative hardware demonstrations and additional qualitative results.}

% Our initialization network (Fig.\ref{unroll_learning}) for SF is conditioned on a fixed number of obstacles. But can the resulting performance withstand minor perturbation in problem setting at inference time. To test this, we considered several problem instances where we added an additional obstacle that was not present during training time (Fig.\ref{extra_obs_results}(a)-(b)). As can be seen, our SF solver is robust to such test-time perturbations and can successfully produce smooth and feasible trajectories. Fig.\ref{extra_obs_results}(c)-(d) further reinforces our assertion about SF robustness as it shows a fast decay of primal and fixed-point residual.

\section{Conclusions and Future Work}

Centralized trajectory optimization offers unparalleled flexibility to coordinate robot swarms. This class of methods can directly optimize for global objectives, such as minimizing the total path length traveled by the entire swarm or reducing overall control effort (e.g., acceleration and velocity). Moreover, they can serve as rich expert datasets for training decentralized controllers \cite{prorok2022holy}. Despite these advantages, the adoption of centralized methods has been severely hampered by a significant barrier: their lack of scalability. This paper introduces a novel framework that challenges this long-standing perception.

We introduced Flow-Opt, a framework that synergistically combines flow matching with differentiable optimization to enable highly scalable centralized trajectory planning for multi-robot systems. Our approach demonstrates significant advancements over existing methods across three key dimensions: (1) Computational Efficiency: Flow-Opt generates trajectories for dozens of robots in tens of milliseconds, achieving speed-ups of up to 30× over optimization-based baselines \cite{Rastgar2020GPUAC, park2020efficient} and 160× over diffusion-based approaches. (2) Solution Quality: The framework produces smoother trajectories with greater inter-robot clearance, all while rigorously satisfying system constraints. (3) Scalability: Our architecture leverages batched GPU operations to process multiple planning problems in parallel. At the heart of our approach is a flow matching model, built upon a diffusion transformer backbone, which allows it to effectively capture complex multi-robot interactions. This generative model is complemented by a differentiable safety filter that ensures rapid convergence to dynamically feasible and collision-free solutions. Extensive experiments across diverse 2D and 3D scenarios validate that Flow-Opt maintains robustness in cluttered environments and generalizes effectively to out-of-distribution challenges.

\noindent \textcolor{black}{\textbf{Real World Deployability:} Our work is highly suited for coordinating robots in structured environments like a warehouse, where it is possible to employ extra GPU compute to improve throughput. For smaller swarm size, our work can also run on embedded hardware making it suitable for tasks like coordinated drone cinematography \cite{di2023cooperative}. Our work also opens door for generating crowd simulations in a scalable manner.}

Our work opens several promising avenues for future research. First, we are exploring an end-to-end joint training of the flow policy and the Safety Filter (SF), which we hypothesize could further enhance performance. However, this direction requires fundamental modifications to the flow training methodology, which is an active area of our current investigation. 
\textcolor{black}{One limitation of our approach is that in its current form, it is suitable for robots represented as a series of integrators}. Thus, we plan to extend our framework to accommodate kinematically constrained, non-holonomic robots. The domain of autonomous driving and connected vehicles, in particular, presents a compelling and natural application for this extension, offering a clear path toward real-world impact.

% Future work will explore more extensions to dynamic environments and integration with decentralized coordination strategies.

%%%%%%%%%%%%%%%%%%%%%%%%%%%%%%%%%%%%%%%%%%%%%%%%%%%%%%%%%%%%%%%%%%%%%%%%%%%%%%%%%%%%%%%%%%%%%%%%%%%%%%%%%%%%%%%%%%%%%%%%%%%%%%%%%%%%%%%%%%%%%%%%%%%%%%%%%%%%%%%

\section{Appendix}\label{Appendix}
\subsection{Derivation of Fixed-Point Iteration Form}

The following derivation builds upon \cite{Rastgar2020GPUAC} but improves it in the following manner.
\begin{itemize}
    \item Inclusion of obstacle avoidance and workspace constraints without disturbing the underlying numerical structure.
    \item Exposing the optimizer steps as a batchable and GPU-accelerated fixed-point iteration.
\end{itemize}

\noindent We begin by reformulating the quadratic inequalities contained in the constraint function $\mathbf{g}$.

\subsubsection{Quadratic Constraints in Spherical Form} The inter-robot collision avoidance constraints \eqref{inter_robot_con} can be re-phrased into the following spherical form.

\begin{dmath}
    \mathbf{p}_{i|k}-\mathbf{p}_{j|k} = d_{ij|k}\begin{bmatrix}
        a \cos\alpha_{ij|k}\sin\beta_{ij|k}\\  a \sin\alpha_{ij|k}\sin\beta_{ij|k} \\ b \cos\beta_{ij|k}
    \end{bmatrix}, 1\leq d_{ij|k}\leq \infty,
    \label{inter_robot_polar}
\end{dmath}

\noindent where $\alpha_{ij|k}, \beta_{ij|k}$ and $ d_{ij|k}$ are the spherical angles and normalized line-of-sight distance between the robots. These are unknown and will be obtained by the SF optimizer along with other variables. We recall that the robots are modeled as spheroids with dimension $(\frac{a}{2}, \frac{a}{2}, \frac{b}{2})$

Following a similar notation, we can rewrite the obstacle avoidance constraints \eqref{collision_con} in the following manner.
\begin{align}
    \mathbf{p}_{i|k}-\mathbf{p}_{o, m|k} = d_{o,m|k}\begin{bmatrix}
        a_o \cos\alpha_{o,m|k}\sin\beta_{o,m|k}\\  a_o \sin\alpha_{o,m|k}\sin\beta_{o,m|k} \\ b_o \cos\beta_{o,m|k}
    \end{bmatrix}, \nonumber \\ 1\leq d_{o,m|k}\leq \infty.
    \label{workspace_polar}
\end{align}
\subsubsection{Reformulated Problem}
\noindent We now get the following reformulation of the SF optimizer\eqref{proj_cost}-\eqref{proj_robot}, obtained by replacing  \eqref{proj_robot} with \eqref{reform_non_convex} derived from \eqref{inter_robot_polar}-\eqref{workspace_polar}.
\small
\begin{align}
    \min_{\overline{\boldsymbol{\xi}}, \Tilde{\boldsymbol{\alpha}}, \Tilde{\boldsymbol{\beta}}, \Tilde{\mathbf{d}}} \frac{1}{2}\left \Vert \overline{\boldsymbol{\xi}}-{\boldsymbol{\xi}}\right\Vert_2^2 \label{reform_cost}\\
    \mathbf{A}\overline{\boldsymbol{\xi}} = \mathbf{b} \label{reform_eq}\\
    \mathbf{G}\overline{\boldsymbol{\xi}} \leq \mathbf{h}\\
    \mathbf{F}\overline{\boldsymbol{\xi}} = \mathbf{e}(\Tilde{\boldsymbol{\alpha}},\Tilde{\boldsymbol{\beta}}, \Tilde{\mathbf{d}})\label{reform_non_convex}  \\
    \mathbf{d}_{min}\leq \Tilde{\mathbf{d}} \leq \mathbf{d}_{max}
\end{align}
\normalsize

 \small
  \begin{align}
      {\textbf{F}} =\hspace{-0.15cm} \begin{bmatrix}
        \begin{bmatrix}
        \mathbf{F}_{r} \\
        \mathbf{F}_o 
    \end{bmatrix} \hspace{-0.25cm}&\hspace{-0.25cm} \textbf{0} \hspace{-0.25cm}&\hspace{-0.25cm} \textbf{0} \\ \textbf{0}\hspace{-0.25cm} & \hspace{-0.25cm}\hspace{-0.15cm}\begin{bmatrix}
         \mathbf{F}_{r} \\
        \mathbf{F}_o 
    \end{bmatrix} \hspace{-0.25cm}& \hspace{-0.25cm} \textbf{0}\\
         \textbf{0}\hspace{-0.25cm}& \hspace{-0.25cm}\textbf{0} \hspace{-0.25cm}&\hspace{-0.25cm} \begin{bmatrix}
         \mathbf{F}_{r} \\
        \mathbf{F}_o 
    \end{bmatrix}
    \end{bmatrix}\hspace{-0.15cm}, 
    {\mathbf{e}} = \hspace{-0.15cm}\begin{bmatrix}
      a\mathbf{d}\cos{ \boldsymbol{\alpha}} \sin{ \boldsymbol{\beta}} \\ 
      \mathbf{x}_o+a_o\mathbf{d}_o\cos{ \boldsymbol{\alpha}_o} \sin{ \boldsymbol{\beta}_o} \\ 
      a\mathbf{d}\sin{ \boldsymbol{\alpha}} \sin{ \boldsymbol{\beta}} \\
      \mathbf{y}_o+a_o\mathbf{d}_o\sin{ \boldsymbol{\alpha}_o} \sin{ \boldsymbol{\beta}_o}\\
      b\hspace{0.1cm} \mathbf{d}\cos{ \boldsymbol{\beta}} \\
      \mathbf{z}_o+b_o\mathbf{d}_o\cos{ \boldsymbol{\beta}_o}    
      \end{bmatrix}\hspace{-0.15cm},\label{f_e}
  \end{align} 
  \normalsize

\noindent $\Tilde{\boldsymbol{\alpha}} = (\boldsymbol{\alpha}, \boldsymbol{\alpha}_o)$, $\Tilde{\boldsymbol{\beta}} = (\boldsymbol{\beta}, \boldsymbol{\beta}_o)$, $\Tilde{\mathbf{d}} = (\mathbf{d}, \mathbf{d}_o)$. The $\boldsymbol{\alpha}$ is formed by stacking $\alpha_{ij|k}$ for all robot pairs $(i, j)$ and all time step $k$. Similarly, $\boldsymbol{\alpha}_o$ is formed by stacking $\alpha_{o, m|k}$ for all $m$ and $k$. We follow similar construction for  $\boldsymbol{\beta} $, $\boldsymbol{\beta}_o$, $\mathbf{d}$, and $\mathbf{d}_o$. Let $(x_{o, m|k}, y_{o, m|k}, z_{o, m|k} )$ be the components of obstacle position $\mathbf{p}_{o, m|k}$. Then, $\mathbf{x}_o$, $\mathbf{y}_o$, $\mathbf{z}_o$  are formed by stacking the respective values for all $m$ and $k$.

\small
\begin{align}
    \mathbf{F}_r = \begin{bmatrix}
    \mathbf{F}_{r, 1} \\
    \mathbf{F}_{r, 2}\\
    \vdots\\
    \mathbf{F}_{r, n-1}
  \end{bmatrix}\otimes \mathbf{P}, \qquad \mathbf{F}_o = \begin{bmatrix}
    \mathbf{P} & & \\
    & \ddots & \\
    & & \mathbf{P}
  \end{bmatrix}
\end{align}
\normalsize

\small
\begin{align}
    \mathbf{F}_{r, i} = \begin{bmatrix}
        \mathbf{F}_i & -\mathbf{I}
    \end{bmatrix}, \mathbf{F}_i = \begin{bmatrix}
        \mathbf{0}_{n-i\times 1} & \mathbf{0}_{n-i\times 1} &\dots &{\mathbf{1}_{n-i\times 1}}
    \end{bmatrix}_{n-i\times i}
\end{align}
\normalsize

\noindent The matrix $\mathbf{F}_o$ is a block-diagonal matrix with a number of blocks equal to the number of robots. The matrix $\mathbf{F}_i$ is formed with $i-1$ blocks of $n-i$ length zero vector and a single vector of ones at the $i^{th}$ column, where $n$ is the number of robots. The symbol $\otimes$ represents the Kronecker product. \textcolor{black}{The matrix $\mathbf{F}_o$ corresponds to obstacle avoidance for each robot. It is formed by vertically stacking the polynomial basis matrix $\mathbf{W}$ as many times as the number of obstacles.}

\subsubsection{Solution Process} We relax the non-convex equality \eqref{reform_non_convex} and affine inequality constraints as penalties and augment them into the cost function using the Augmented Lagrangian method 
\begin{dmath}
    \mathcal{L} = \frac{1}{2}\left \Vert \overline{\boldsymbol{\xi}}-{\boldsymbol{\xi}}\right\Vert_2^2+\frac{\rho}{2}\left \Vert \mathbf{F}\overline{\boldsymbol{\xi}}-\mathbf{e}(\Tilde{\boldsymbol{\alpha}}, \Tilde{\boldsymbol{\beta}}, \Tilde{\mathbf{d}} )\right\Vert_2^2+\frac{\rho}{2}\left \Vert \mathbf{G}\overline{\boldsymbol{\xi}}-\mathbf{h}+\mathbf{s}\right\Vert_2^2-\langle \boldsymbol{\lambda}, \overline{\boldsymbol{\xi}}\rangle,
    \label{aug_lag}
\end{dmath}
\noindent where $\rho$ is a known constant, the variable $\boldsymbol{\lambda}$ are so-called Lagrange multipliers and $\mathbf{s}$ is an unknown positive slack variable. We minimize \eqref{aug_lag} subject to \eqref{reform_eq} through an Alternating Minimization (AM) approach, wherein at each step, only one variable group among $\overline{\boldsymbol{\xi}}, \Tilde{\boldsymbol{\alpha}}, \Tilde{\boldsymbol{\beta}}, \Tilde{\mathbf{d}}$ is optimized while others are held fixed. Specifically, the AM routine decomposes into the following iterative steps, wherein the left superscript $l$ tracks the values of a variable across iterations. For example, ${^l}\overline{\boldsymbol{\xi}}$ is the value of $\overline{\boldsymbol{\xi}}$ at iteration $l$.

\small
\begin{subequations}
\begin{align}
    {^{l+1}}\Tilde{\boldsymbol{\alpha}} &= \argmin_{\Tilde{\boldsymbol{\alpha}}} \mathcal{L}( {^{l}}\overline{\boldsymbol{\xi}}, \Tilde{\boldsymbol{\alpha}}, {^l}\Tilde{\boldsymbol{\beta}}, {^l}\Tilde{\mathbf{d}}, {^l}\boldsymbol{\lambda} ) = f_1({^l}\overline{\boldsymbol{\xi}})\label{alpha_step}\\
    {^{l+1}}\Tilde{\boldsymbol{\beta}} &= \argmin_{\Tilde{\boldsymbol{\beta}}} \mathcal{L}({^{l}}\overline{\boldsymbol{\xi}}, {^{l+1}}\Tilde{\boldsymbol{\alpha}}, \Tilde{\boldsymbol{\beta}}, {^l}\Tilde{\mathbf{d}}, {^l}\boldsymbol{\lambda} ) = f_2({^l}\overline{\boldsymbol{\xi}})\label{beta_step}\\
    {^{l+1}}\Tilde{\mathbf{d}} &= \argmin_{\mathbf{d}_{min}\leq \Tilde{\mathbf{d}}\leq \mathbf{d}_{max}} \mathcal{L}({^{l}}\overline{\boldsymbol{\xi}}, {^{l+1}}\Tilde{\boldsymbol{\alpha}}, {^{l+1}}\Tilde{\boldsymbol{\beta}}, \Tilde{\mathbf{d}}, {^l}\boldsymbol{\lambda} ) = f_3({^l}\overline{\boldsymbol{\xi}})\label{d_step}\\
    {^{l+1 }} \mathbf{s} &= \max(0, -\mathbf{G}{^l}\overline{\boldsymbol{\xi}}-\mathbf{h}) \label{s_step} \\
    {^{l+1}}\boldsymbol{\lambda} &= {^{l}}\boldsymbol{\lambda}-\rho \mathbf{F}^T \mathbf{r}_1-\rho \mathbf{G}^T\mathbf{r}_2 \label{lambda_step} \\
    {^{l+1}}\overline{\boldsymbol{\xi}} &= \argmin_{\mathbf{A}\overline{\boldsymbol{\xi}} = \mathbf{b}} \mathcal{L}(\overline{\boldsymbol{\xi}}, {^{l+1}}\mathbf{e}, {^{l+1}}\boldsymbol{\lambda} ) \label{xi_step}\\
    &= \mathbf{M}^{-1}\boldsymbol{\eta} 
\end{align}
\end{subequations}

\begin{align}
    {^{l+1}}\mathbf{e} =   \mathbf{e}( {^{l+1}}\Tilde{\boldsymbol{\alpha}}, {^{l+1}}\Tilde{\boldsymbol{\beta}}, {^{l+1}}\Tilde{\mathbf{d}}    )
\end{align}

\small
\begin{align}
    \mathbf{r}_1 = \mathbf{F}{^l}\boldsymbol{\xi}-{^{l+1}}\mathbf{e}    ,\qquad \mathbf{r}_2 = \mathbf{G}{^l}\overline{\boldsymbol{\xi}}-\mathbf{h}+{^{l+1}}\mathbf{s} 
\end{align}
\normalsize
\normalsize
\small
\begin{align}
    \textbf{M} = \begin{bmatrix}
        \mathbf{I}+\rho\mathbf{F}^T\mathbf{F} & \textbf{A}^{T} \\ 
        \mathbf{A} & \mathbf{0}
    \end{bmatrix}^{-1} \hspace{-0.19cm}, \boldsymbol{\eta} = \begin{bmatrix}
        \rho\mathbf{F}^T \hspace{0.1cm}{^{l+1}}\mathbf{e}+{^{l+1}}\boldsymbol{\lambda}+{\boldsymbol{\xi}}\\
        \mathbf{b}
    \end{bmatrix} 
\end{align}
\normalsize

The minimization \eqref{alpha_step}-\eqref{d_step} have a closed-form solution which can be expressed as a function of ${^l}\overline{\boldsymbol{\xi}}$  \cite{Rastgar2020GPUAC}. For example, one part of minimization \eqref{alpha_step} reduces to

\begin{align}
    {^{l+1}}\alpha_{ij|k} = \arctan2({^l}y_{i|k}-{^l}y_{j|k}, {^l}x_{i|k}-{^l}x_{j|K}), \forall i, j, k  
\end{align}

\noindent where $({^l}x_{i|k}, {^l}y_{i|k})$ are the components of the position ${^l}\mathbf{p}_{i|k}$ and are completely characterized by the trajectory coefficient ${^l}\overline{\boldsymbol{\xi}}$ at iteration $l$ of the AM optimizer.

Similarly, \eqref{xi_step} is simply an equality-constrained QP and thus has an explicit formula for its solution. Moreover, since ${^{l+1}}\mathbf{e}$ and ${^{l+1}}\boldsymbol{\lambda}$ are explicit functions of ${^l}\overline{\boldsymbol{\xi}}$, ${^l}\boldsymbol{\lambda}$, \eqref{lambda_step}-\eqref{xi_step} constitutes the fixed-point iteration $\mathcal{T}$ presented in \eqref{fixed_point}.

A few points about the AM steps are worth noting. 

\begin{itemize}
    \item \textbf{Differentiability:} Since every step has a closed-form solution, we can easily unroll them into a differentiable computational graph, \textcolor{black}{as shown in Fig.\ref{unroll_learning}}.
    \item Steps \eqref{alpha_step}-\eqref{lambda_step} do not involve any matrix factorization and only require element-wise operation. Thus, they can be trivially batched across GPUs. Moreover, in step \eqref{xi_step}, the matrix $\mathbf{M}$ is independent of the input ${\boldsymbol{\xi}}$ sampled from the pre-trained flow policy. Thus, its factorizations can be pre-stored. This in turn reduces the batch version of \eqref{xi_step} to simply the following matrix-matrix product. 
    \begin{align}
        \begin{bmatrix}
         \overline{\boldsymbol{\xi}}^1\\
         \overline{\boldsymbol{\xi}}^2\\
         \overline{\boldsymbol{\xi}}^3\\
         \vdots\\
         \overline{\boldsymbol{\xi}}^n 
        \end{bmatrix} = \begin{bmatrix}
             \boldsymbol{\eta}^T(\boldsymbol{\xi}^1)\\
             \boldsymbol{\eta}^T(\boldsymbol{\xi}^2)\\
             \boldsymbol{\eta}^T(\boldsymbol{\xi}^3)\\
             \vdots\\ 
             \boldsymbol{\eta}(\boldsymbol{\xi}^n)
        \end{bmatrix}(\mathbf{M}^{-1})^T
         \label{batch_comp}
    \end{align}
    
    % \begin{align}
    %     [\overline{\boldsymbol{\xi}}^1| \overline{\boldsymbol{\xi}}^2| \overline{\boldsymbol{\xi}}^2| \cdots ] =[ \boldsymbol{\eta}(\boldsymbol{\xi}^1)| \boldsymbol{\eta}(\boldsymbol{\xi}^2)| \boldsymbol{\eta}(\boldsymbol{\xi}^3), \cdots ] (\mathbf{M}^{-1})^T
    %     \label{batch_comp}
    % \end{align}
    where $\overline{\boldsymbol{\xi}}^i$ is the projected output corresponding to flow input ${\boldsymbol{\xi}}^i$. 
\end{itemize}

% First, since every step has a closed-form solution, we can easily unroll them into a differentiable computational graph. Second, steps \eqref{alpha_step}-\eqref{lambda_step} do not involve any matrix factorization and only require element-wise operation. Thus, they can be trivially batched across GPUs. Moreover, in step \eqref{xi_step}, the matrix $\mathbf{M}$ is independent of the input $\overline{\boldsymbol{\xi}}$ sampled from the VQ-VAE/CVAE. Thus, its factorizations can be pre-stored. This also implies that the batched version of minimization \eqref{xi_step} only requires matrix-matrix products, which is easy to compute over GPU

\noindent \textbf{Primal Residual:} The primal residual vector $\mathbf{r}$ at $l^{th}$ iteration is given by the following.
\begin{align}
    \mathbf{r}_p = \mathbf{r}_1+\mathbf{r}_2
    \label{primal_residual}
\end{align}
\noindent Essentially, $\mathbf{r}_p$ dictates how well the non-convex equality constraints are satisfied. It is easy to see that $\Vert\mathbf{r}_p\Vert_2 = 0$ implies that our reformulation \eqref{inter_robot_polar}-\eqref{workspace_polar} holds and the original inter-robot \eqref{inter_robot_con}, obstacle avoidance  \eqref{collision_con} and workspace constraints \eqref{workspace_con} are satisfied.

\textcolor{black}{
\begin{remark}
    The matrix factorization of $\mathbf{M}$ would have to be recomputed if we change the penalty $\rho$, or the planning horizon (which in turn changes polynomial basis-matrix $\mathbf{W}$, recall \eqref{eq::poly}).
\end{remark}}

%====================================================================================================================================================================================
\subsection{Differences between the Train and Test Set}\label{appendix:train_test_novelty}

\noindent In this subsection, we verify that our training and test set, although drawn from the same distribution, are indeed quite different. In other words, our test set is sufficiently novel and the results presented in the earlier sections, indeed show generalization and not overfitting.

% \textcolor{revision_color}{To empirically verify that test scenarios are genuinely novel despite sharing the same generative distribution as the training set, we provide two complementary analyses. This argument applies uniformly across all experimental configurations (varying numbers of robots and obstacle counts); the figures below illustrate the analysis for the 16-robot case, which is representative of the full dataset.}

\begin{figure}[!t]
    \centering
    \includegraphics[width=\columnwidth]{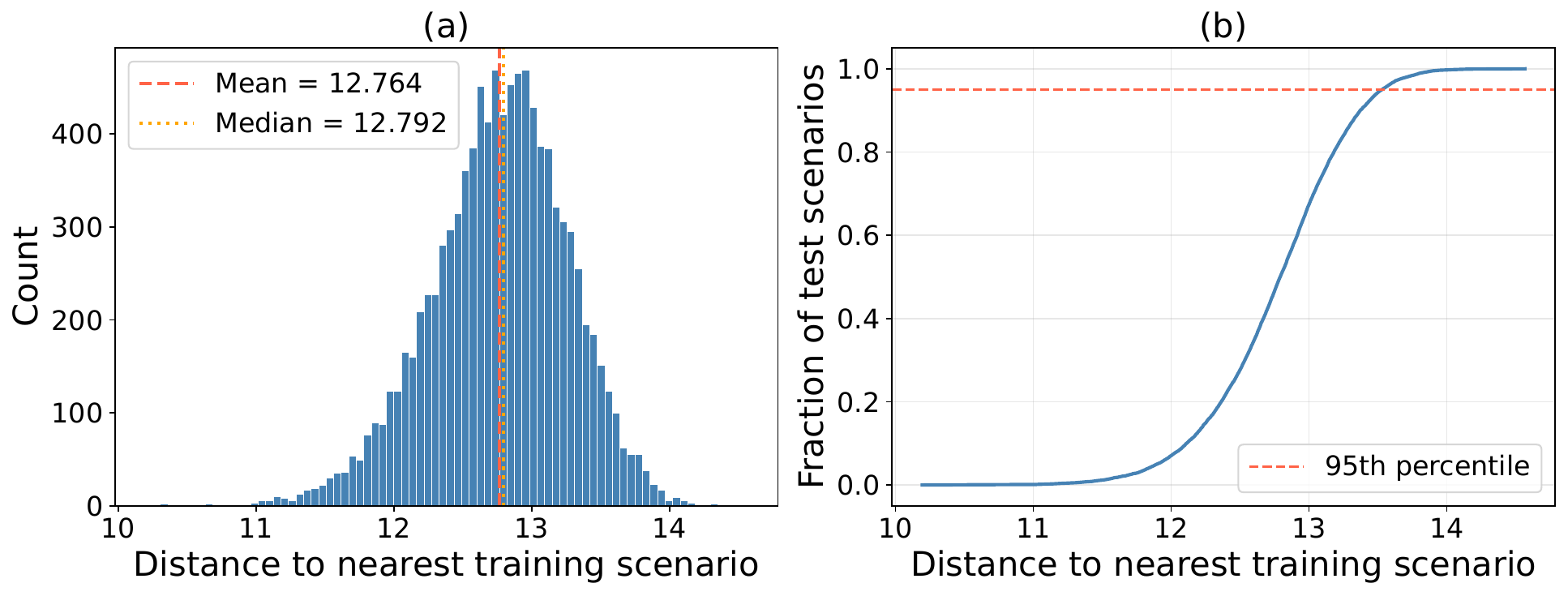}
    \caption{\footnotesize{Instance-level novelty of test scenarios. \textbf{Left:} Histogram of nearest-neighbor distances from each test scenario to its closest training counterpart (mean $= 12.76$, min $= 10.19$). \textbf{Right:} CDF of the same distances, showing that fewer than $5\%$ of test scenarios lie within a threshold of $11.90$ of any training scenario. Together these confirm that no test scenario is a near-duplicate of a training instance.}}
    \label{fig:nn_novelty}
\end{figure}

\begin{figure}[!t]
    \centering
    \includegraphics[width=\columnwidth]{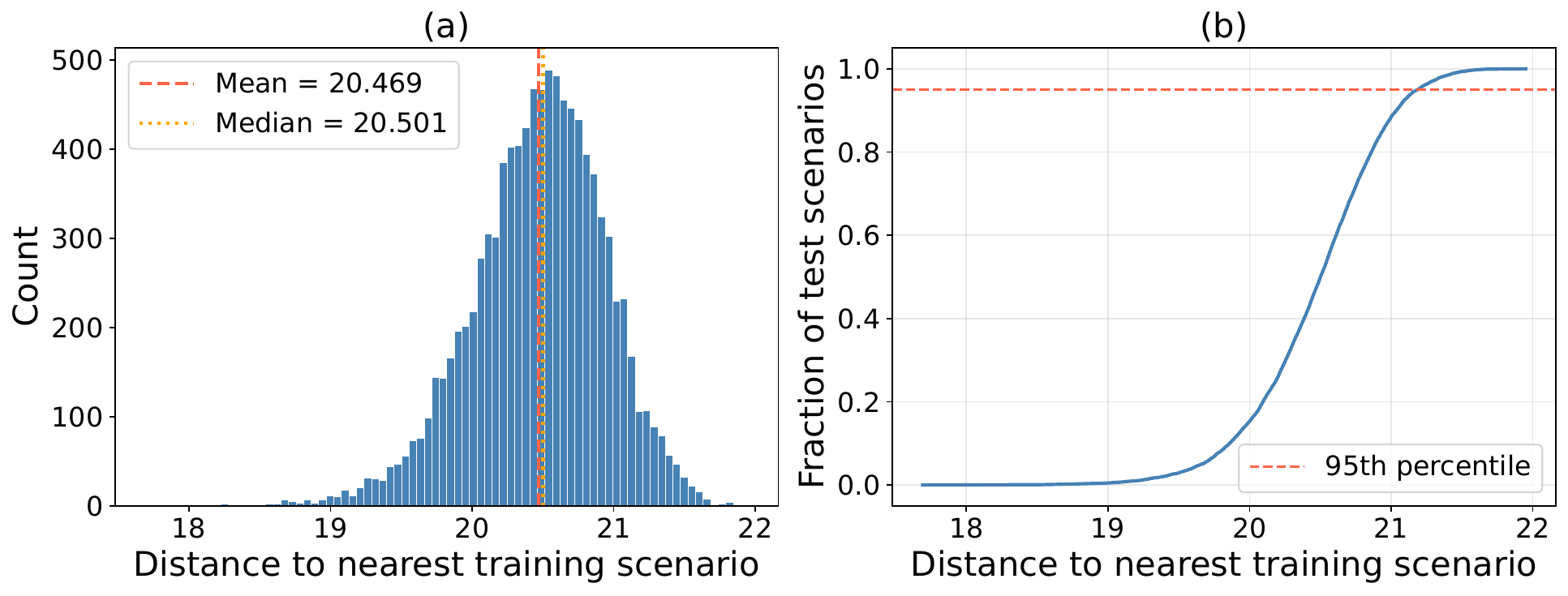}
    \caption{\footnotesize{\textcolor{revision_color}{Instance-level novelty of test scenarios for 32 agents. \textbf{Left:} Histogram of nearest-neighbor distances from each test scenario to its closest training counterpart (mean $= 20.47$, min $= 17.69$). \textbf{Right:} CDF of the same distances, showing that fewer than $5\%$ of test scenarios lie within a threshold of $19.66$ of any training scenario. Together these confirm that no test scenario is a near-duplicate of a training instance.}}}
    \label{fig:nn_novelty_32}
\end{figure}

\noindent \textbf{Instance-level novelty}: \textcolor{revision_color}{To directly quantify how distinct individual test scenarios are from training ones, we embed each scenario as a vector in $\mathbb{R}^{4n}$ ($n$ agents $\times$ 4 values(start and goal pair in 2D) per robot) and compute the nearest-neighbor distance from every test scenario to its closest training counterpart (Fig.~\ref{fig:nn_novelty} and Fig.~\ref{fig:nn_novelty_32}). For the 16-robot dataset, the resulting histogram is centered well above zero (mean $= 12.76$, min $= 10.19$), and the CDF shows that fewer than $5\%$ of test scenarios lie within a threshold of $11.90$ of any training scenario. The same qualitative pattern holds across all other examples discussed in the previous subsections (e.g, see \ref{fig:nn_novelty_32}): in each case the minimum nearest neighbor distance is strictly positive, ruling out near-duplication. This analysis confirms that the test sets are in-distribution yet consist of genuinely unseen instances.}

\subsection{Network Parameters}
% \noindent Table \ref{tab:flow_model_parameters}-\ref{tab:init_model_parameters} present the network parameters for the flow policy and initialization network respectively. Both the flow and initialization model can be conditioned with start-goal positions ($2 \cdot n_d$) or start-goal and initial velocities of the robots ($3 \cdot n_d$).

\noindent Tables \ref{tab:flow_model_parameters}-\ref{tab:init_model_parameters} summarize the network parameters. In the Flow model, the Start-Goal CNN encodes robot start/goal positions with $2 \cdot n_d$ channels without velocities and $3 \cdot n_d$ with velocities, producing a 256 dimension vector for the DiT. The Obstacle PointNet maps $2 \cdot n_d$ inputs per obstacle to 256 dimension features. The DiT has 4 layers, 256 dimension embeddings, 8 heads, and processes sequences of length $S$, while the Feed-Forward outputs $S \times (n_d \cdot n_{\xi})$. In the Initialization model, the Transformer replaces the DiT (1 layer, 256 dimension, 8 heads), and the Feed-Forward outputs $S \times (2 \cdot n_d \cdot n_{\xi})$, predicting both $\boldsymbol{\xi}$ and $\boldsymbol{\lambda}$. In both models, $S$ is typically the number of robots $n$.

\begin{table*}[!t]
\centering
\caption{\textbf{Detail of the Flow model}: where $n$ denotes the number of robots, $n_d$ dimension of workspace, $n_{obs}$ for the number of obstacles, $n_{\xi}$ for the polynomial basis order, and $S$ is a tunable hyperparameter, but often set to be equal to the number of robots}
\label{tab:flow_model_parameters}
\renewcommand{\arraystretch}{1.2} % Increase row spacing
    \begin{tabular}{|l|c|c|c|c|c|c|}
    \hline
    Model & \begin{tabular}[c]{@{}c@{}}In Channel\\ or In Seq Len\end{tabular} & 
    \begin{tabular}[c]{@{}c@{}}Out Channel\\ or Out Seq Len\end{tabular} & 
    \# Layers & Input Size & Output Size & \# Heads \\
    \hline
    DiT & $S$ & $S$ & 4 & $256$ & $256$ & $8$ \\ 
    \hline
    Start-Goal CNN & $2 \cdot n_d$ / $3 \cdot n_d$ & -- & 6 & $n$ & $256$ & -- \\ 
    \hline
    Obstacle PointNet & $2 \cdot n_d$ & -- & 6 & $n_{\text{obs}}$ & $256$ & -- \\ 
    \hline
    Feed Forward & -- & -- & 1 & $S \times 256$ & $S \times \cdot n_d \cdot n_{\xi}$ & -- \\ \hline
    \end{tabular}
\end{table*}

\begin{table*}[!t]
\centering
\caption{\textbf{Detail of the initialization model used}: where $n$ denotes the number of robots, $n_d$ dimension of workspace, $n_{obs}$ for the number of obstacles, $n_{\xi}$ for the polynomial basis order, and $S$ is a tunable hyperparameter but often set to be equal to the number of robots}
\label{tab:init_model_parameters}
\renewcommand{\arraystretch}{1.2} % Increase row spacing
    \begin{tabular}{|l|c|c|c|c|c|c|}
    \hline
    Model & \begin{tabular}[c]{@{}c@{}}In Channel\\ or In Seq Len\end{tabular} & 
    \begin{tabular}[c]{@{}c@{}}Out Channel\\ or Out Seq Len\end{tabular} & 
    \# Layers & Input Size & Output Size & \# Heads \\
    \hline
    Transformer & $S$ & $S$ & 1 & $256$ & $256$ & $8$ \\ 
    \hline
    Start-Goal CNN & $2 \cdot n_d$ / $3 \cdot n_d$ & -- & 6 & $n$ & $256$ & -- \\ 
    \hline
    Obstacle PointNet & $2 \cdot n_d$ & -- & 6 & $n_{\text{obs}}$ & $256$ & -- \\ 
    \hline
    Feed Forward & -- & -- & 1 & $ S \times 256$ & $S \times 2 \cdot n_d \cdot n_{\xi}$ & -- \\ \hline
    \end{tabular}
\end{table*}

\begin{table*}
    \centering
    \caption{Dimensions of Matrices used in Appendix. Dimensions of the vectors that act along with these matrices can be deduced easily. Some symbols are defined in Table \ref{tab:notation}}
    \label{tab:symbols_dimension}
    \scriptsize
    \renewcommand{\arraystretch}{1.2}
    \begin{tabular}{|c|c|}
        \hline
        \text{Symbol} & \text{Dimension} \\ 
        \hline
        $\mathbf{P}$ & Rows: number of obstacles ($m$) times planning horizon. Columns : $n_{\xi}$ (number of polynomial coefficients which acts as optimization variables) \\ \hline
        $\mathbf{A}$ &  Rows : number of boundary conditions times number of robots ($n$). Columns: $n_{\xi}*n$ \\ \hline
        $\mathbf{G}$ & Rows : workspace dimension $n_d$ times $n$ times planning horizon. Columns: $n_{\xi}*n$ \\ \hline
        $\mathbf{h}$ &  Rows : workspace dimension $n_d$ times $n$ times planning horizon \\ \hline
        $\mathbf{F_r}$ & Rows : $n(n-1)/2$ times planning horizon. Columns  $n_{\xi}*n$ \\ \hline
    \end{tabular}
    \normalsize
\end{table*}

 % $\bar{\boldsymbol{\alpha}}$ &  \\ \hline
 %        $\bar{\boldsymbol{\beta}}$ &  \\ \hline
 %        $\bar{\mathbf{d}}$ &  \\ \hline
 %        $\mathbf{e}$ &  \\ \hline
 %        $\boldsymbol{\lambda}$ &  \\ \hline
 %        $\mathbf{s}$ &  \\ \hline
    % $\mathbf{b}$ & Rows : number of boundary conditions times number of robots ($n$) \\ \hline

%==============================================================================================================================================================================

\balance
\bibliography{references}
\bibliographystyle{IEEEtran}

\begin{IEEEbiography}[{\includegraphics[width=1in,height=1.25in,clip,keepaspectratio]{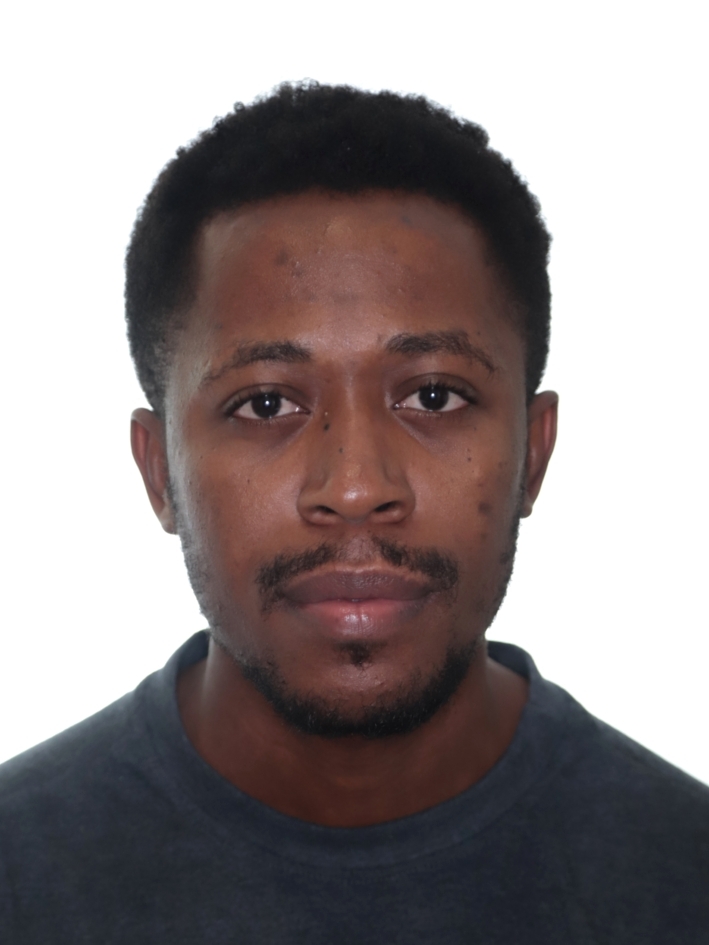}}]{Simon Idoko} received the B.Sc. degree in Electrical/Electronics Engineering in 2019 and the M.Sc. degree in Robotics and Computer Vision in 2022, with a thesis on nonlinear control of robots. He is currently pursuing the Ph.D. degree in Computer Engineering. His research interests include motion planning, control, machine learning, and reinforcement learning.
\end{IEEEbiography}

\begin{IEEEbiography}[{\includegraphics[width=1in,height=1.25in,clip,keepaspectratio]{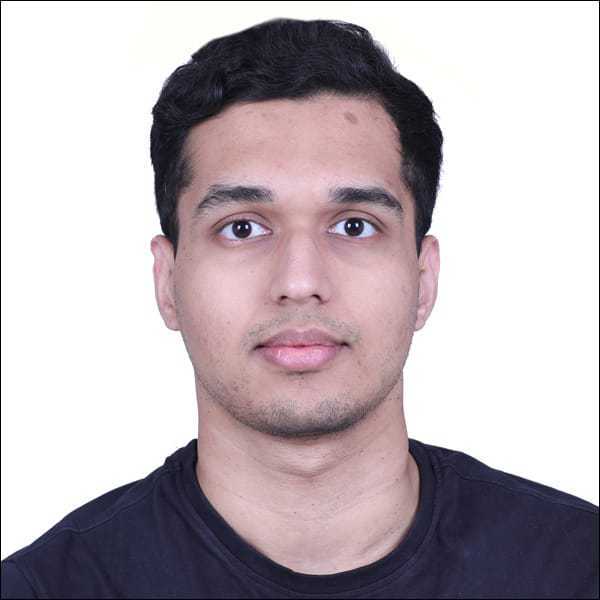}}]{Prajyot Jadhav} received the Bachelor of Technology degree in Electronics and Communication Engineering from Visvesvaraya National Institute of Technology in 2024. He is currently working as a Research Engineer with University of Tartu. His research interests include robotics, motion planning, and control.
\end{IEEEbiography}

\begin{IEEEbiography}[{\includegraphics[width=1in,height=1.25in,clip]{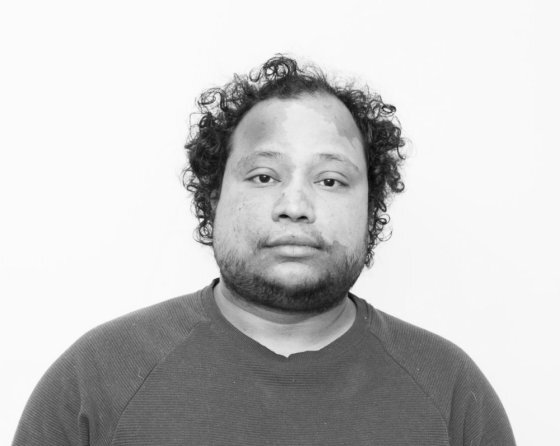}}]{Arun Kumar Singh} received PhD from IIIT-Hyderabad, India in December 2014. He is currently an Associate Professor at University of Tartu, Estonia. His research interests lie in developing optimization algorithms customized for robotic applications, embedding classical algorithmic and control theoretic reasoning into neural network pipelines, and planning under uncertainty. His works span a diverse application domain ranging from manipulation, drone navigation to autonomous driving.
\end{IEEEbiography}

\end{document}